\documentclass{article}

\PassOptionsToPackage{numbers}{natbib}
\usepackage[preprint]{neurips_2024}

\usepackage[utf8]{inputenc} 
\usepackage[T1]{fontenc}    
\usepackage{hyperref}       
\usepackage{url}            
\usepackage{booktabs}       
\usepackage{amsfonts}       
\usepackage{nicefrac}       
\usepackage{microtype}      
\usepackage{xcolor}         
\usepackage{amsfonts}
\usepackage{xspace}
\usepackage{graphicx}
\usepackage{subcaption}
\usepackage{booktabs}
\usepackage{multirow}
\usepackage{amsmath}
\usepackage{xcolor}
\usepackage{colortbl}
\usepackage{hyperref}
\usepackage{adjustbox} 
\usepackage{rotating}  
\newcommand{\PSUTMM}{PSU-TMM100\xspace}

\title{FootFormer: Estimating Stability from Visual Input}

\author{%
  Keaton Kraiger \\
  School of Electrical Engineering \\and Computer Science \\
  Pennsylvania State University \\
  University Park, PA 16802 \\
  \texttt{kbk5531@psu.edu}
  \And
  Jingjing Li \\
  College of Artificial Intelligence,\\ Cybersecurity and Computing\\
  University of South Florida \\
  Tampa, FL 33620 \\
  \texttt{jingjingli@usf.edu}
  \AND
  Skanda Bharadwaj \\
  School of Electrical Engineering \\and Computer Science \\
  Pennsylvania State University \\
  University Park, PA 16802 \\
  \texttt{ssb248@psu.edu}
  \And
  Jesse Scott \\
  Scientific Applications \\\& Research Associates (SARA), Inc. \\
  Cypress, CA  90630 \\
  \texttt{jescott@sara.com}
  \And
  Robert T. Collins \\
  School of Electrical Engineering \\and Computer Science \\
  Pennsylvania State University \\
  University Park, PA 16802 \\
  \texttt{rtc12@psu.edu}
  \And
  Yanxi Liu \\
  School of Electrical Engineering \\and Computer Science \\
  Pennsylvania State University \\
  University Park, PA 16802 \\
  \texttt{yul11@psu.edu}
}

\begin{document}

\maketitle

\begin{abstract}
We propose FootFormer, a cross-modality approach for  jointly predicting human motion dynamics directly from visual input. On multiple datasets, FootFormer achieves  statistically significantly better or equivalent estimates  of foot pressure distributions, foot contact maps, and center of mass (CoM), as compared with existing methods that generate one or two of those measures.  Furthermore, FootFormer achieves SOTA performance in estimating stability-predictive components (CoP, CoM, BoS) used in classic kinesiology metrics. Code and data are available at \href{https://github.com/keatonkraiger/Vision-to-Stability.git}{https://github.com/keatonkraiger/Vision-to-Stability.git}.

\end{abstract}

\begin{figure}[h!]
    \centering
    \includegraphics[width=0.95\linewidth]{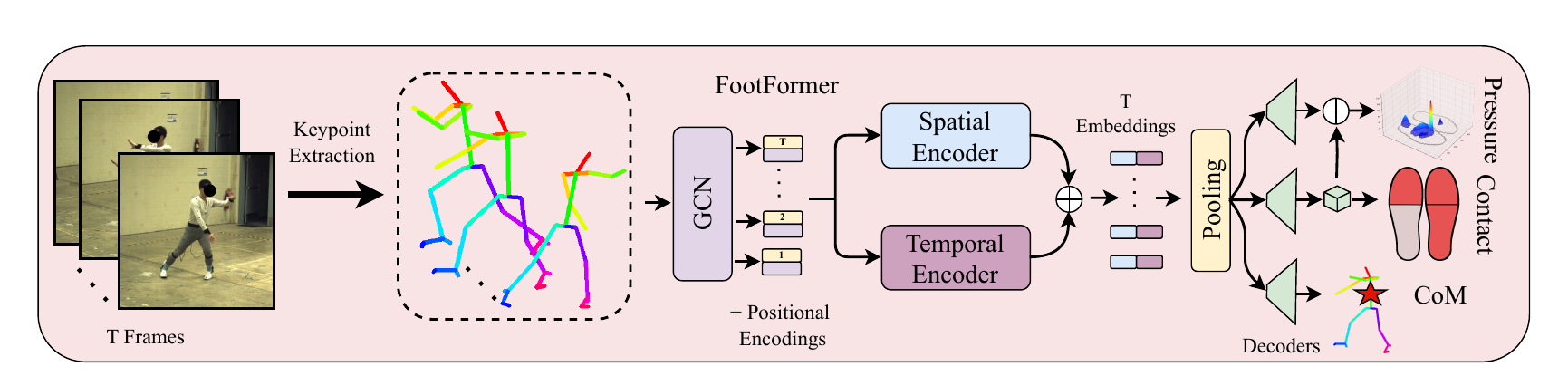}
    \vspace{0.2cm}
    \caption{The proposed cross-modality architecture FootFormer captures spatiotemporal information from visual input to directly estimate predictive measurements of human dynamics and stability. FootFormer embeds pose sequences and passes them through a spatiotemporal transformer, which is then decoded into a dense foot pressure map, contact estimation, and 3D center of mass (CoM) location, respectively.}
    \label{fig:arch}
\end{figure}

\section{Introduction}
\label{sec:intro}

Despite extensive work on estimating human body pose and motion~\cite{openpose, ChenNIPS14, chu2017multi,iqbal2017posetrack,JainTLB14,papandreou2017towards,PfisterCZ15,song2017thin,DeepPose14,convposemachines2016,Newell2016, Bulat2016}, significantly less attention has been paid to inferring physical quantities such as foot pressure and stability. Consider the Center of Pressure (CoP), which marks the net point of reactive force between a person and the ground plane. CoP's location relative to the whole body center of mass has been identified as a determinant of stability in human motion~\cite{Hof2007,Hof2008,Pai2003}. Typically, CoP is obtained in a lab setting by force plates or insole foot pressure sensors~\cite{GroundLink,UnderPressure}. Yet, as CoP is correlated with whole body kinematics, specifically mass and acceleration of body parts~\cite{winter2009biomechanics}, estimating it visually in natural settings is plausible~\cite{ScottECCV2020, JesseRehabJournal}. Vision models that infer motion dynamics quantities like CoP may enable scalable video-based analyses of human balance and stability, with applications in kinesiology, animation, and biomedical analysis. While prior work has explored video-based ground contact force estimation~\cite{UnderPressure,GroundLink,li2019motionforcesfromvideo}, the methods often only regress global scalar forces, omitting the rich structure of foot-ground interaction. Recently, large datasets such as \PSUTMM~\cite{ScottECCV2020} and MMVP~\cite{mmvp} have collected synchronized video, motion capture (MoCap), and high-resolution insole foot pressure data. However, research on these datasets is often limited to regressing a single modality from single-frame input~\cite{ScottECCV2020,JesseRehabJournal} or to augment 3D pose estimation methods~\cite{mmvp}.

We make the following contributions: 
\begin{enumerate}
    \item We propose a new cross-modality network, FootFormer (Figure~\ref{fig:arch}), for jointly estimating motion dynamics (foot pressure, foot contact, and center of mass) from visual input, unlike prior methods that predict only one or two modalities (Table~\ref{tab:model_capabilities}).
    \item We validate FootFormer on  \PSUTMM~\cite{ScottECCV2020}, MMVP~\cite{mmvp},  UnderPressure~\cite{UnderPressure}, and a newly collected Ordinary Movements dataset, and demonstrate FootFormer's efficacy compared to other methods in achieving significantly better or equivalent performance on all three output forms, especially its statistically significant SOTA performance on stability estimation (Table~\ref{tab:stability}).
    \item For foot pressure distribution prediction, in particular, we demonstrate FootFormer's ability to generalize by evaluating the pretrained model on new video-pressure sequences containing previously unseen, ordinary movements.
\end{enumerate}

\begin{table}[h!]
\centering
\small
\resizebox{0.95\linewidth}{!}{%
\begin{tabular}{l|c|c|c|c}
\toprule
\textbf{Method} & 
\textbf{\begin{tabular}{c}Foot\\Pressure\end{tabular}} & 
\textbf{\begin{tabular}{c}Foot\\Contact\end{tabular}} & 
\textbf{\begin{tabular}{c}Center of\\Mass\end{tabular}} & 
\textbf{\begin{tabular}{c}CoP/\\BoS$^*$ \end{tabular}} \\
\midrule
PNS~\cite{ScottECCV2020} & \cellcolor{green!25}\checkmark & \cellcolor{red!25}$\times$ & \cellcolor{red!25}$\times$ & \cellcolor{green!25}\checkmark \\
FPP-Net~\cite{mmvp} & \cellcolor{green!25}\checkmark & \cellcolor{green!25}\checkmark & \cellcolor{red!25}$\times$ & \cellcolor{green!25}\checkmark \\
UP~\cite{UnderPressure} & \cellcolor{green!25}\checkmark & \cellcolor{green!25}\checkmark & \cellcolor{red!25}$\times$ & \cellcolor{green!25}\checkmark\\
CoMNet~\cite{JesseRehabJournal} & \cellcolor{red!25}$\times$ & \cellcolor{red!25}$\times$ & \cellcolor{green!25}\checkmark & \cellcolor{red!25}$\times$ \\
\textbf{FootFormer (Ours)} & \cellcolor{green!25}\checkmark & \cellcolor{green!25}\checkmark & \cellcolor{green!25}\checkmark & \cellcolor{green!25}\checkmark \\
\bottomrule
\end{tabular}
}
\vspace{0.3cm}
\caption{Model output capabilities across different modalities. \checkmark indicates direct output, $\times$ indicates no output capability. $^*$CoP/BoS derived from foot pressure predictions.}
\label{tab:model_capabilities}
\end{table}

\section{Related Work}
\label{sec:relatedwork}

\subsection{From Kinematics to Ground Contact Dynamics} 
\label{subsec:related_work_kinematics}
Prior works have explored estimating contact forces from kinematic and video inputs~\cite{brubakerContactDynamics,vondrakPhysicalSimulation,brubakerAnthropomorphicWalker,VideoMocapWeiChai2010,DataDrivenInverseDynamicsSiggraphAsia,li2019motionforcesfromvideo}, typically estimating simple vertical ground reaction forces (vGRFs) or binary foot contact, unlike the dense pressure distributions or foot-region contact used in our method. Dynamics constraints are often applied in postprocessing~\cite{RempeContactDynamics2020,PhysCap2020,ShimadaPhysionical} to enforce physically plausible solutions. Alternatively,~\cite{SimPOEcvpr21,dynamicsRegulatedEgopose} interleave kinematic predictions with physics-based simulation on a causal, frame-by-frame basis, designing and learning humanoid controllers in simulation~\cite{VondrakMocapBipedControl2012,SimPOEcvpr21}. Other studies~\cite{Zell2020HumDynamics,UnderPressure,GroundLink} analyze dynamics by observing MoCap data to estimate motion dynamics and exterior forces. While similar, our objective is to enable stability estimation with a more complete approximation of motion-stability that includes foot pressure, yielding center of pressure, base of support, foot contact, and center of mass (Table~\ref{tab:model_capabilities}). More recent work~\cite{Lhoste2024DeepLearningEO} utilizes an LSTM to predict 
a scalar gait stance interpolation value for exoskeleton control. Similarly,~\cite{mmvp} proposes a GRU-based network to estimate foot pressure and contact to augment 3D pose estimation. Conversely, in~\cite{RenExoskeletonTransformer}, a transformer is used to predict hip and knee joint angles given plantar pressure for purposes of exoskeleton control. In a similar vein, we utilize spatiotemporal pose inputs but focus attention on estimating dynamics that determine human stability.

\subsection{Measuring Human Stability and Balance} 
\label{subsec:related_work_stability}
Humans naturally sense and maintain balance~\cite{Asslander_2015}, and the human visual cortex is attuned to observing other people's balance and stability~\cite{Firestone_2016}. Quantitative evaluation of stability in research and clinical applications  often use force plates to capture 3D forces for each foot while capturing body movements with MoCap technology~\cite{Chaudhry_2011,chaudhry_2004,POPOVIC_2000}. A broad selection of mathematical models have been developed for stability, and a wide set of stability metrics have been defined in the literature~\cite{Bruijn_2013,Hof2008}. Recently reported works, including novel pendulum models~\cite{JessicaInvertedPendulum} and deep learning for "On-Demand Balance Evaluation"~\cite{wei2020using,mihalec2022balance}, are almost all limited to gait movements, synthesis (simulation)-oriented, dependent on lab force plates and MoCap systems, and most important, do not take video as a primary input.  

Scott et al.~\cite{ScottECCV2020} proposed PressNet-Simple to estimate foot pressure distributions and subsequently compute Center of Pressure (CoP) and Base of Support (BoS) on \PSUTMM dataset, which contains a large variety of pose orientations, two key components for stability analysis. Later work~\cite{JesseRehabJournal} added estimation of 3D body Center of Mass (CoM) to compute two classic stability measures, CoM-CoP and CoM-BoS. Du et al.~\cite{MultiTaskCoPGraphConvNetworks,ViewInvariantCoP} use predominantly frontal pose sequences with both feet stationary on the ground to estimate CoP  measures such as path length and sway area, which can indicate balance problems.

\section{Data}
\label{sec:data}

\subsection{\PSUTMM}
\label{subsec:psutmm}
\PSUTMM is a multimodal dataset of 100 human motion sequences (each 5min long) in which 10 participants perform 24-form Taiji (Tai Chi) (Figure~\ref{fig:dataset_comparison} top 2 rows)~\cite{ScottECCV2020}. \PSUTMM includes spatiotemporally-synchronized measurements of foot pressure insoles~\cite{tekscan_2020}, MoCap markers~\cite{Vicon_2020}, and two RGB video views. Because we are primarily interested in estimating foot pressure directly from vision, we use predicted OpenPose~\cite{openpose} 2D and 3D-triangulated joint positions provided in the dataset, instead of the MoCap data.

We follow the preprocessing proposed with \PSUTMM: the OpenPose 2D and 3D joints are centered about the hip and z-score normalized per joint dimension. The raw pressure data is first clipped to the insole's recording range (0-862 kPa). Then, because we predict foot pressure distributions as opposed to absolute pressure, we divide each frame by the total pressure. This removes the need for our model to implicitly learn to estimate subject weight and instead focus on spatial distribution of pressure. To define foot contact regions, we divide contact maps into $N$ uniform regions and classify a region in contact if the maximum pressure value exceeds a given threshold (10kPa in our case). Ground truth CoM is derived from Vicon's Plug-in-Gait model~\cite{Vicon_2020}. 

\subsection{Ordinary movements}
\label{subsec:ordinary}
\begin{figure}[h!]
\centering
\centering
\setlength{\tabcolsep}{0.5pt}
\renewcommand{\arraystretch}{0.1}

\begin{tabular}{ccccccccccc}
\includegraphics[width=0.09\linewidth]{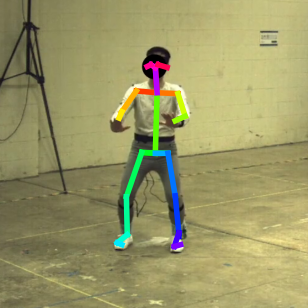} &
\includegraphics[width=0.09\linewidth]{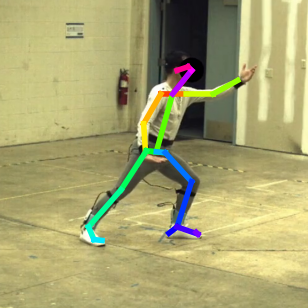} &
\includegraphics[width=0.09\linewidth]{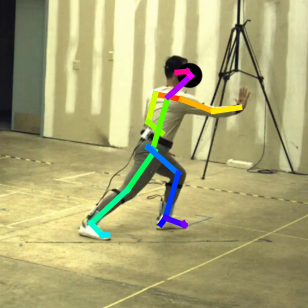} &
\includegraphics[width=0.09\linewidth]{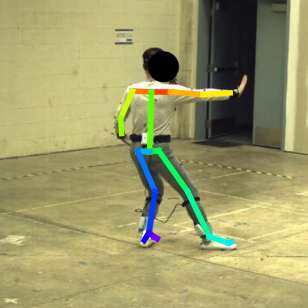} &
\includegraphics[width=0.09\linewidth]{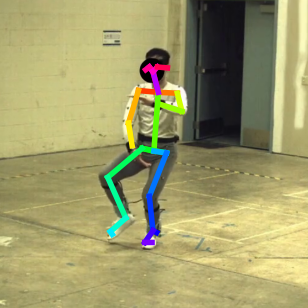} &
\includegraphics[width=0.09\linewidth]{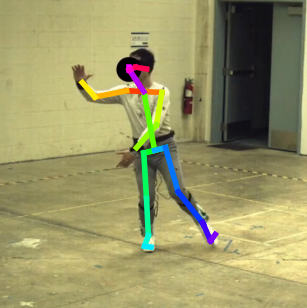} &
\includegraphics[width=0.09\linewidth]{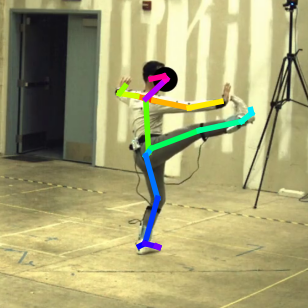} &
\includegraphics[width=0.09\linewidth]{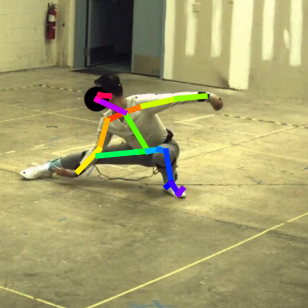} &
\includegraphics[width=0.09\linewidth]{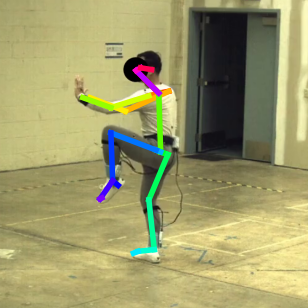} &
\includegraphics[width=0.09\linewidth]{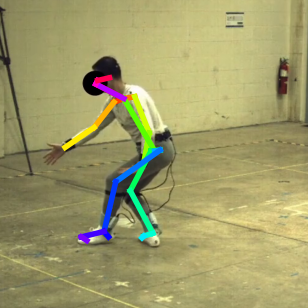} &
\includegraphics[width=0.09\linewidth]{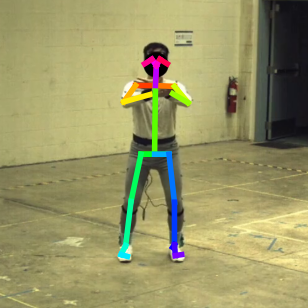}
\end{tabular}

\begin{tabular}{ccccccccccc}
\includegraphics[width=0.09\linewidth]{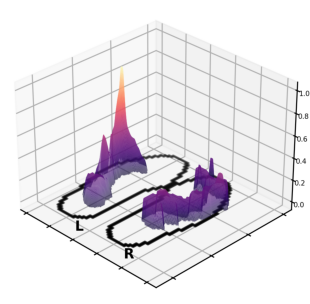} &
\includegraphics[width=0.09\linewidth]{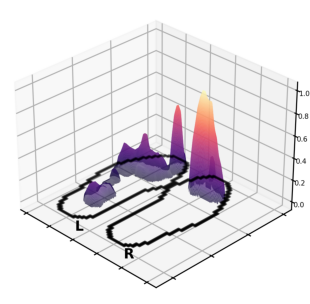} &
\includegraphics[width=0.09\linewidth]{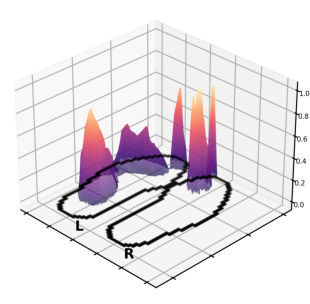} &
\includegraphics[width=0.09\linewidth]{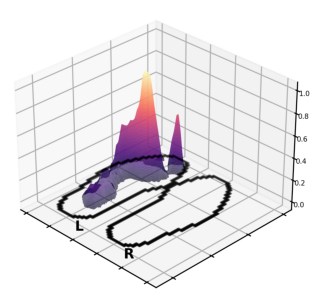} &
\includegraphics[width=0.09\linewidth]{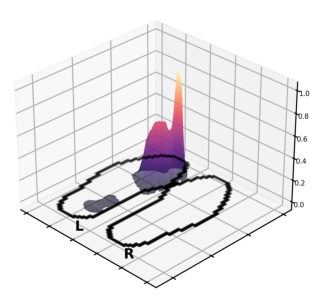} &
\includegraphics[width=0.09\linewidth]{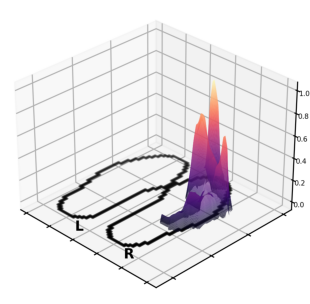} &
\includegraphics[width=0.09\linewidth]{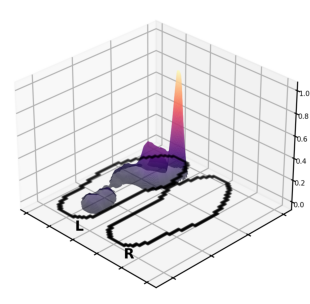} &
\includegraphics[width=0.09\linewidth]{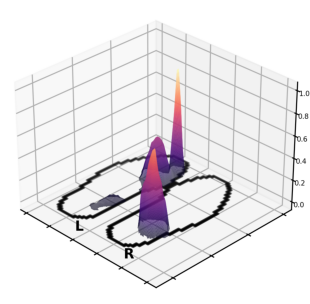} &
\includegraphics[width=0.09\linewidth]{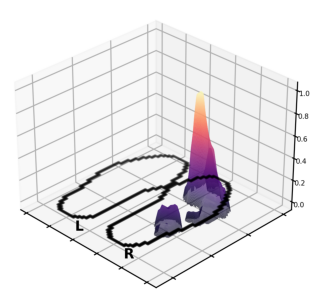} &
\includegraphics[width=0.09\linewidth]{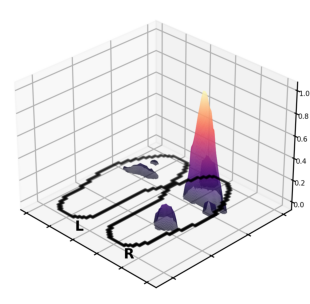} &
\includegraphics[width=0.09\linewidth]{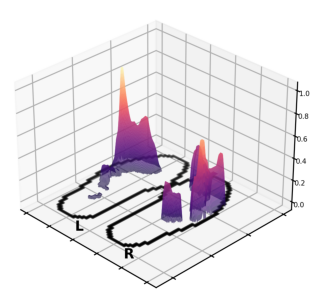}
\end{tabular}

\begin{tabular}{ccccccccccc}
\includegraphics[width=0.09\linewidth]{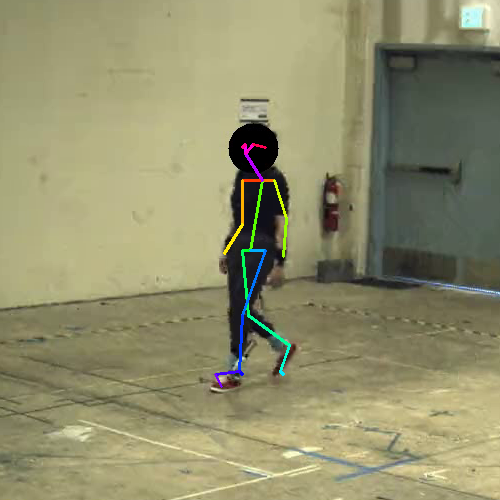} &
\includegraphics[width=0.09\linewidth]{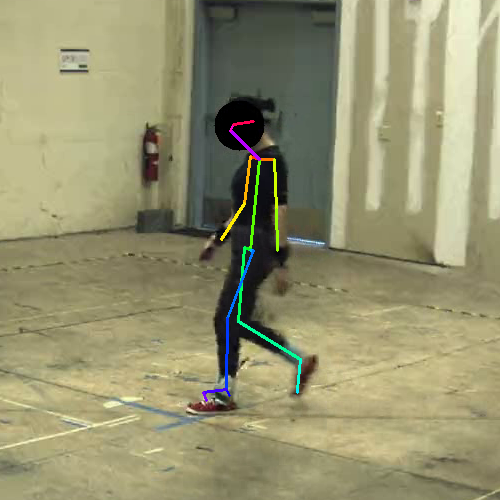} &
\includegraphics[width=0.09\linewidth]{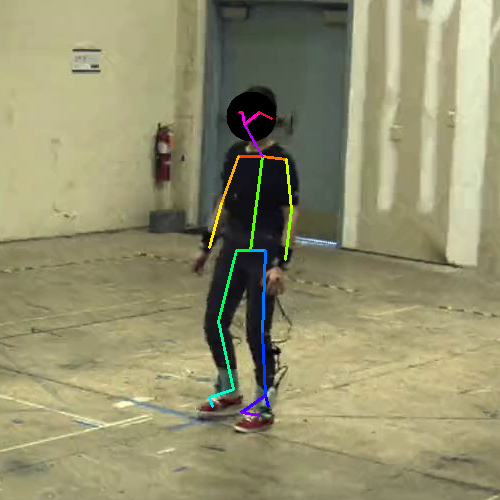} &
\includegraphics[width=0.09\linewidth]{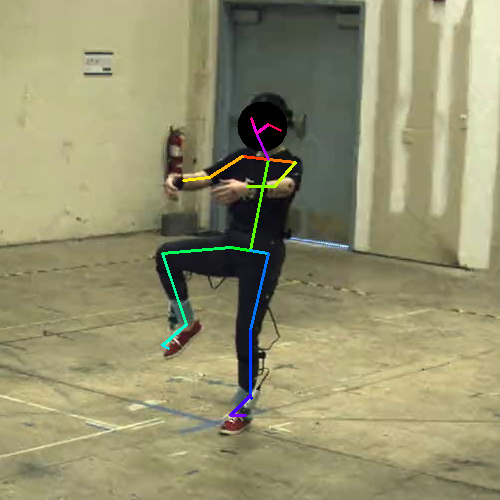} &
\includegraphics[width=0.09\linewidth]{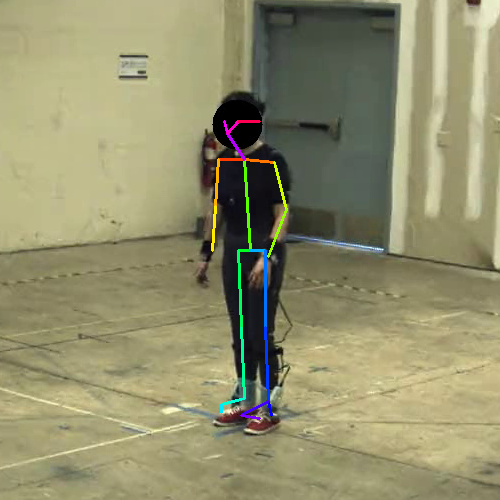} &
\includegraphics[width=0.09\linewidth]{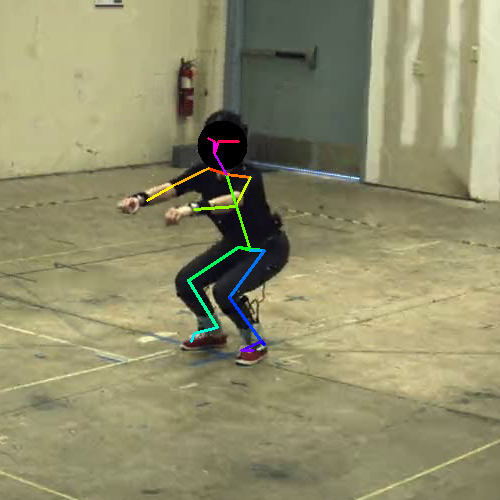} &
\includegraphics[width=0.09\linewidth]{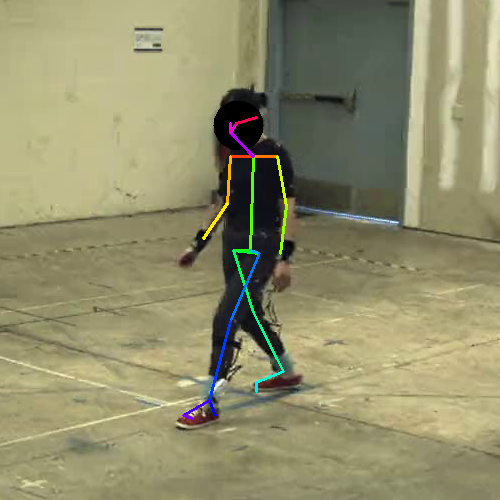} &
\includegraphics[width=0.09\linewidth]{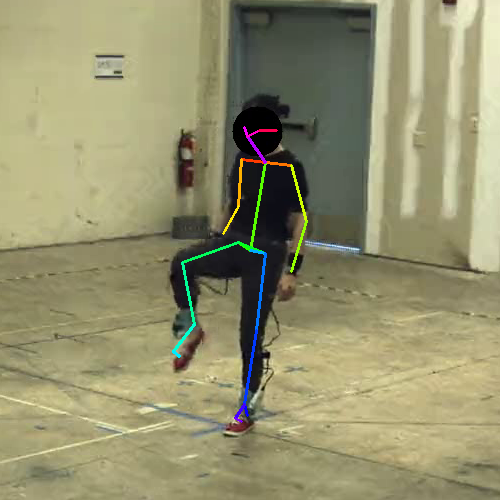} &
\includegraphics[width=0.09\linewidth]{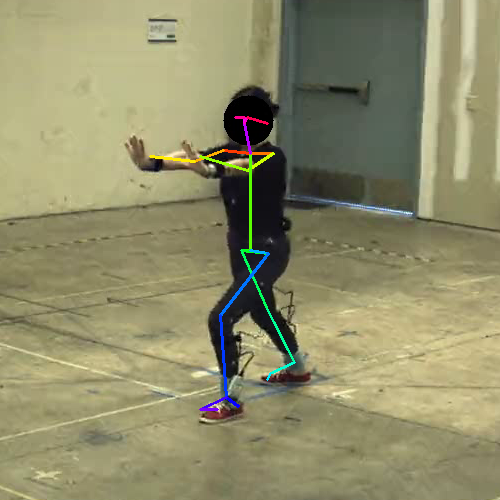} &
\includegraphics[width=0.09\linewidth]{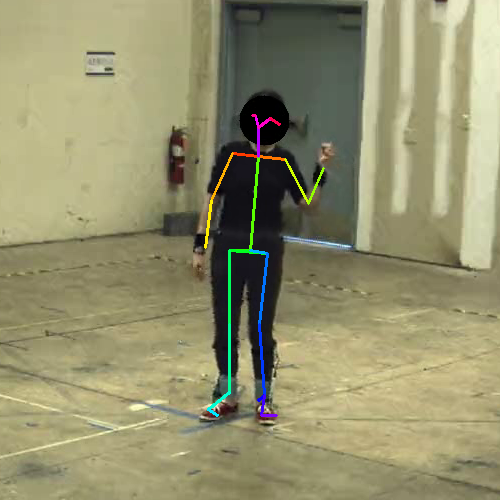} &
\includegraphics[width=0.09\linewidth]{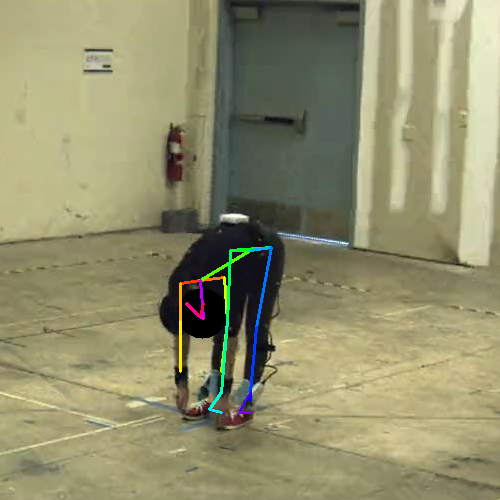}
\end{tabular}

\begin{tabular}{*{11}{p{0.09\linewidth}}}
\centering\scriptsize Circular Walking & 
\centering\scriptsize Straight Walking & 
\centering\scriptsize Lateral Step & 
\centering\scriptsize Single Leg Stand & 
\centering\scriptsize Calf\\ Raise & 
\centering\scriptsize Squat \\ Rep & 
\centering\scriptsize Forward Lunge & 
\centering\scriptsize Leg \\ Kick & 
\centering\scriptsize Push \& Pull & 
\centering\scriptsize Throwing \\Ball& 
\centering\scriptsize Full-Body Stretches
\end{tabular}

\begin{tabular}{ccccccccccc}
\includegraphics[width=0.09\linewidth]{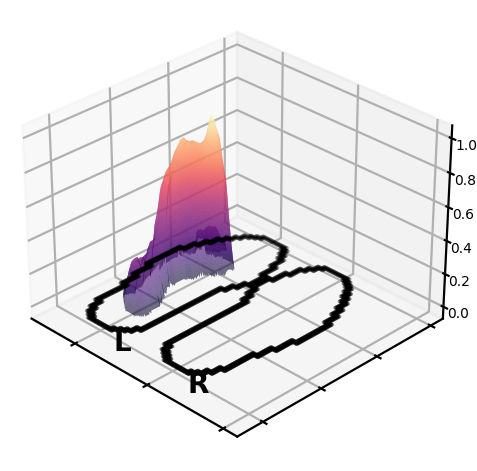} &
\includegraphics[width=0.09\linewidth]{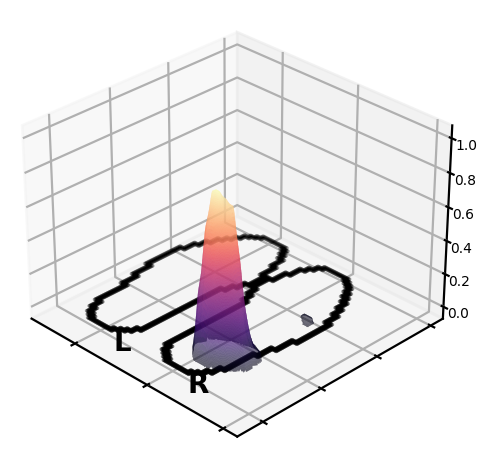} &
\includegraphics[width=0.09\linewidth]{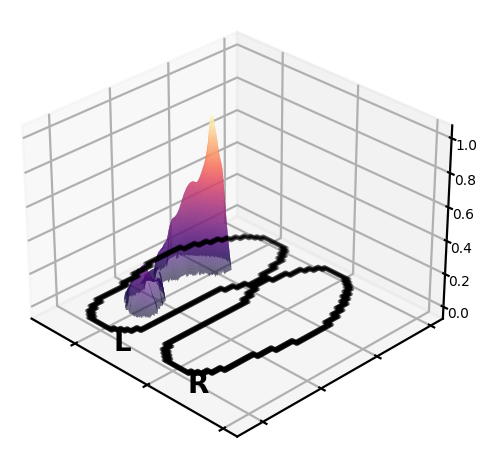} &
\includegraphics[width=0.09\linewidth]{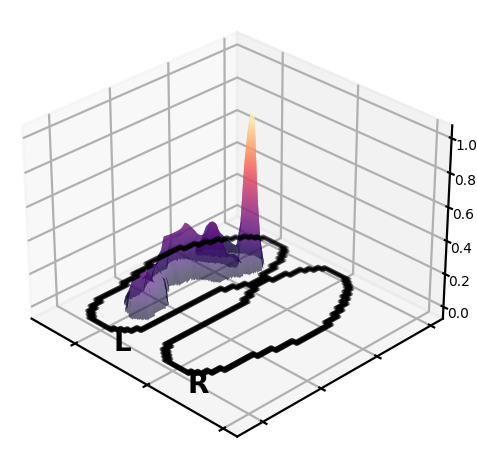} &
\includegraphics[width=0.09\linewidth]{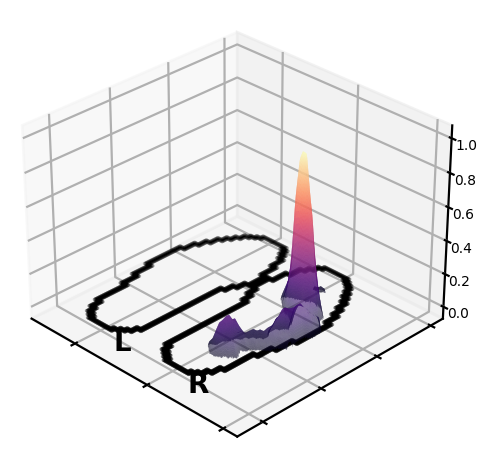} &
\includegraphics[width=0.09\linewidth]{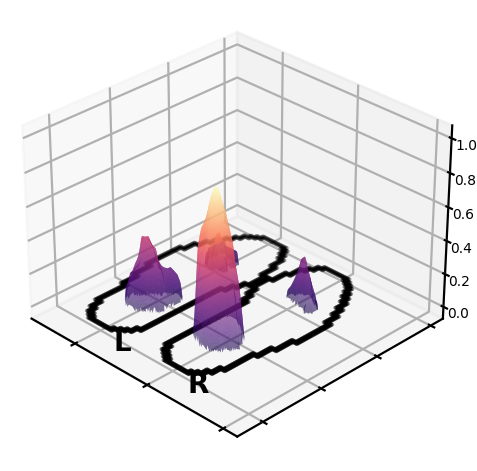} &
\includegraphics[width=0.09\linewidth]{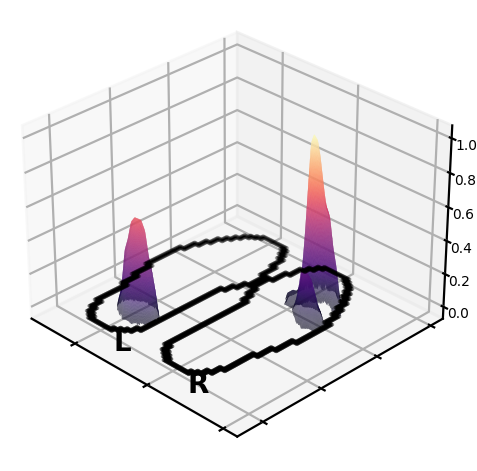} &
\includegraphics[width=0.09\linewidth]{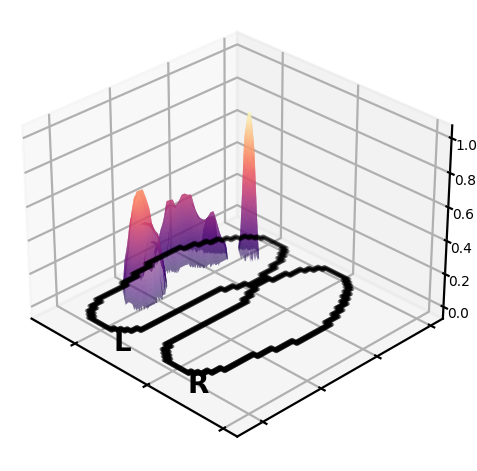} &
\includegraphics[width=0.09\linewidth]{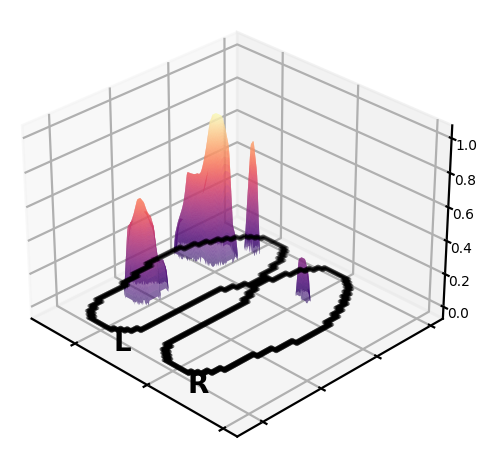} &
\includegraphics[width=0.09\linewidth]{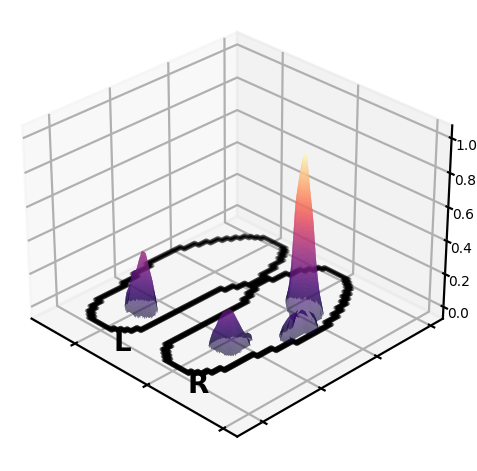} &
\includegraphics[width=0.09\linewidth]{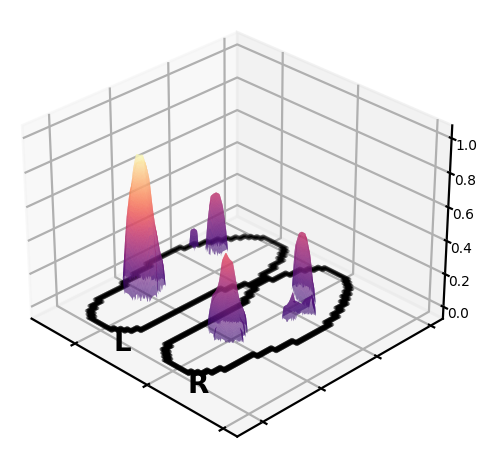}
\end{tabular}

\vspace{3mm}

\caption{Sample Taiji poses and matching foot pressures measured from \PSUTMM (rows 1-2) versus sample Ordinary Movements (rows 3-4) poses and matching foot pressures.}
\label{fig:dataset_comparison}
\end{figure}

To further evaluate our method and its ability to generalize, we collect a set of ordinary movements (OM) (Figure~\ref{fig:dataset_comparison}, rows 3-4). The data is composed of basic motions and exercises such as walking, squatting, and lunging. Similar to \PSUTMM, OM includes spatiotemporally synchronized measurements of foot pressure and two calibrated video views. Unlike the Taiji performances in \PSUTMM, the OM recordings are much shorter, with an average recording length of 40 seconds, and have faster movements. The data preprocessing is identical to that of the \PSUTMM data. We use this dataset exclusively for cross-dataset generalization validations (excluded from training). 
Additional details are in the Supplementary Material.

\subsection{UnderPressure}
\label{subsec:UP}
UnderPressure (UP) is a MoCap-foot contact synchronized dataset~\cite{UnderPressure}. It consists of 10 participants performing a range of activities including locomotion, sitting, and interacting with objects such as stairs. The dataset uses an Xsens MVN MoCap system~\cite{schepers2018xsens}, providing 3D locations of 23 joints, and Moticon OpenGo Sensor Insoles~\cite{moticon2021opengo} consisting of 16 plantar pressure sensors. UnderPressure evaluates pressure detection on both region contact and vertical ground reaction force (vGRF). 

We use the preprocessing steps from UnderPressure~\cite{UnderPressure}: joint positions are computed through forward kinematics from joint angles, with data augmentation applied via skeleton morphology variations using precomputed SVD basis vectors. Binary contact labels for the heel and toe-region of each foot are determined by whether the sum of smoothed vGRFs within those regions exceed a threshold of $5\%$ for the respective subject's body weight.

\subsection{MMVP}
\label{subsec:MMVP}
The dataset closest to \PSUTMM is the Multimodal MoCap Dataset with Vision and Pressure (MMVP)~\cite{mmvp}. MMVP is composed of 16 participants and provides synchronized Azure Kinect~\cite{microsoft2025azurekinect} RGBD video and Xsensor pressure insole~\cite{xsensor2025gaitmotion} data. MMVP provides mostly short (10 second maximum) motions, including dancing, jumping, and other exercises. The dataset provides relatively dense insole pressure maps ($\sim500$ pressure sensors), foot contact labels extracted from 3D body meshes,  accurate 2D and 3D body representations of dynamics, and comparatively fast-paced motions to that of \PSUTMM.

We follow the preprocessing steps originally proposed in MMVP~\cite{mmvp} for obtaining foot contact maps. For each frame, the insole pressure data is normalized to [0,1] by first dividing the pressure data by the respective subject's body weight and applying a Sigmoid operation to the weight-normalized pressure. Foot contact labels are then determined (empirically~\cite{mmvp}) by setting the contact threshold to 0.5. For 2D keypoint extraction, we opt to use OpenPose~\cite{openpose} instead of RTMPose~\cite{jiang2023rtmpose}.

\section{Method}
\label{sec:methods}

Our proposed method, FootFormer (Figure~\ref{fig:arch}), learns a mapping of human poses to foot pressure, foot contact, and center of mass (CoM), respectively, enabling quantification of human stability. 

\subsection{Problem Formulation}
\label{subsec:sequence}
Motion can be represented as a temporal sequence $S=\{x_i\}_{i=1}^T$ where $x_i$ denotes a pose, at time step $i$, represented as 2D joint coordinates extracted using  OpenPose~\cite{openpose} keypoints, and $T$ is the sequence length. Given a  sequence $S$ centered on the target pose $x_t$, FootFormer regresses three stability-related modalities: foot pressure distributions $P_t \in \mathbb{R}^{P'}$, foot contact maps $C_t \in \{0,1\}^{N}$, and the 3D center of mass $m_t \in \mathbb{R}^3$. The pressure map $P_t$ contains $P'$ flattened pressure values across both feet, while the contact map $C_t$ indicates binary contact states for $N$ discrete regions across both feet. We empirically set $T=9$, allowing the model to leverage four frames before and after the target frame.

Formally, our proposed FootFormer is a neural network parameterized by $\theta$ that maps a sequence of poses to the swtability-related modalities of the center pose: $\Phi_{\theta}(S)=\{P_t, C_t, m_t\}$.

\subsection{Architecture}
\label{subsec:arch}
FootFormer processes temporal pose data through an encoder-decoder architecture (Figure~\ref{fig:arch}) with: (1) a pose encoder extracting spatial embeddings, (2) a spatiotemporal transformer for sequence modeling, and (3) task-specific decoders.

\textbf{Graph Convolutional Network (GCN)}: We encode spatial structures using a GCN with learnable connectivity~\cite{mao2019learning}. Each pose $x_i$ in the input sequence $S=\{x_i\}_{i=1}^T$ is represented as a fully connected graph with $K$ joints and weighted adjacency matrix $A\in \mathbb{R}^{K\times K}$. For input features $X_{in}\in \mathbb{R}^{K\times F}$ where $F$ is the joint feature dimension, the GCN outputs $X_{out}=AX_{in}W$ using learnable weights $W\in \mathbb{R}^{F\times d}$. Applied to each frame, this produces spatially-enhanced embeddings $E \in \mathbb{R}^{T \times d}$, where $d$ is the embedding dimension (512). To encode temporal structure, we add positional encodings $E=E+P$.

\textbf{Spatiotemporal Transformer (STT)}: The STT applies multi-head self-attention to $E$ to model spatial and temporal dependencies. We apply a temporal attention mask to constrain attention to local temporal windows, preventing information leakage from future frames. Each layer contains a position-wise MLP applied independently to each token. Residual connections and layer normalization are included to stabilize learning.  The STT outputs refined embeddings $E' \in \mathbb{R}^{T \times d}$. We apply attention-based pooling to $E'$, producing $h_{pool} \in \mathbb{R}^{d}$ via learnable attention weights.

\textbf{Multi-Head Decoder}: Given $h_{pool}$, we use task-specific heads for pressure ($P$), contact ($C$), and CoM ($m$). CoM and contact use simple MLPs. For pressure-contact alignment, we use cross-attention where pressure features form queries $q=W_ph_{pool}$ and contact predictions form keys and values $k=v=W_c C$, where $W_p, W_c$ are learnable projections. Cross-attention refines the pressure representation $q'=\text{MHA}(q,k,k)$, then pressure predictions are gated: $P=\text{softmax}(W_f q' \odot \sigma(W_g q'))$ where $W_f, W_g$ are projection matrices.

\subsection{Loss}
\label{subsec:loss}
 FootFormer outputs foot pressure, regional foot contact, and center of mass; thus, we utilize a multi-component loss function to unify optimization across these modalities. To model pressure distribution, we employ Kullback-Leibler divergence loss $\mathcal{L}_p=D_{KL}(\log \hat{P} \parallel P)$, where $\hat{P},P$ denote the predicted and ground-truth pressure distributions, respectively. We formulate contact as a multi-label classification problem, using binary cross-entropy loss $\mathcal{L}_c=\text{BCE}(\hat{C}, C)$ where $\hat{C},C$ represent the predicted and ground-truth contact maps, respectively. When available, the CoM regression is supervised using MSE $\mathcal{L}_{com}= \| \hat{\mathbf{m}} - \mathbf{m} \|_2$ between predicted and ground-truth CoM points $\hat{m},m$. The total loss is then the weighted sum over all modalities: $\mathcal{L} = \lambda_p \mathcal{L}_p + \lambda_c \mathcal{L}_c + \lambda_{com} \mathcal{L}_{com}$.

\section{Experiments}
\label{sec:results}
\subsection{Implementation Details}
\label{subsec:implementation}
\textbf{Training Protocol:} For \PSUTMM evaluation, we follow Scott et al.'s Leave-One-Subject-Out (LOSO) cross-validation scheme~\cite{ScottECCV2020}, training FootFormer on 9 subjects and testing on the remaining, left out subject, 10 times in round robin fashion. For the UnderPressure~\cite{UnderPressure} and MMVP~\cite{mmvp} datasets' foot contact evaluation, we train our model from scratch using their respective original training protocols and code implementations. 

\textbf{Model Training:} All models are trained on an Nvidia A6000 GPU with a batch size of 512. For FootFormer, we optimize our multi-task loss ($\lambda_p,\lambda_c,\lambda_{com}=1$) using AdamW, while baselines use Adam with $\mathcal{L}_p$ for pressure prediction and $\mathcal{L}_c$ for contact (FPP-Net~\cite{mmvp} only), following their original protocols. Model-specific hyperparameters are in the Supplementary Material.

\textbf{Cross-Dataset Evaluation:} To assess cross-dataset transfer capability, FootFormer as trained above on \PSUTMM is evaluated directly on OM without fine-tuning.

\subsection{Evaluation}
\label{subsec:evaluation}
\textbf{Baselines:} We compare FootFormer trained on \PSUTMM to the following prior works: (1) \textbf{PNS}~\cite{ScottECCV2020}, a 4-layer fully connected network with added residual connections; (2) \textbf{UP}~\cite{UnderPressure} (dubbed after its dataset), a 1D CNN + MLP model originally used to estimate vGRFs from pose sequences; and (3) \textbf{FPP-Net}~\cite{mmvp}, which encodes pose via a 1D CNN encoder before using a GRU to handle the sequential data followed by a dual-headed MLP regressor which jointly predicts foot pressure and binary contact. We adapt baselines minimally to ensure a fair comparison, only adjusting input and output sizes to fit \PSUTMM. Table~\ref{tab:model_capabilities} provides an overview of model output capabilities.

\textbf{Metrics:} We report performance for three modalities: foot pressure, binary foot contact, and 3D center of mass (CoM). For foot pressure, we are interested in quantifying the normalized \textbf{pressure} distribution that facilitates Center of Pressure estimation, we report KLD distance of the predicted and ground-truth pressure distribution. 
We follow standard \textbf{foot contact} evaluation practice to report precision, recall, F1 score, and Intersection over Union (IoU) between the ground truth and predicted contact points. When evaluating \textbf{Center of Mass}, we use Euclidean error between our predicted CoM points and 3D CoM points provided in \PSUTMM measured with a Vicon MoCap system. FootFormer is the only model which jointly optimizes and outputs all three modalities (Table~\ref{tab:model_capabilities}), directly enabling the quantification of human stability metrics~\cite{jian1993trajectory,lugade2011center,Bruijn_2013}. 

\subsection{Foot Pressure}
\label{subsec:footpressure}
Table~\ref{tab:combined_kld} (Left) reports the mean KLD across all 10 subjects on LOSO experiments for both the 2D and 3D keypoints in \PSUTMM across all baselines (Table~\ref{tab:model_capabilities}). FootFormer performs statistically significantly better or equivalently across both inputs on \PSUTMM for foot pressure estimation.

\begin{table}[h!]
\centering
\small
\resizebox{0.95\linewidth}{!}{%
\begin{tabular}{lcc|cc}
\toprule
\textbf{Dataset (Metric)} 
& \multicolumn{2}{c|}{\textbf{\PSUTMM (KLD} $\downarrow$)} 
& \multicolumn{2}{c}{\textbf{OM (KLD} $\downarrow$)} \\
\cmidrule(lr){2-3} \cmidrule(lr){4-5}
\textbf{Method} 
& \textbf{2D} & \textbf{3D} 
& \textbf{2D} & \textbf{3D} \\
\midrule
PNS~\cite{ScottECCV2020}            & 2.82 $\pm$ 0.86$^{\dagger}$ & 2.68 $\pm$ 0.94$^{\dagger}$ & 3.53 $\pm$ 1.23$^{\dagger}$ & 2.59 $\pm$ 0.94$^{\dagger}$ \\
FPP-Net~\cite{mmvp}                 & 1.40 $\pm$ 0.32 & 1.60 $\pm$ 0.48$^{\dagger}$ & \textbf{1.52 $\pm$ 0.37} & 1.54 $\pm$ 0.34 \\
UP~\cite{UnderPressure}  & 1.45 $\pm$ 0.35 & 1.50 $\pm$ 0.35$^{\dagger}$ & 1.69 $\pm$ 0.34 & \textbf{1.48 $\pm$ 0.41} \\
Ours                       & \textbf{1.36 $\pm$ 0.29} & \textbf{1.22 $\pm$ 0.32} & 1.56 $\pm$ 0.40 & 1.53 $\pm$ 0.22 \\
\bottomrule
\end{tabular}
}
\vspace{0.3cm}
\caption{Foot pressure estimation evaluated using KL Divergence (KLD) on \PSUTMM~\cite{ScottECCV2020} and Ordinary Movements (OM) datasets. Results are averaged across all subjects using leave-one-subject-out (LOSO) cross-validation. \textbf{Bold} indicates the best (lowest KLD); $^{\dagger}$ denotes a statistically significant difference from FootFormer (Ours) under paired \textit{t}-test (\textit{p}~<~0.05).  FootFormer performs statistically significantly better or equivalently across the two datasets.}
\label{tab:combined_kld}
\end{table}
 
We consider how well the model generalizes to non-Taij movements by training on the \PSUTMM dataset and testing on a new dataset of eleven Ordinary Movements. Table~\ref{tab:combined_kld} (Right) reports KLD for each baseline on 2D and 3D input over all collected movements. Despite an overall lower performance than on Taiji sequences, FootFormer is still able to generalize to these completely unseen movements, achieving significantly better than PNS and equivalent to FPP-Net and UP. We provide qualitative examples of the different model predictions for each baseline model in the Supplementary Material. 

\subsection{Foot Contact}
\label{subsec:contact}
Estimating foot contact (FC), or whether a foot or specific parts of the foot are in contact with the ground plane, is essential for applications in locomotion analysis, rehabilitation, graphics, and animation. We compare FootFormer against FPP-Net~\cite{mmvp} and UP~\cite{UnderPressure} (Table~\ref{tab:model_capabilities}) on their respective datasets to assess effectiveness in contact prediction. Table~\ref{tab:contact} presents the precision (prec.), recall, F1-score, and IoU evaluation scores. 

\begin{table}[h!]
\small
\centering
\resizebox{0.95\linewidth}{!}{%
\begin{tabular}{llcccc}
\hline
    \textbf{Model} & \textbf{Dataset} & \textbf{prec.} $\uparrow$ & \textbf{recall} $\uparrow$ & \textbf{F1} $\uparrow$ & \textbf{IoU} $\uparrow$ \\
\hline\hline
    FPP-Net\cite{mmvp} & MMVP & $0.635^{\dagger}$ & \textbf{0.600} & 0.583 & 0.448 \\
    Ours & MMVP & \textbf{0.650} & 0.588 & \textbf{0.586} & \textbf{0.450} \\
\hline
    UP\cite{UnderPressure} & UnderPressure & $0.936^{\dagger}$ & $0.954^{\dagger}$ & $0.945^{\dagger}$ & $0.896^{\dagger}$ \\
    Ours & UnderPressure & \textbf{0.942} & \textbf{0.972} & \textbf{0.956} & \textbf{0.917}\\
\hline
\end{tabular}
}
\vspace{0.3cm}
\caption{Foot contact estimation results. We train and evaluate on the MMVP~\cite{mmvp} and UnderPressure~\cite{UnderPressure} datasets and compare against their respective baseline models. \textbf{Bold} indicates the best (highest metric); $^{\dagger}$ denotes a statistically significant difference from Ours under paired \textit{t}-test (\textit{p}~<~0.05).}
\label{tab:contact}
\end{table}

On the MMVP dataset, FootFormer achieves significantly better precision and equivalent F1, IoU, and recall. Moreover, our model achieves statistically significant improvements over UP across all 4 metrics. Exact \textit{p}-values are provided in the Supplementary Material. Our results for FPP-Net on the MMVP dataset may be superior to those reported in the original paper~\cite{mmvp} as we fully retrain and evaluate with OpenPose~\cite{openpose} keypoints instead of their original keypose extraction method~\cite{jiang2023rtmpose}.

\subsection{Stability Components}
\label{subsec:stabilitycomponents}

\begin{figure}[h!]
    \centering
    \includegraphics[width=0.5\linewidth]{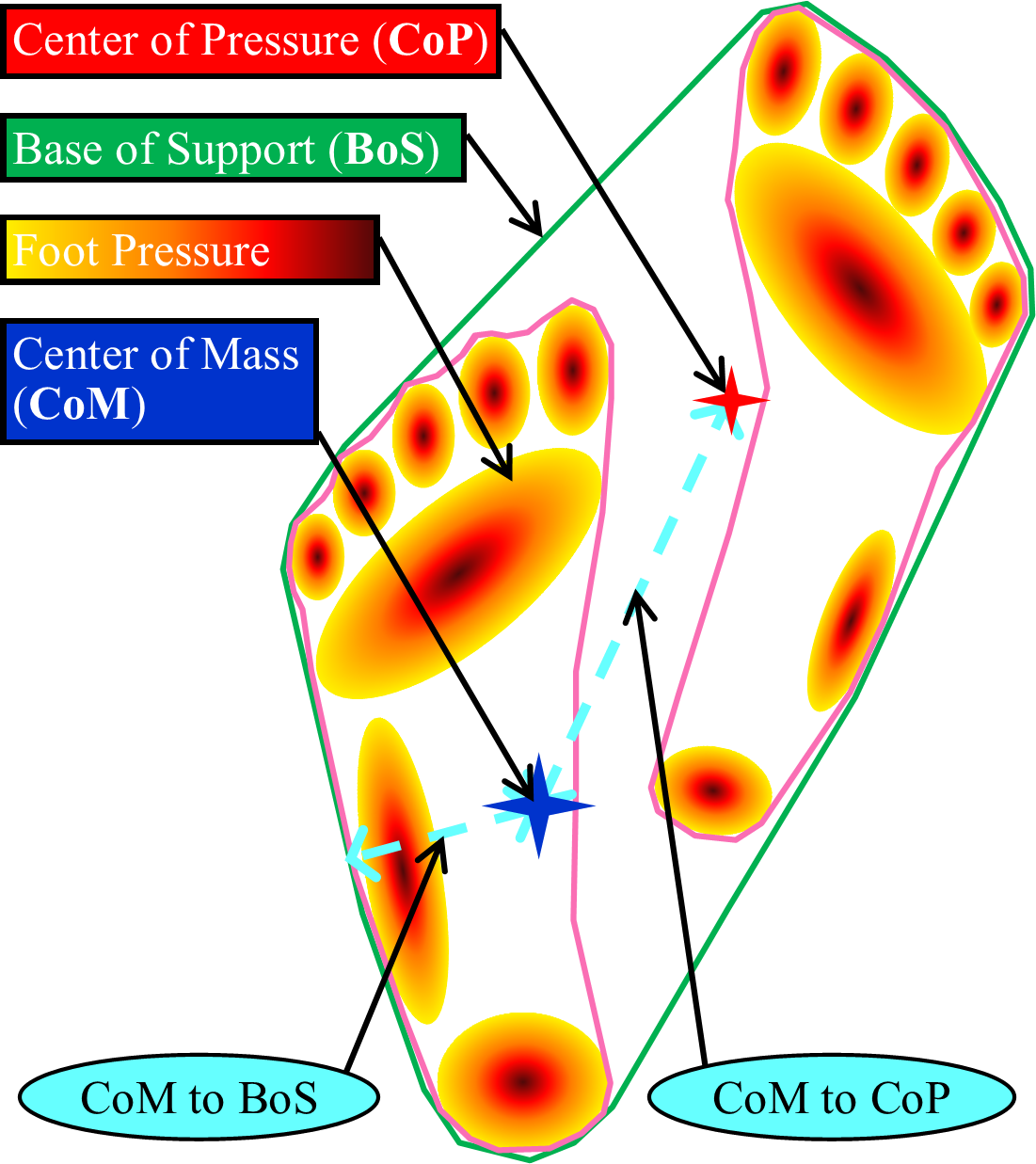}
    \caption{Foot plantar pressure annotated with Center of Pressure (CoP), Base of Support (BoS), and Center of Mass (CoM) projected onto the floor plane. Two stability metrics are shown: Com-BoS (2D distance from CoM to BoS boundary) and CoM-CoP (2D distance from CoM to CoP).}
    \label{fig:stability_diagram}
\end{figure}

Our goal is comprehensive stability analysis.  To this end, we evaluate estimates of three stability components, CoP, CoM and BoS, that form the foundation for quantifying human postural stability and balance~\cite{jian1993trajectory,lugade2011center,Bruijn_2013}. Figure~\ref{fig:stability_diagram} depicts foot pressure with these stability components and two stability components used in their calculation. 

\begin{figure}[h]
\centering
\begin{tabular}{cc}
    \includegraphics[width=0.75\linewidth]{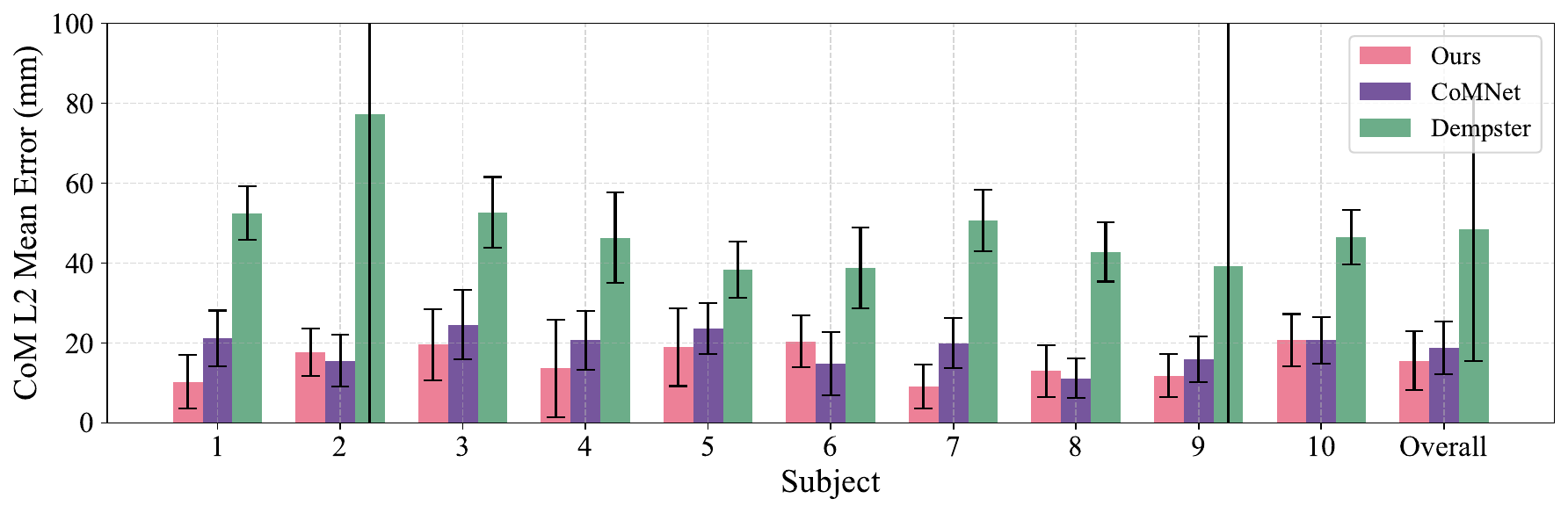} \\
    (a) Mean Center of Mass L$_2$ Error \\
    
    \includegraphics[width=0.75\linewidth]{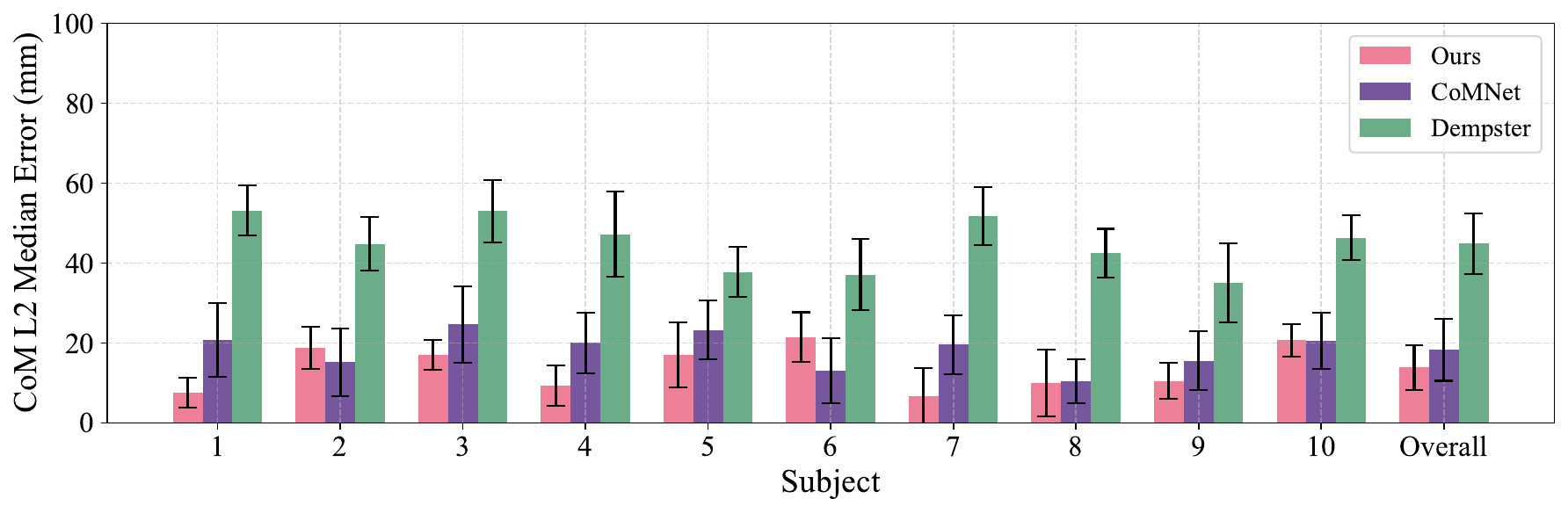} \\
    (b) Median Center of Mass L$_2$ Error \\
    
    \includegraphics[width=0.75\linewidth]{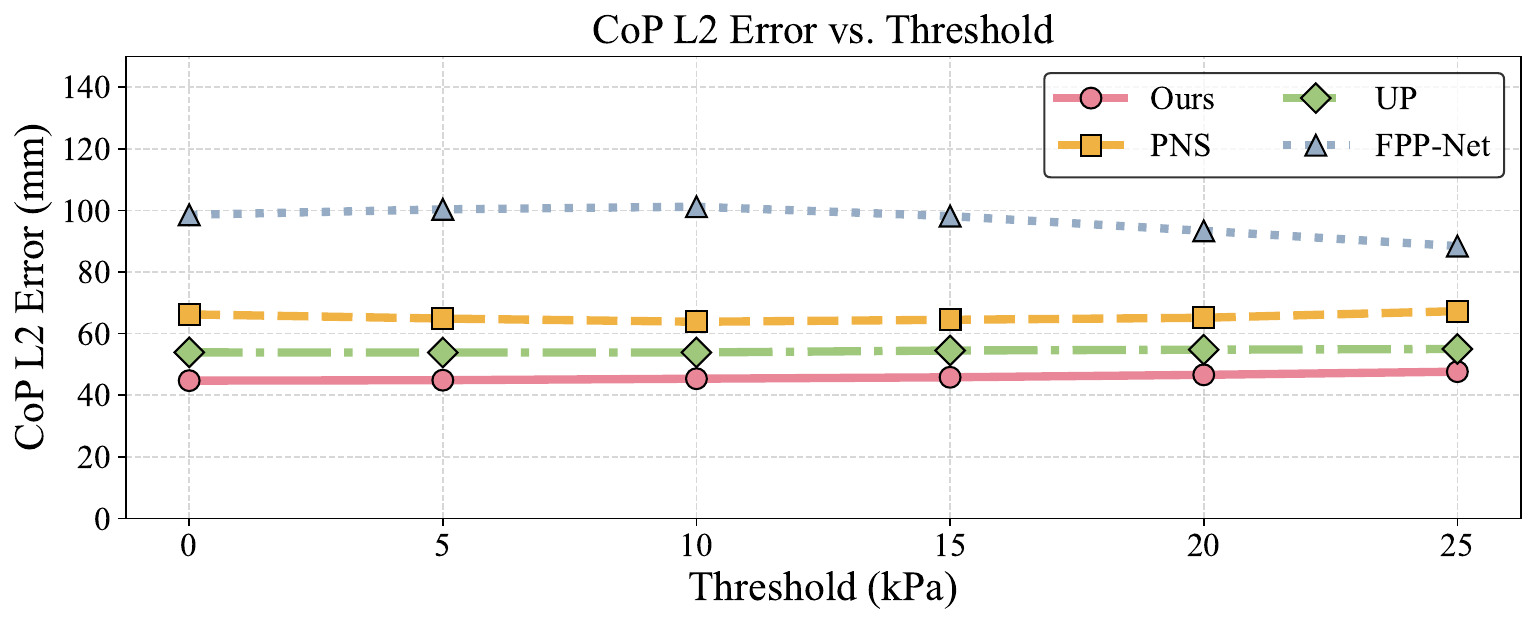} \\
    (c) Center of Pressure L$_2$ Error\\
    \includegraphics[width=0.75\linewidth]{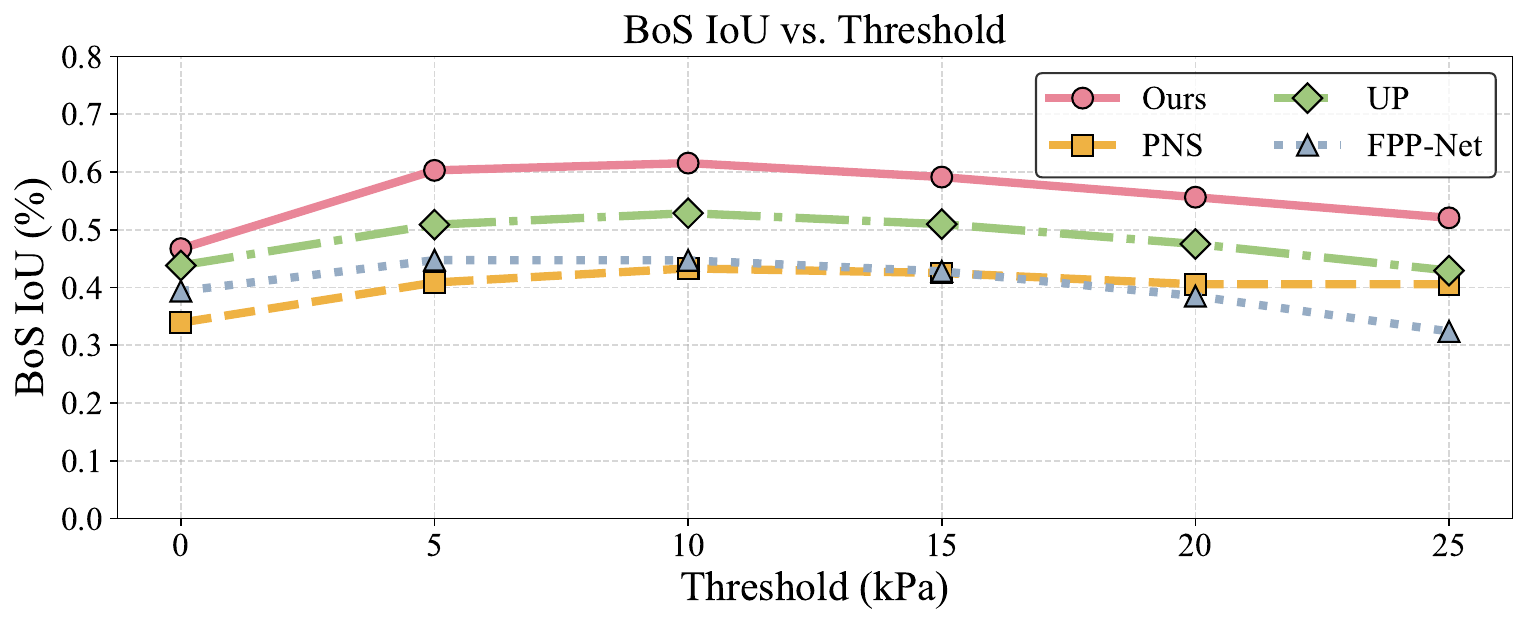} \\
    (d) Base of Support IoU\\
\end{tabular}
\vspace{0.3cm}
\caption{Performance comparison on ~\PSUTMM~\cite{ScottECCV2020}. (a,b) Mean and median CoM L$_2$ error across subjects. (c) CoP L$_2$ error and (d) BoS IoU across varying pressure thresholds. FootFormer achieves statistically significantly better results across all stability components (see Supplementary Material for exact \textit{p}-values).}
\label{fig:stability_components}
\end{figure}

\textbf{Center of Mass (CoM):} CoM represents the weighted average position of body mass and is a crucial factor in a person's ability to maintain balance. Unlike CoP and BoS, CoM is directly regressed from keypoint sequences. We compare against two baseline methods, CoMNet~\cite{JesseRehabJournal}, a fully connected network (Table~\ref{tab:model_capabilities}), and Dempster's method~\cite{winter2009biomechanics,drillis1964body,dempster1967properties}, a classical anthropomorphic method that estimates CoM from weighted sums of segmental centers of mass across the body. For CoM evaluation, Figure~\ref{fig:stability_components}(a,b) presents both mean and median L$_2$ errors reported in millimeters (mm). FootFormer demonstrates statistically significant  improvements over both classical (Dempster's method~\cite{winter2009biomechanics,drillis1964body,dempster1967properties}) and learned (CoMNet~\cite{JesseRehabJournal}) baselines.

\textbf{Center of Pressure (CoP):} CoP is calculated as the weighted mean of the pressure elements in the XY ground plane. Accurate estimation of CoP is essential for understanding balance, as shifts in CoP can indicate changes in stability or an impending need for postural adjustment~\cite{jian1993trajectory}. \textbf{Base of Support (BoS):} BoS represents the area under the feet that supports the body. Estimating BoS is critical for determining the boundaries within which an individual can maintain balance~\cite{lugade2011center}. We follow~\cite{JesseRehabJournal} and perform foot localization to align the predicted and GT pressure maps with the floor plane using 3D-triangulated keypoints from two viewpoints. Foot position is estimated by foreshortening the pressure map based on dorsiflexion and plantar flexion angles, and rotating it according to ankle orientation ensuring spatial consistency.

We report CoP and BoS evaluation across varying pressure thresholds, with CoP error reported in mm and BoS measured as the IoU between convex hulls surrounding the predicted and ground-truth pressure maps. Figure~\ref{fig:stability_components}(c,d) shows FootFormer achieves both the lowest CoP error and highest BoS IoU across all tested pressure thresholds. We observe the best performance across all baselines when thresholding the pressure at 5-10 kPa, reducing noise in the foot insole measurements.

\subsection{Stability}
\label{subsec:stability}

Moving beyond simple kinematic estimates, the multiple outputs of FootFormer enable us to directly estimate stability. We calculate two popular measures of postural stability (Figure~\ref{fig:stability_diagram}). First, we estimate CoM-CoP defined as $\| CoM - CoP\|_2$ or the Euclidean distance from the 2D CoM projected onto the floor plane to CoP~\cite{jian1993trajectory}. Typically, the further apart these two points are, the greater the potential for becoming unstable. Second, we measure $\| CoM-BoS_{nearest}\|_2$ or the Euclidean distance from the 2D CoM to the boundary of the BoS~\cite{lugade2011center}. Intuitively, the CoM-BoS captures the magnitude of instability.

\begin{table}[h!]
    \small
    \centering
    \resizebox{0.95\linewidth}{!}{%
    \begin{tabular}{l l c|c}
    \toprule
    \textbf{Metric} & \textbf{Model} & \textbf{Mean Absolute Error (mm)} $\downarrow$ & \textbf{Median Absolute Error (mm)} $\downarrow$ \\
    \midrule

    \multirow{2}{*}{CoM-CoP} 
    & CoMNet+PNS~\cite{JesseRehabJournal}    & $46.00 \pm 22.0^{\dagger}$ & $29.86 \pm 13.2^{\dagger}$ \\
    & Ours      & $\mathbf{31.80 \pm 27.6}$ & $\mathbf{24.33 \pm 8.3}$ \\
    \midrule
    \multirow{2}{*}{CoM-BoS} 
    & CoMNet+PNS~\cite{JesseRehabJournal}    & $34.73 \pm 21.7^{\dagger}$ & $19.79 \pm 11.7$ \\
    & Ours      & $\mathbf{23.97 \pm 23.16}$ & $\mathbf{17.69 \pm 11.3}$ \\
    \bottomrule
    \end{tabular}
    }
    \vspace{0.3cm}
    \caption{Stability quantification results on \PSUTMM~\cite{ScottECCV2020}. CoM-CoP and CoM-BoS are reported in mm. \textbf{Bold} indicates the best (lowest error); $^{\dagger}$ denotes a statistically significant difference from Ours under paired \textit{t}-test (\textit{p}~<~0.05).}
    \label{tab:stability}
\end{table}

We compare with Scott et al.~\cite{JesseRehabJournal} who use CoMNet to estimate CoM and PNS~\cite{ScottECCV2020} to regress foot pressure (Table~\ref{tab:model_capabilities}). Table~\ref{tab:stability} follows~\cite{JesseRehabJournal} and reports the mean $\pm$ std and median $\pm$ rSTD for CoM-CoP and CoM-BoS error in mm, where rSTD represents robust standard deviation calculated as the median absolute deviation from the median, multiplied by 1.4826~\cite{rousseeuw1993alternatives}. Error is computed as the absolute distance between predicted and ground-truth positions derived from the insole sensors and MoCap system used in \PSUTMM. FootFormer achieves statistically significantly improvements in both stability metrics compared to the combined multi-model approach of CoMNet+PNS. We believe this validates the efficacy of learning coupled motion dynamics within a unified structure, as our joint optimization approach outperforms separate models trained independently for each component.

\subsection{Ablation Experiment}
\label{subsec:ablations}
To validate design choices of the proposed network, we systematically replace key components and evaluate on \PSUTMM's KLD foot pressure and mean L$_2$ CoM error.

\textbf{Pose Embedding:} We compare our learnable GCN against a 1D CNN~\cite{mmvp,UnderPressure} and a linear MLP~\cite{ScottECCV2020}. \textbf{Temporal Modeling}: We replace the STT with a standard transformer and GRU~\cite{mmvp} to assess the efficacy of the spatiotemporal attention mechanism. \textbf{Contact Conditioning:} We evaluate the pressure decoder with and without the contact-based gating to investigate the cross-modal alignment benefits.

\begin{figure}[h!]
\centering
\begin{tabular}{cc}
    \includegraphics[width=0.45\linewidth]{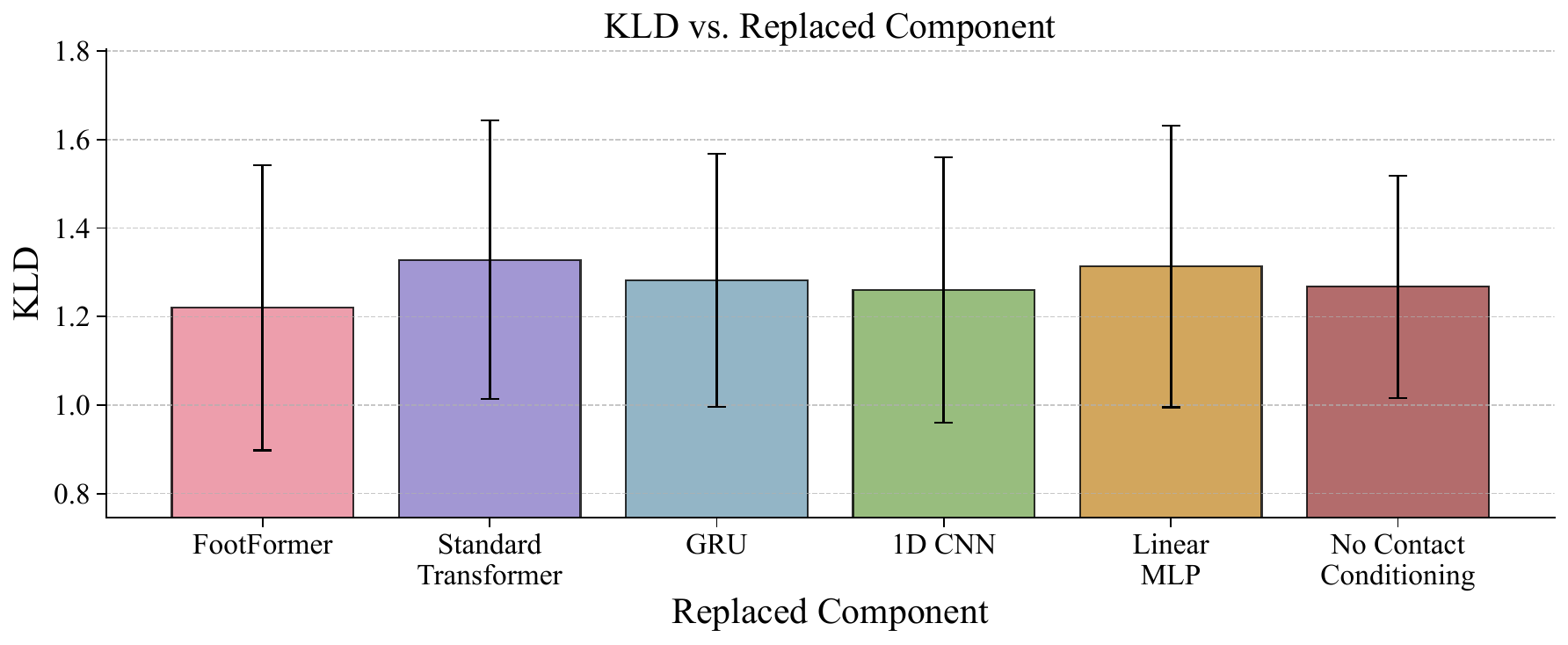} &
    \includegraphics[width=0.45\linewidth]{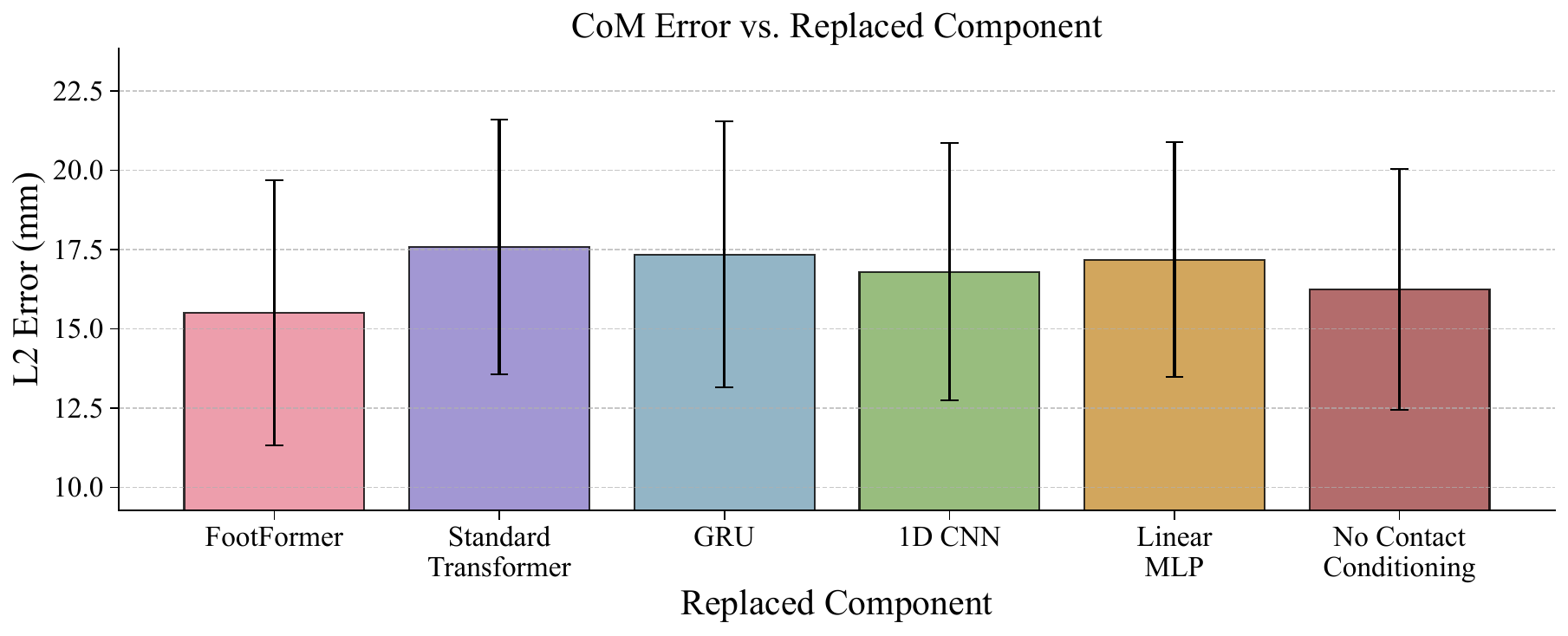} \\
(a) KL-Divergence  & (b) Center of Mass
\end{tabular}
\vspace{0.2cm}
\caption{Mean KLD and CoM Euclidean error when varying FootFormer's components.}
\label{fig:ablation}
\end{figure}

Figure~\ref{fig:ablation} shows the KLD and CoM results for FootFormer and all variants of the swapped-out components. We observe that replacing any of the key components of the network results in a degradation of both pressure and CoM prediction. Further, the contact-conditioned decoder improves both pressure and CoM prediction, demonstrating cross-model alignment.

\section{Conclusion}
\label{sec:conclusion}

We present FootFormer, a cross-modality method that jointly estimates foot pressure, foot contact maps, and center of mass from visual input in a unified model. Unlike prior approaches requiring separate networks for individual modalities (Table~\ref{tab:model_capabilities}), FootFormer achieves statistically significantly better or equivalent performance across three datasets using one single model (Tables~\ref{tab:combined_kld} and~\ref{tab:contact}, Figure~\ref{fig:com_cop_bos_comparison}). Notably, our unified approach achieves SOTA performance over combined multi-model baselines on human motion stability quantification (Table~\ref{tab:stability}). 

\textbf{Limitations and future work}: In this work, we do not incorporate additional motion-rich data sources such as IMUs or biometric sensors commonly embedded in everyday devices. Vision-based methods struggle to detect non-visual phenomena such as vertigo, for which biometric or inertial data could provide useful indirect signals. We believe learning to integrate these additional modalities deserves greater attention and plan to pursue this direction in future work.

\textbf{Acknowledgments} This work was funded in part by NSF grant 2312967, From Vision to Dynamics. 
Dr. Yanxi Liu owns AR TAIJ, LLC, which offers the free app AR TAIJI. The Penn State University Individual Conflict of Interest Committee has reviewed this research and determined that it could be perceived to be related to AR TAIJI, LLC, and this is being managed by the Committee.

\bibliographystyle{unsrtnat}
\bibliography{egbib}

\newpage
\appendix

\section*{Supplementary Material}
\section{Additional Implementation Details}
Below, we provide additional implementation details of FootFormer and each of the  baseline models.

\subsection{Detailed Architecture Specifications}
FootFormer uses a Graph Convolutional Network (GCN) with a learnable attention matrix for spatial pose encoding, producing a 256-dimensional pose embedding per frame. The Spatial-Temporal Transformer (STT) consists of 8 transformer encoder layers with 16 attention heads, dropout of 0.1, and MLP hidden size of 1024. We employ learnable positional encodings prior to passing through the STT. Since the prediction is only on the middle frame, average pooling is performed on the sequence of embeddings prior to entering the task-specific decoders, each having a hidden size of 128.

To adapt the UP~\cite{UnderPressure} model to \PSUTMM, we make simple changes to resize the input and final regression layer to fit \PSUTMM's pose input and insole pressure maps, respectively; all other network components are maintained as is. Similarly, we modify FPP-Net's~\cite{mmvp} first layer to fit the joint scheme present in the data. We then adapt the network's pressure and contact regressor to fit the insole shape and contact regions (like that of FootFormer).

\subsection{Hyperparameter Tuning}
To optimize the proposed FootFormer model we employed a staged hyperparameter tuning strategy consisting of a coarse-to-fine search. In the initial coarse phase, we performed a broad sweep over key architectural and optimization parameters to identify general performance trends. This included varying model depth, hidden dimensions, learning rates, and regularization parameters. Based on these observations, we conducted a fine-grained search in a narrower range around the best-performing settings. All tuning was conducted using validation performance averaged across subjects in a leave-one-subject-out (LOSO) setup to avoid subject-specific overfitting.

The final hyperparameters used for FootFormer were a learning rate of 2e-4 and AdamW $\beta_1=0.9$ with $\lambda_p,\lambda_c,\lambda_{com}=1$. For UP and FPP-Net, the original architectural hyperparameters were preserved, and a fine tuning of the learning rate was done in addition to tuning of $\lambda_p$ and $\lambda_c$ for FPP-Net. A final learning rate of 1e-5 and 1e-4 were used for FPP-Net and UP, respectively, with $\lambda_p=0.4$ and $\lambda_c=0.6$ being used in FPP-Net's loss weighting (a value similar to that reported in their original paper).

\section{Additional Foot Pressure Estimation Results}
\label{subsec:eval}

\begin{figure}[h!]
    \centering
    \includegraphics[width=0.95\linewidth]{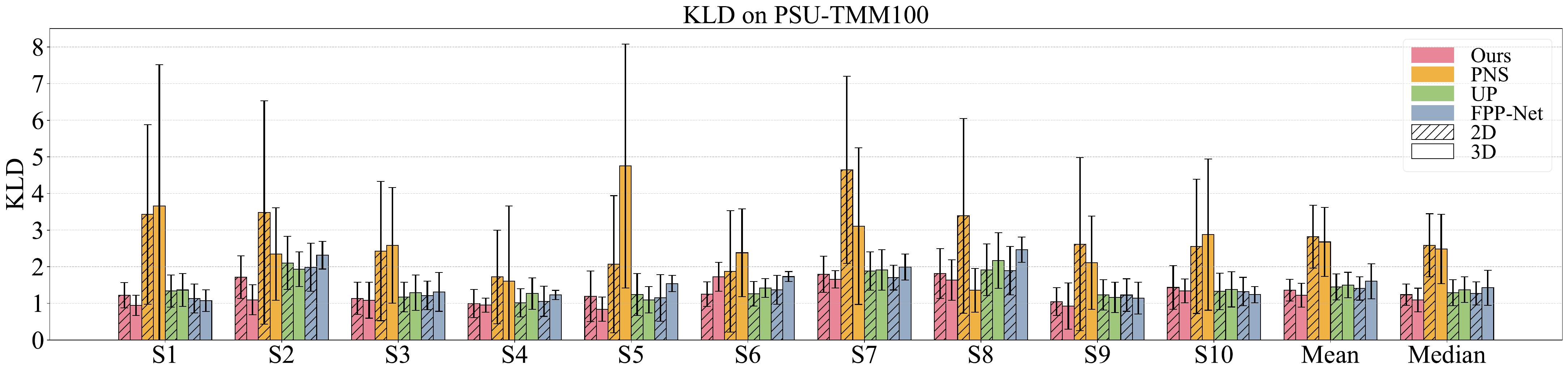}
    \vspace{0.3cm}
    \caption{KLD foot pressure estimation results on \PSUTMM~\cite{ScottECCV2020} across each subject with mean and median values. We train and evaluate on both 2D and 3D detected keypoints. Lower is better. Statistical significance values are presented in Table~\ref{tab:pval_TMM_combined}.}
    \label{fig:PSU_pressure_metrics}
\end{figure}

We report the full comparison with Scott et al.~\cite{ScottECCV2020}, UP~\cite{UnderPressure}, and FPP-Net~\cite{mmvp} on the \PSUTMM for each subject in Figure~\ref{fig:PSU_pressure_metrics}. Table~\ref{tab:pval_TMM_combined} reports statistical significance values of paired \textit{t}-tests comparing FootFormer with the baseline models.

\begin{table}[h!]
\centering
\small
\resizebox{0.95\linewidth}{!}{%
\begin{tabular}{lcc}
\toprule
\textbf{Model} & \textbf{2D KLD $\downarrow$ (\textit{p}-value vs Ours)} & \textbf{3D KLD $\downarrow$ (\textit{p}-value vs Ours)} \\
\midrule
PNS~\cite{ScottECCV2020}      
& 2.82 $\pm$ 0.86$^{\dagger}$ \, {\scriptsize($1.45\mathrm{e}{-04}$)} 
& 2.68 $\pm$ 0.94$^{\dagger}$ \, {\scriptsize($9.38\mathrm{e}{-03}$)} \\
FPP-Net~\cite{mmvp}           
& 1.40 $\pm$ 0.32 \, {\scriptsize($2.93\mathrm{e}{-01}$)} 
& 1.60 $\pm$ 0.48$^{\dagger}$ \, {\scriptsize($1.61\mathrm{e}{-02}$)} \\
UP~\cite{UnderPressure}       
& 1.45 $\pm$ 0.35 \, {\scriptsize($6.06\mathrm{e}{-02}$)} 
& 1.50 $\pm$ 0.35$^{\dagger}$ \, {\scriptsize($1.59\mathrm{e}{-02}$)} \\
Ours                 
& \textbf{1.36 $\pm$ 0.29} & \textbf{1.22 $\pm$ 0.32} \\
\bottomrule
\end{tabular}
}
\vspace{0.3cm}
\caption{Comparison of FootFormer (Ours) with baselines on \PSUTMM~\cite{ScottECCV2020}. 
Each entry reports mean $\pm$ std KLD and the paired \textit{t}-test \textit{p}-value vs.~Ours. 
\textbf{Bold} indicates the best (lowest KLD); $^{\dagger}$ denotes a statistically significant difference from Ours (\textit{p}~<~0.05).
Per-subject results are shown in Fig.~\ref{fig:PSU_pressure_metrics}.}
\label{tab:pval_TMM_combined}
\end{table}

\section{Ordinary Movements Data}
\begin{table}[h!]
\small
\centering
\begin{tabular}{r|l|r}
\hline
\textbf{\#} & \textbf{Activity} & \textbf{\# of Frames} \\
\hline
1 & Circular Walking & 4698 \\
2 & Straight Walking & 2359 \\
3 & Lateral Step & 1942 \\
4 & Single Leg Stand & 3669 \\
5 & Calf Raise & 4072 \\
6 & Squat Rep & 2136 \\
7 & Forward Lunge & 2232 \\
8 & Leg Kick & 1970 \\
9 & Push \& Pull & 1169 \\
10 & Throwing Ball & 2731 \\
11 & Full-Body Stretches & 2216 \\
\hline
\multicolumn{2}{c|}{\textbf{Total}} & \textbf{29194} \\
\hline
\end{tabular}
\vspace{0.3cm}
\caption{Summary of the collected ordinary movement motions with frame counts for each action set.}
\label{tab:om_summary}
\end{table}

The collected Ordinary Movements (OM) data is introduced separately from \PSUTMM to enable rigorously evaluating the generalization of vision-based foot pressure estimation methods beyond scripted and repetitive motions. Unlike Taiji, which consists of a long-form choreographed sequence, the OM dataset captures short, natural, everyday activities that present a broader range of human motion styles and contact patterns.

The OM dataset contains 11 distinct movement types (summarized in Table~\ref{tab:om_summary}) such as walking, squatting, lunging, kicking, single-leg stance, and push-pull motions. These activities were selected to reflect everyday physical behaviors encountered in real-world environments. Each activity lasts approximately 40 seconds on average, and the total dataset contains a total of 29,194 frames.

Each frame in OM includes the following synchronized modalities:
\begin{itemize}
    \item \textbf{Foot pressure maps:} Recorded using Tekscan F-Scan 7.0~\cite{tekscan_2020} insole sensors at 50~Hz. Each foot has a high-resolution prexel grid of size $60 \times 21$, capturing pressure intensities in kilopascals.
    \item \textbf{2D and 3D body pose:} 2D joints are extracted using OpenPose~\cite{openpose} BODY25, while 3D joints are reconstructed via stereo triangulation from the two camera views (like in \PSUTMM).
    \item \textbf{Video:} 1080p RGB video is captured at 50~Hz from two calibrated and synchronized Vicon Vue cameras, ensuring accurate visual data for each frame.
\end{itemize}

\begin{figure}[h!]
\centering
\setlength{\tabcolsep}{0.6pt}
\renewcommand{\arraystretch}{0.5} 

\begin{tabular}{c ccccc}
& \rotatebox{45}{\small{Lateral
Step}} & \rotatebox{45}{\small{Single Leg Stand}} & \rotatebox{45}{\small{Calf Raise}} & \rotatebox{45}{\small{Leg Kick}} & \rotatebox{45}{\small{Throwing Ball}} \\

\rotatebox{45}{\small{View 1}} &
\includegraphics[width=0.16\linewidth]{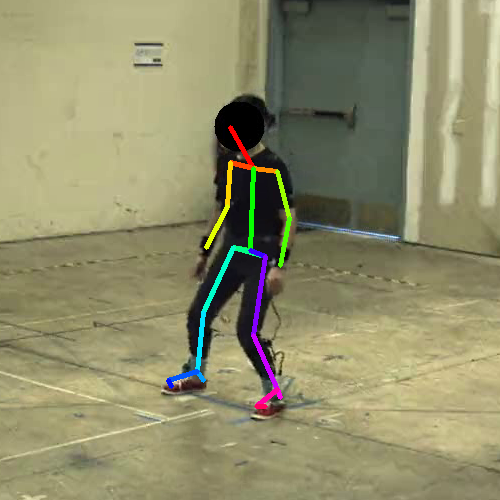} &
\includegraphics[width=0.16\linewidth]{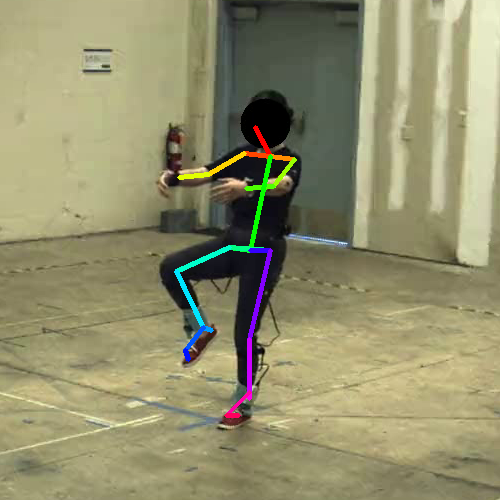} &
\includegraphics[width=0.16\linewidth]{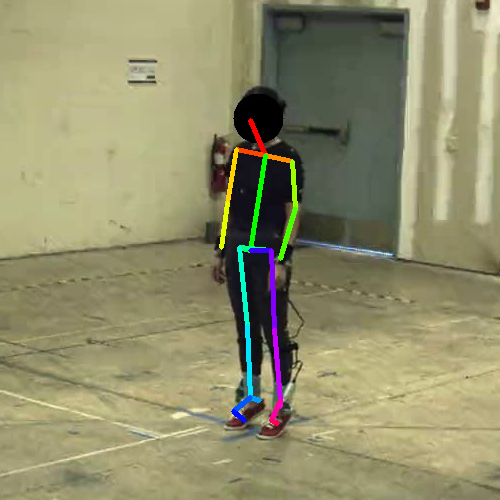} &
\includegraphics[width=0.16\linewidth]{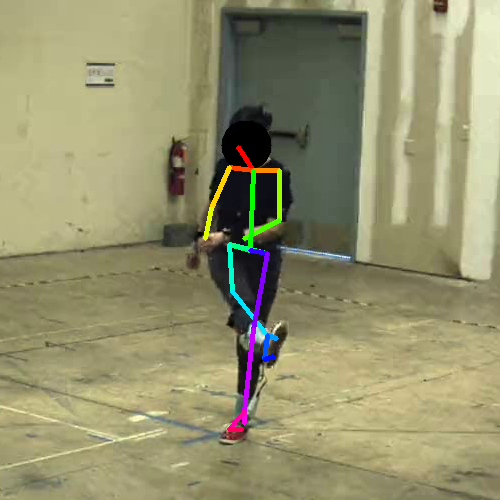} &
\includegraphics[width=0.16\linewidth]{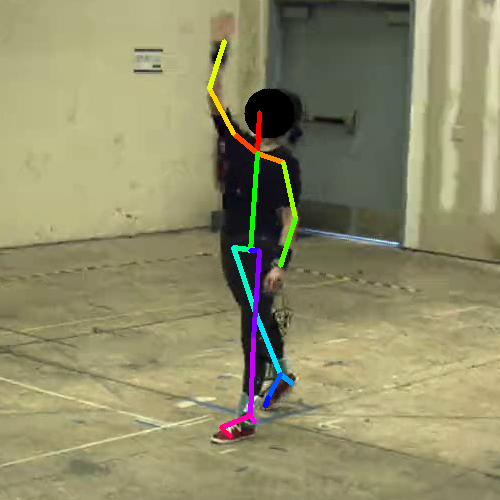} \\

\rotatebox{45}{\small{View 2}} &
\includegraphics[width=0.16\linewidth]{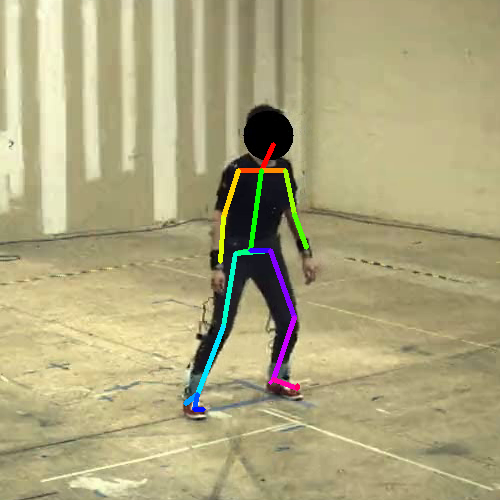} &
\includegraphics[width=0.16\linewidth]{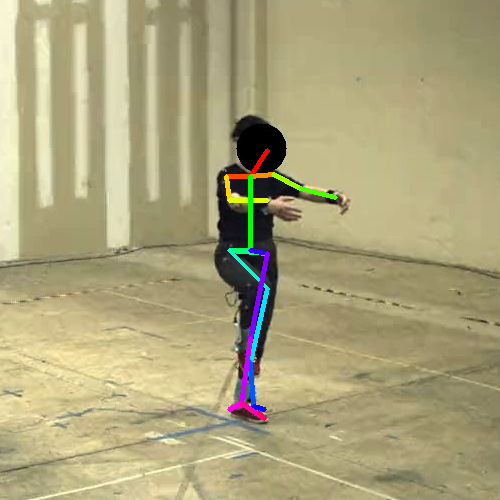} &
\includegraphics[width=0.16\linewidth]{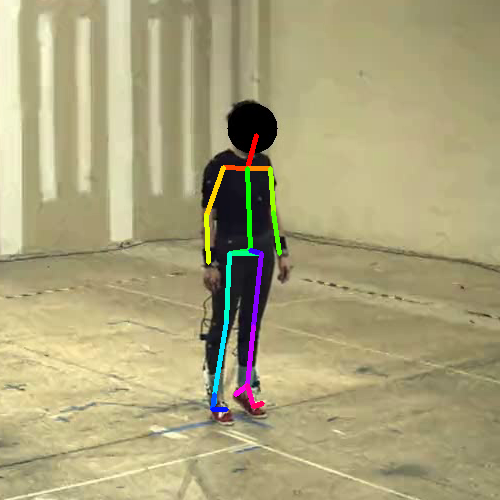} &
\includegraphics[width=0.16\linewidth]{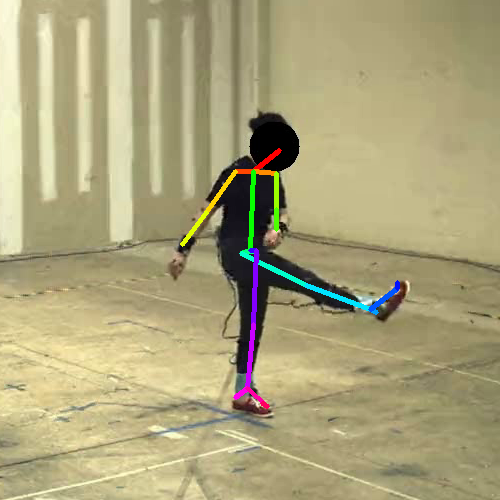} &
\includegraphics[width=0.16\linewidth]{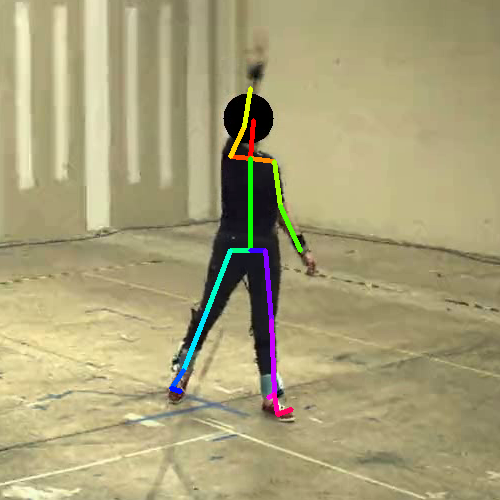} \\

\rotatebox{45}{\small{GT}} &
\includegraphics[width=0.16\linewidth]{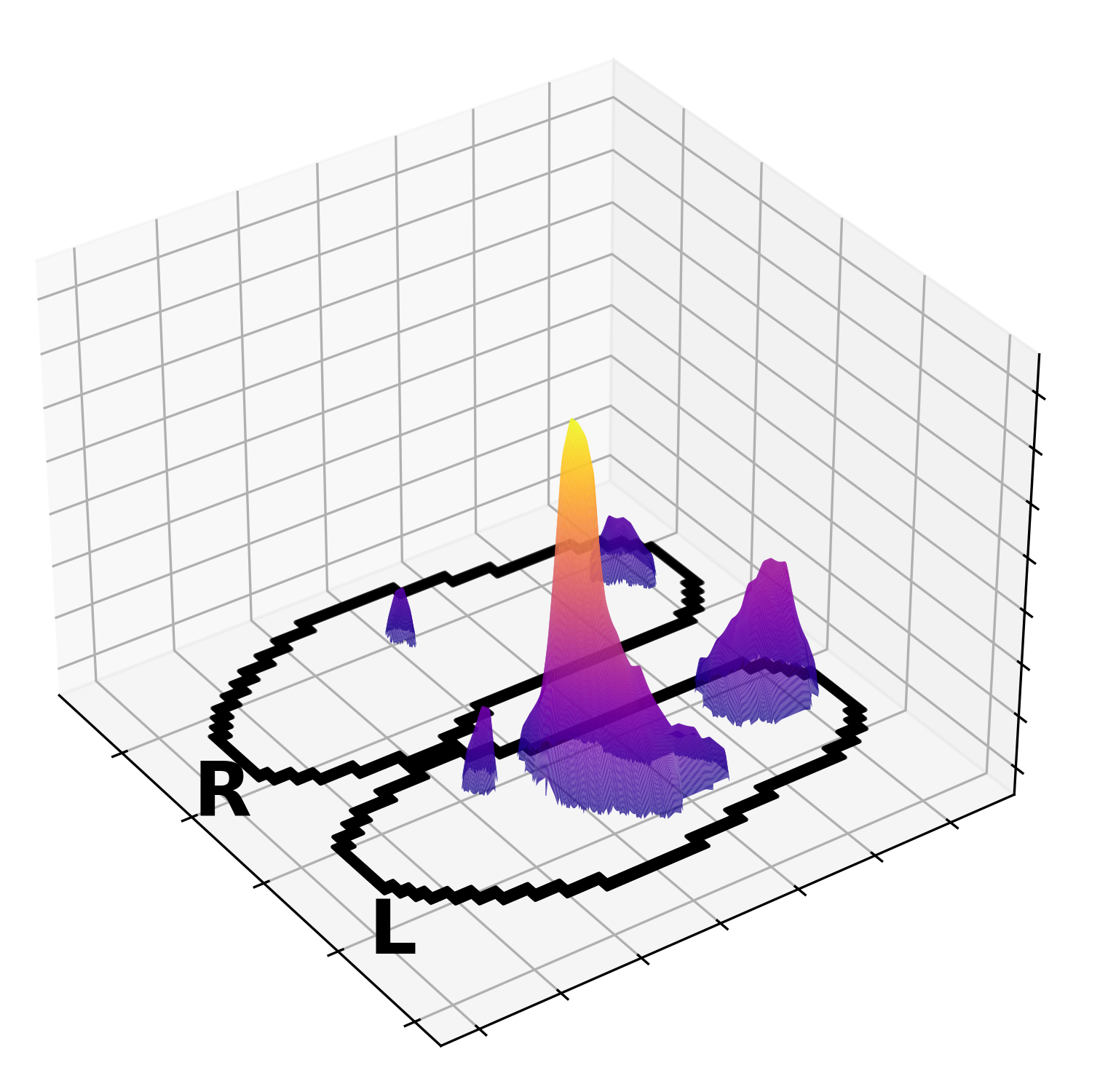} &
\includegraphics[width=0.16\linewidth]{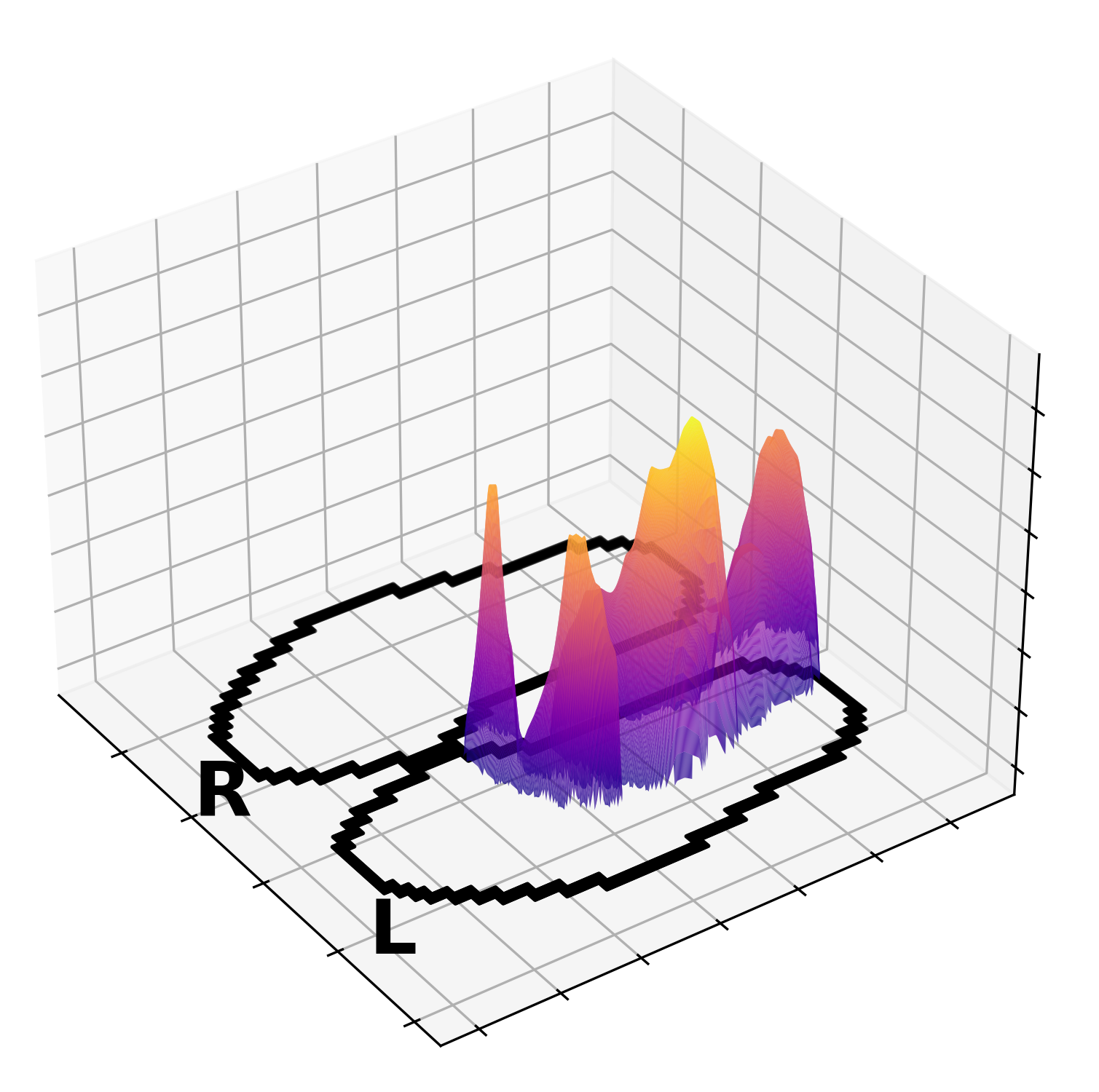} &
\includegraphics[width=0.16\linewidth]{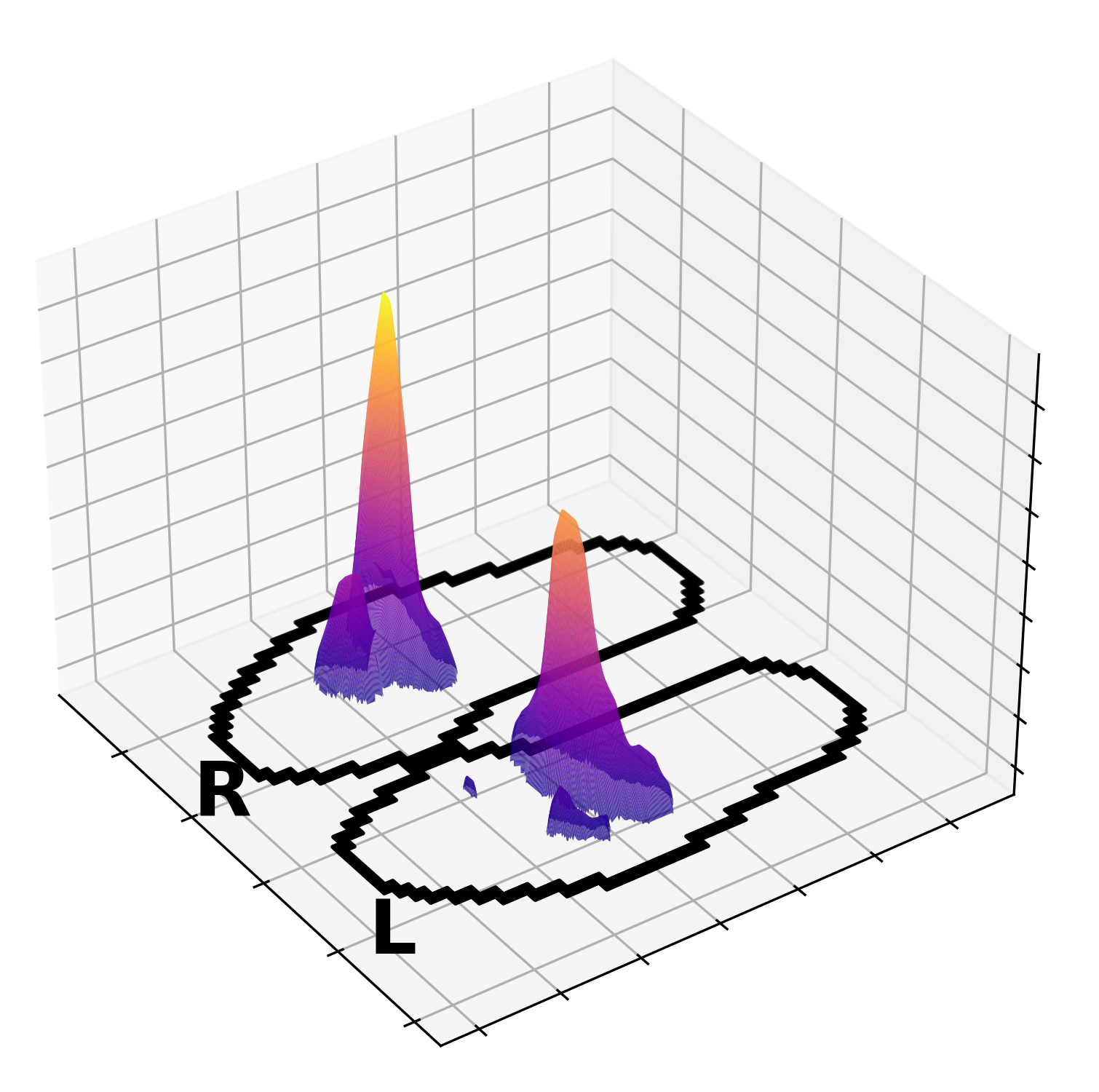} &
\includegraphics[width=0.16\linewidth]{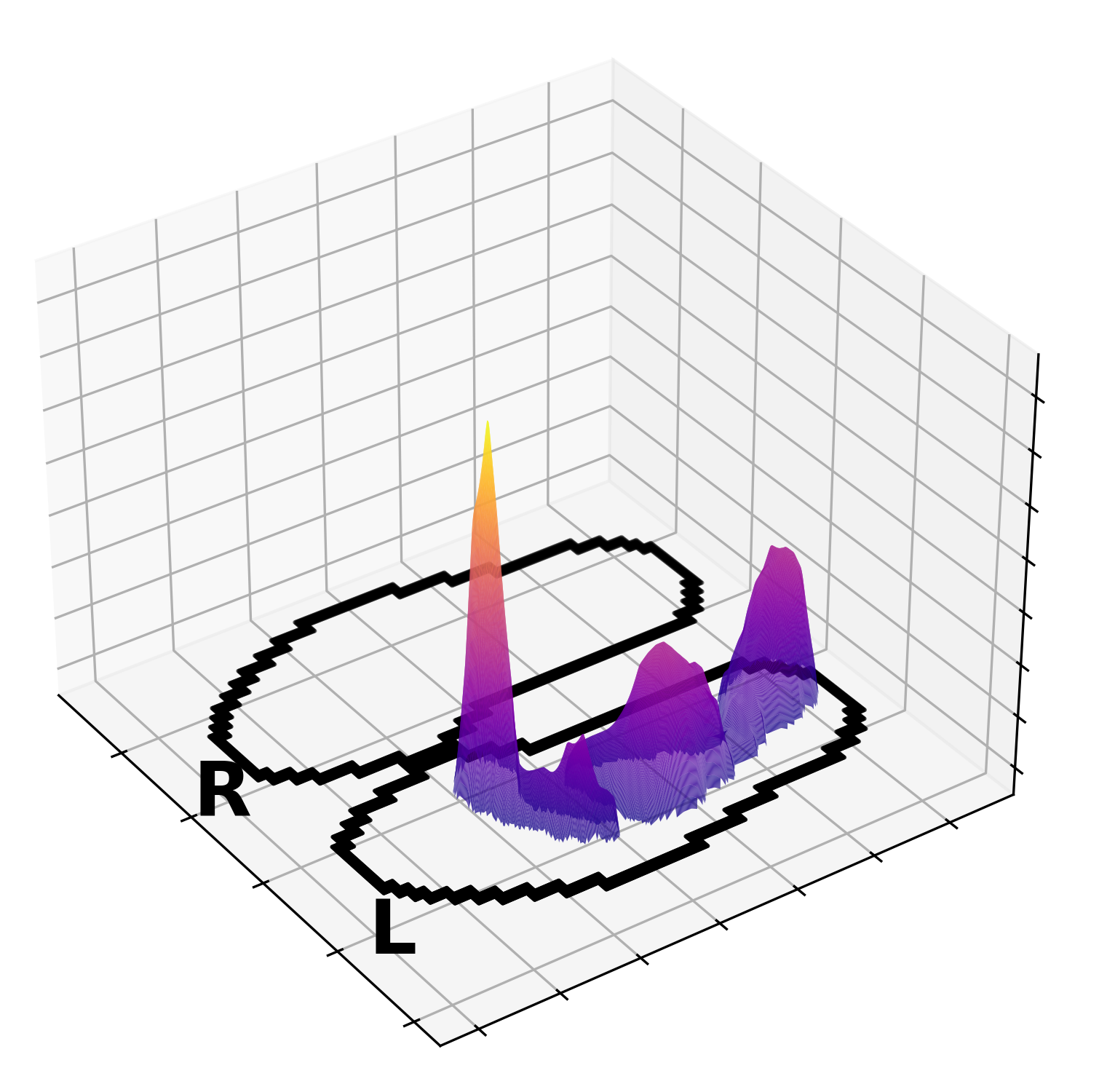} &
\includegraphics[width=0.16\linewidth]{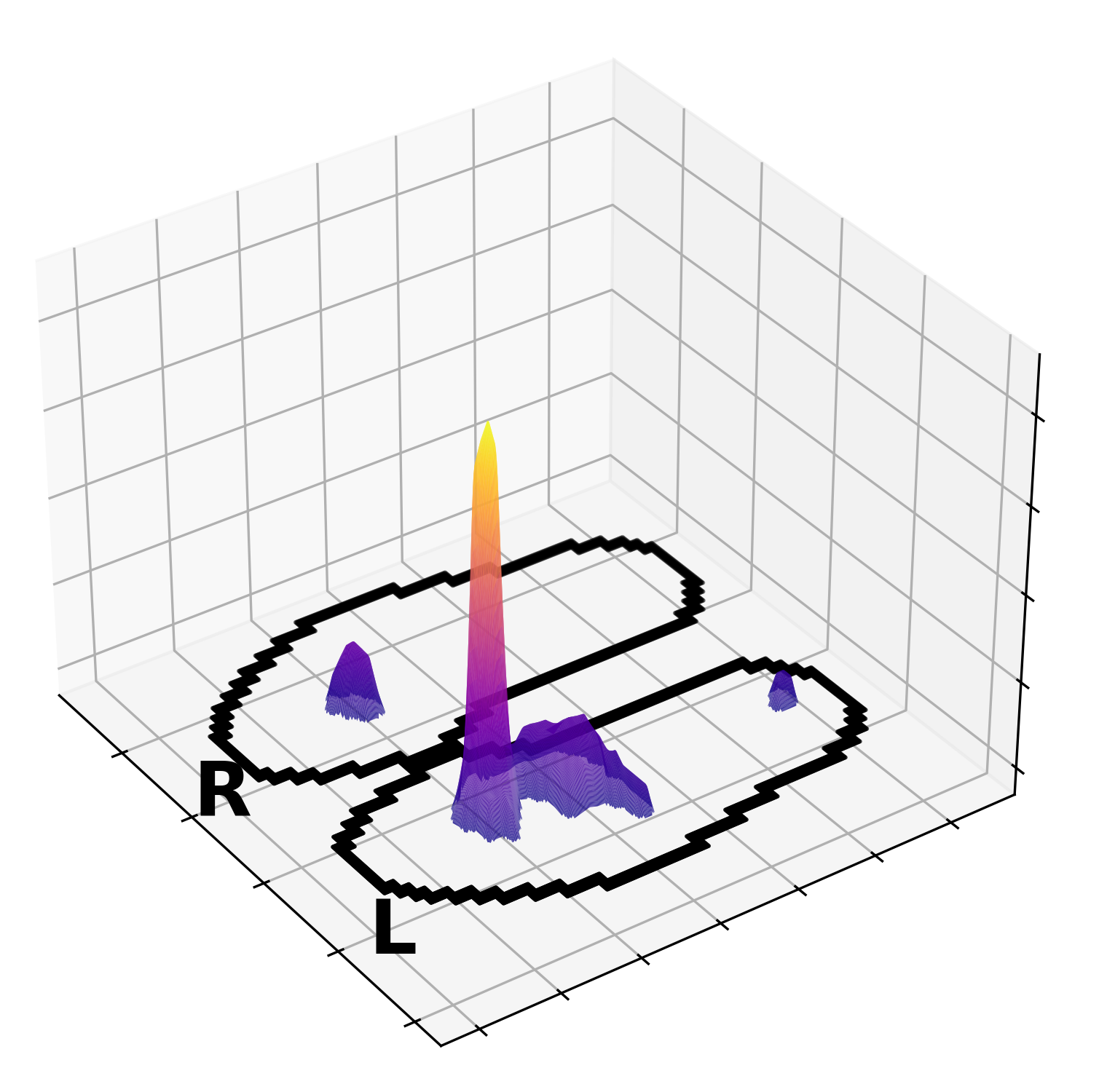} \\

\rotatebox{45}{\small{FootFormer}} &
\includegraphics[width=0.16\linewidth]{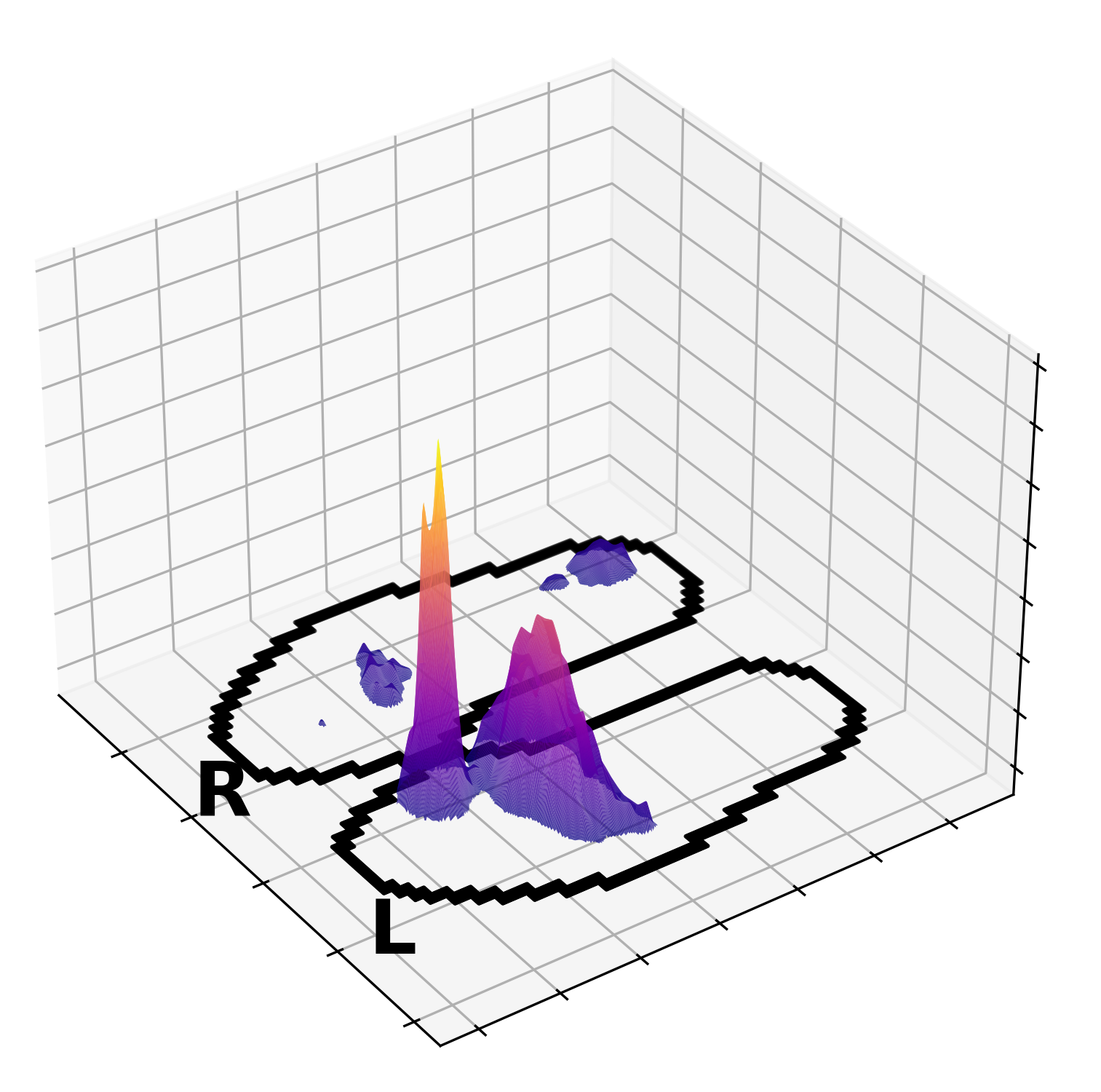} &
\includegraphics[width=0.16\linewidth]{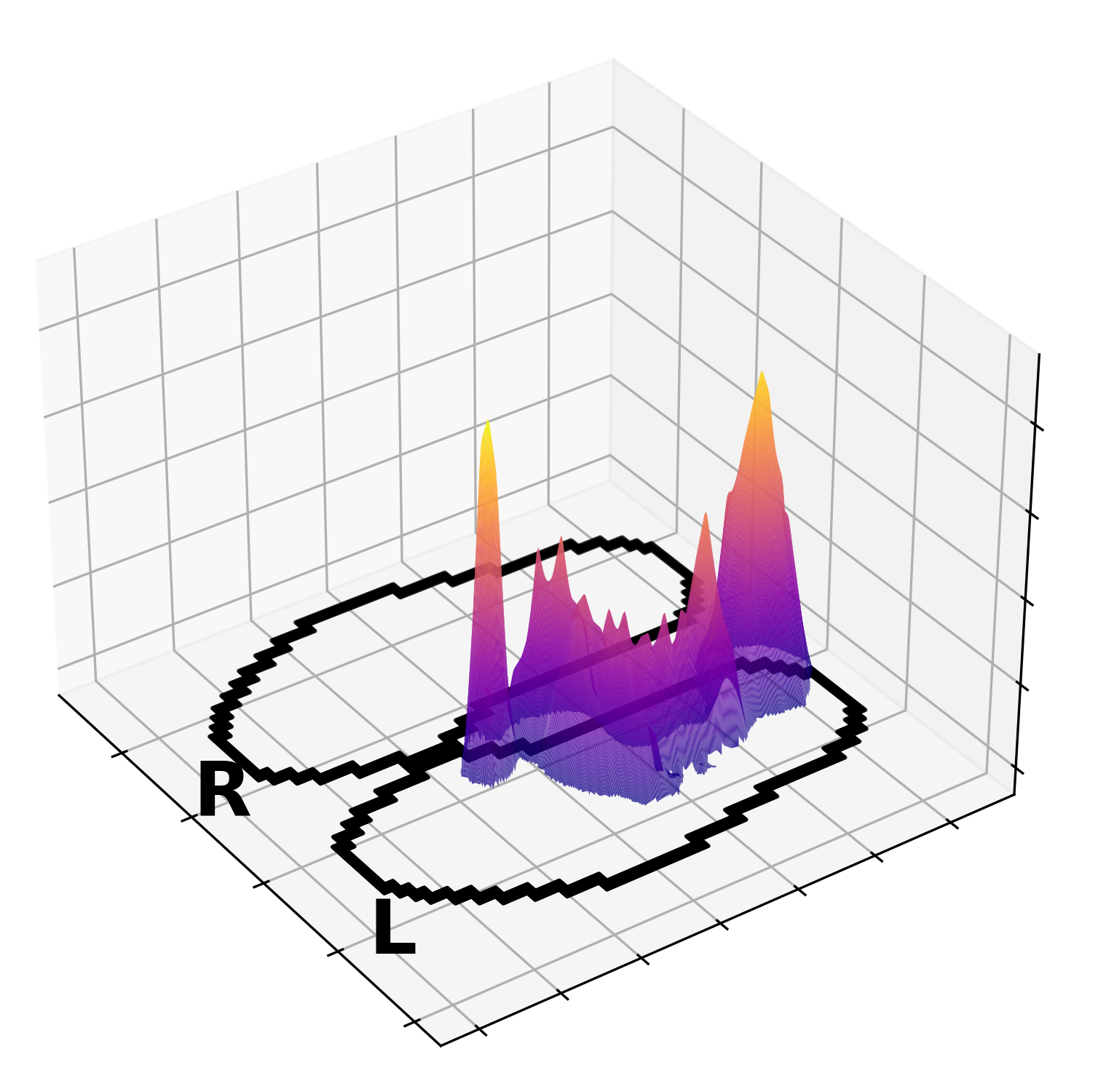} &
\includegraphics[width=0.16\linewidth]{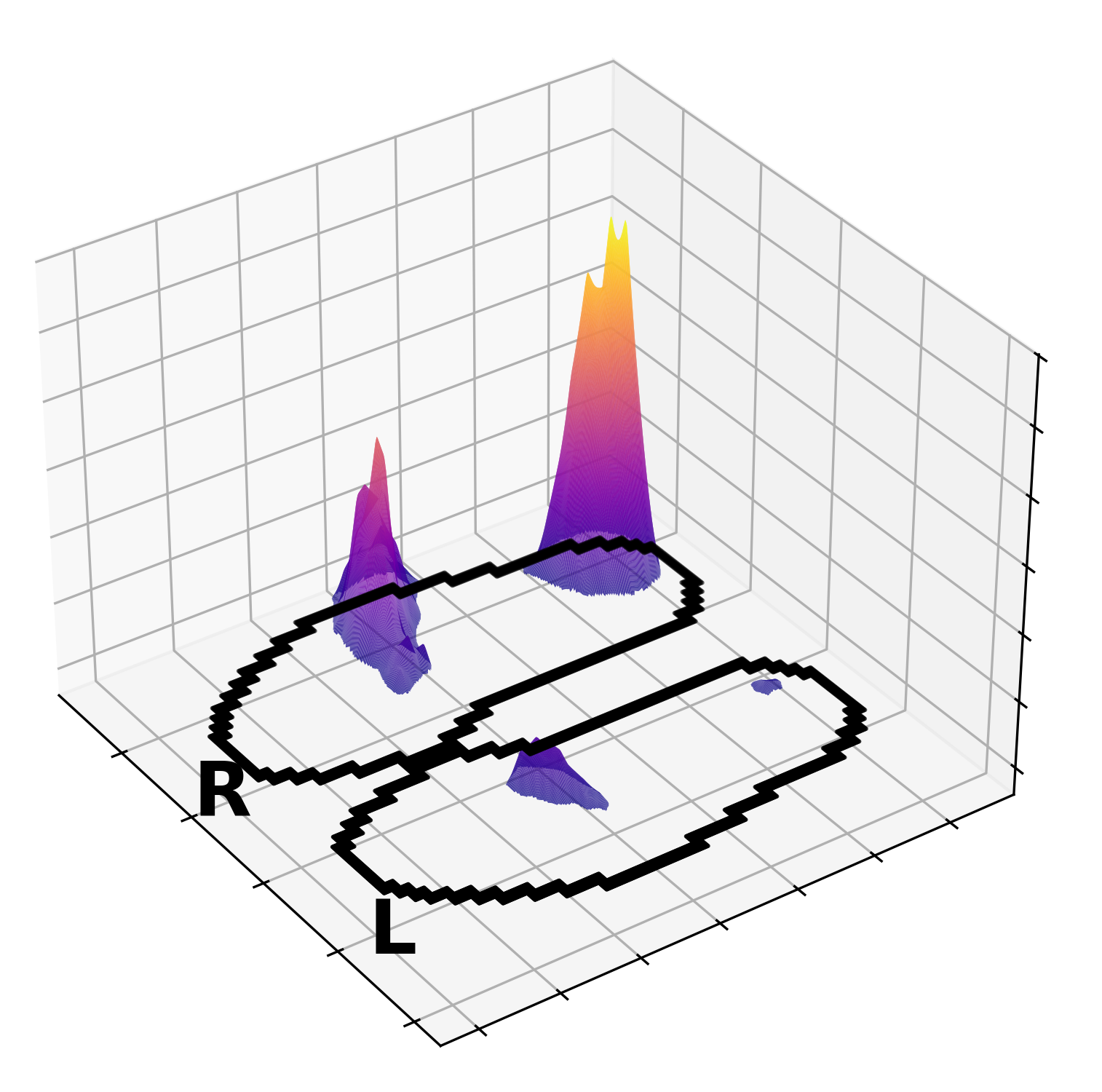} &
\includegraphics[width=0.16\linewidth]{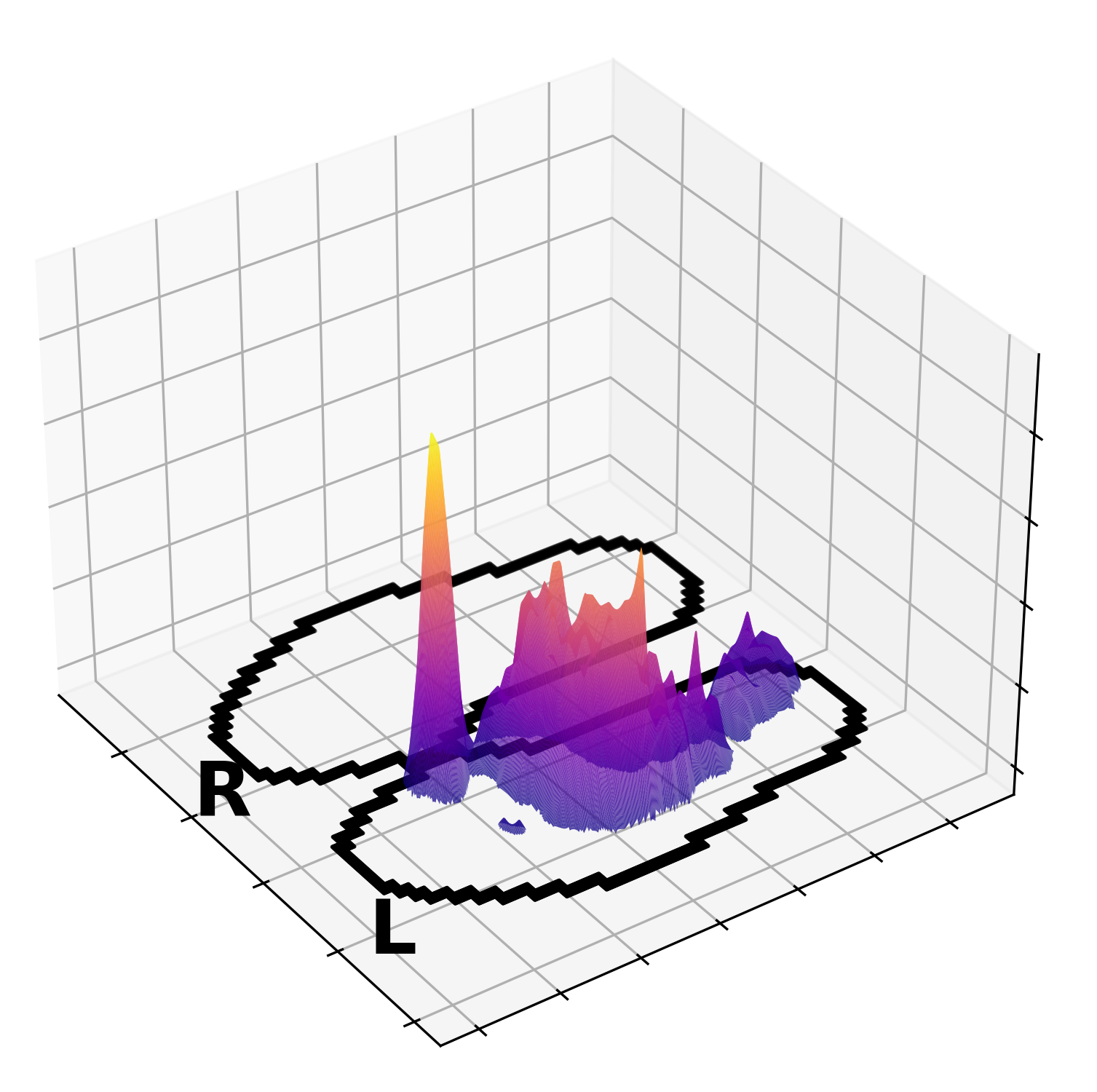} &
\includegraphics[width=0.16\linewidth]{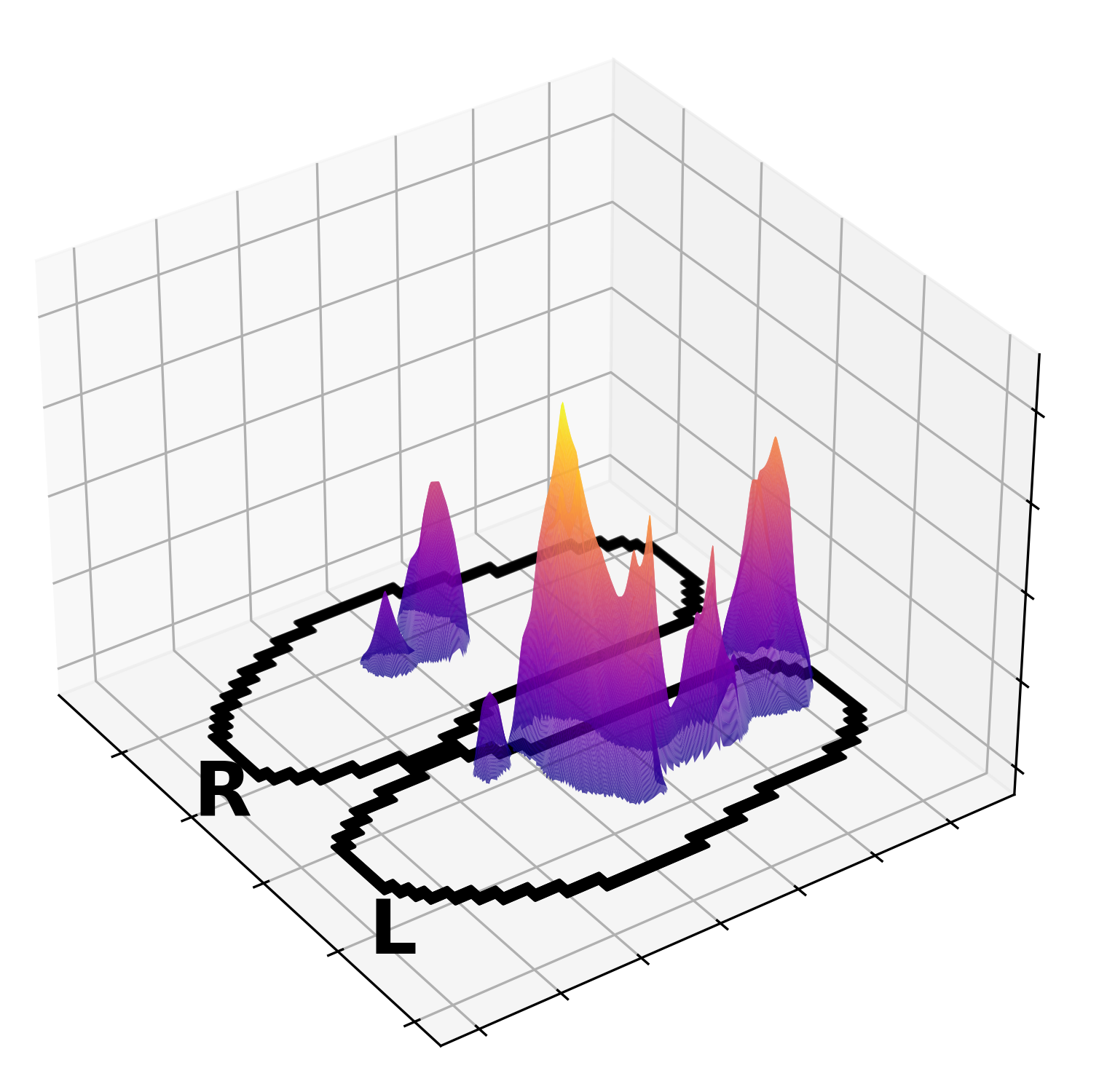} \\

\rotatebox{45}{\small{PNS}} &
\includegraphics[width=0.16\linewidth]{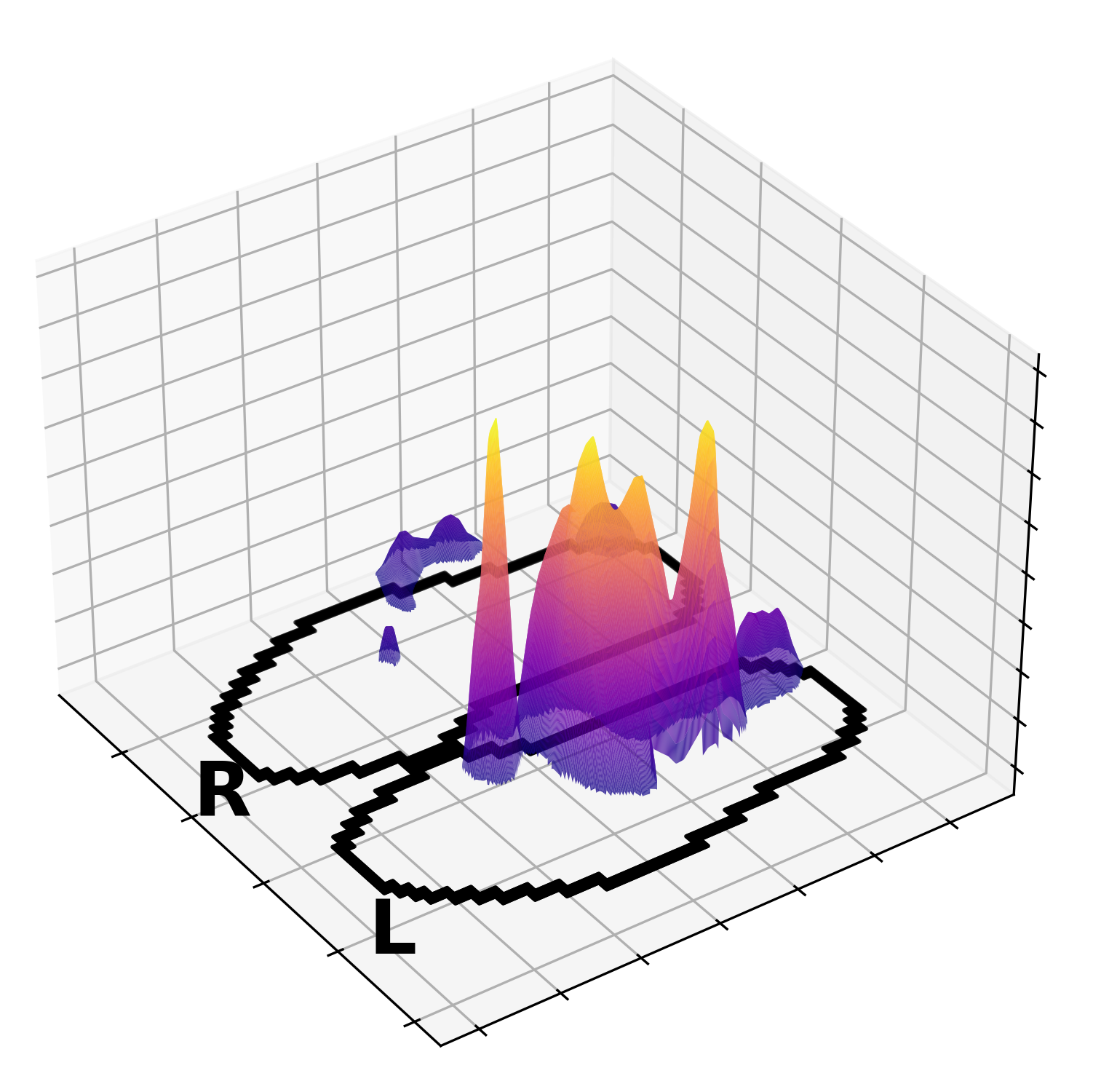} &
\includegraphics[width=0.16\linewidth]{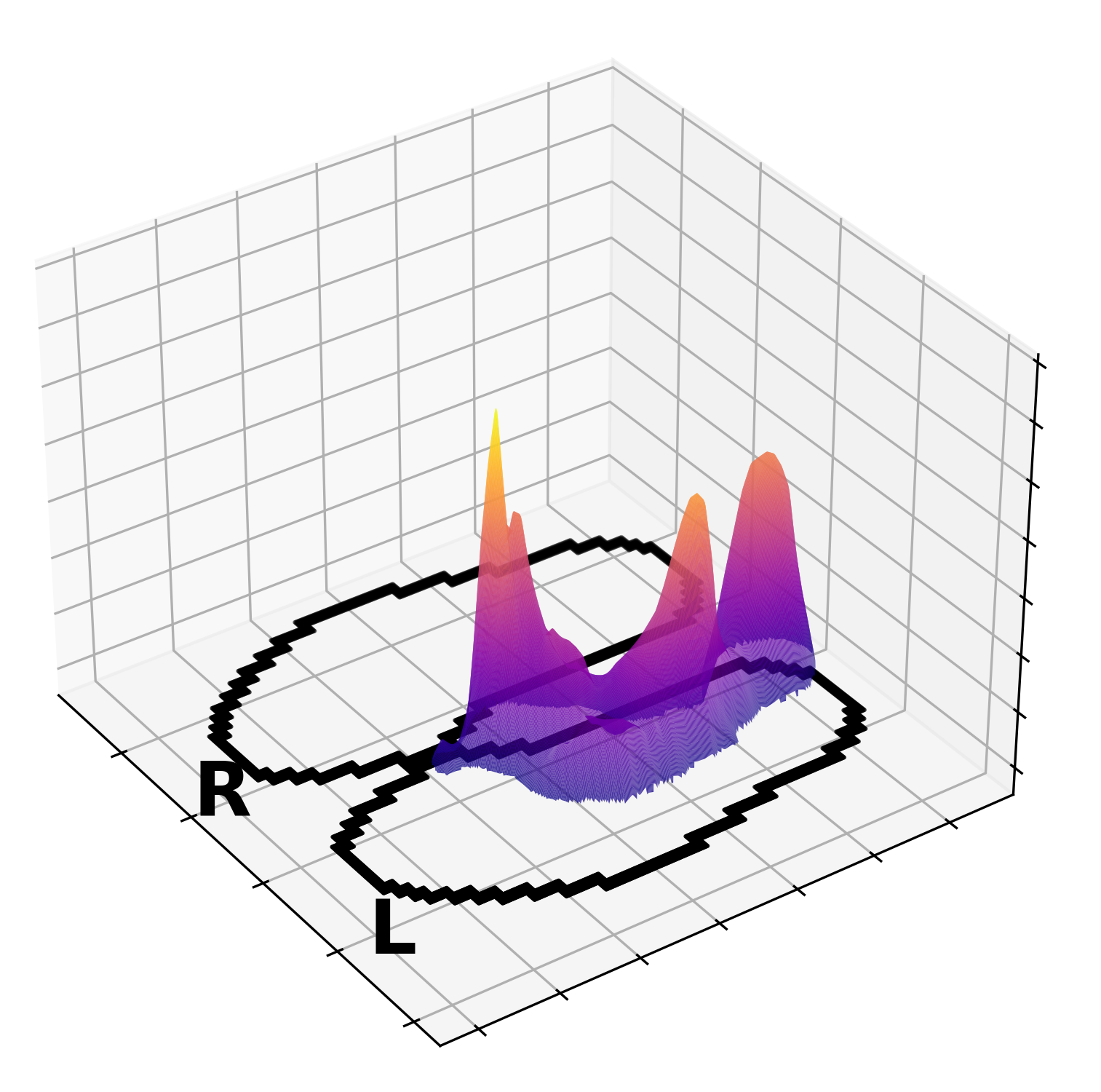} &
\includegraphics[width=0.16\linewidth]{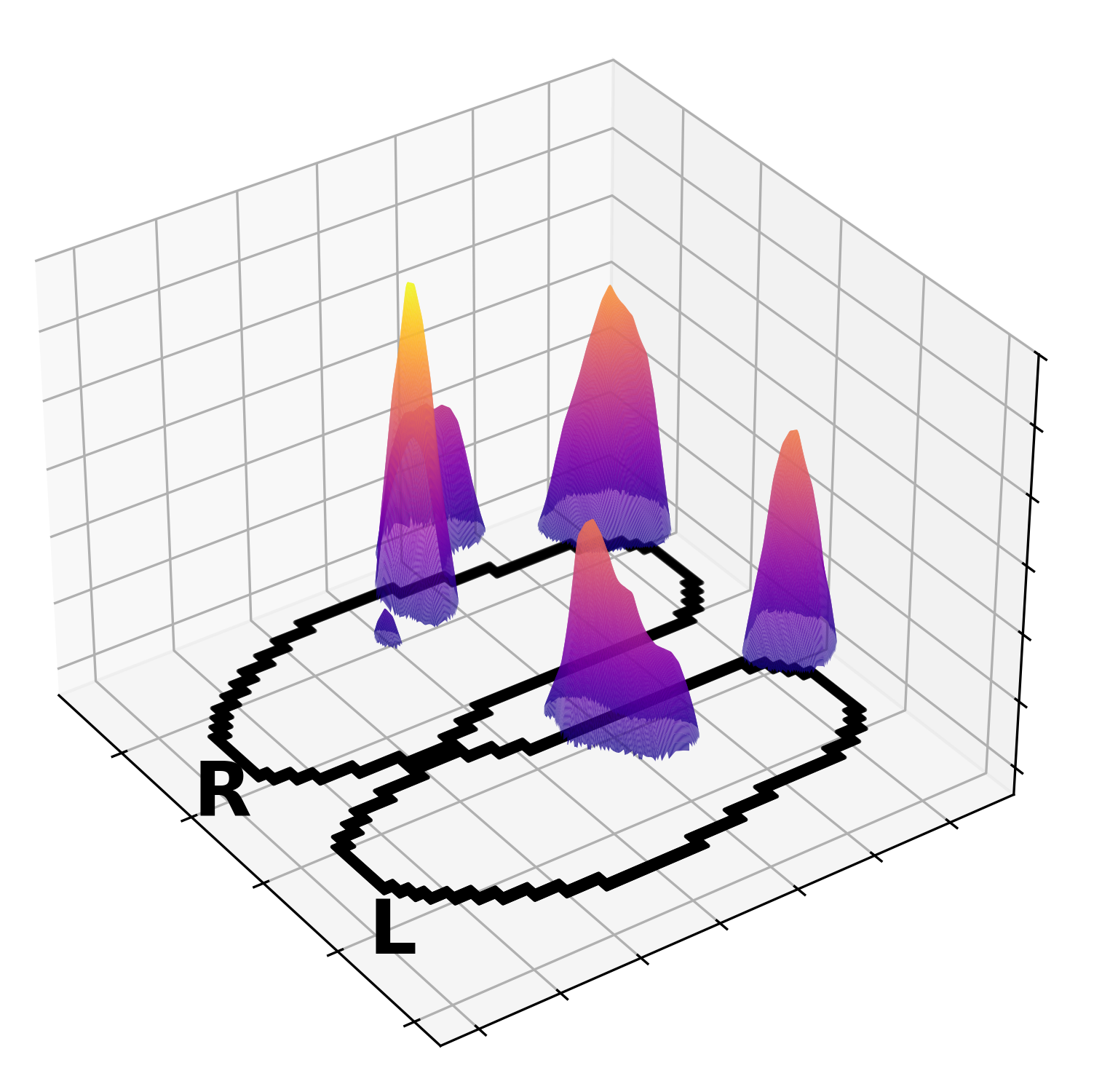} &
\includegraphics[width=0.16\linewidth]{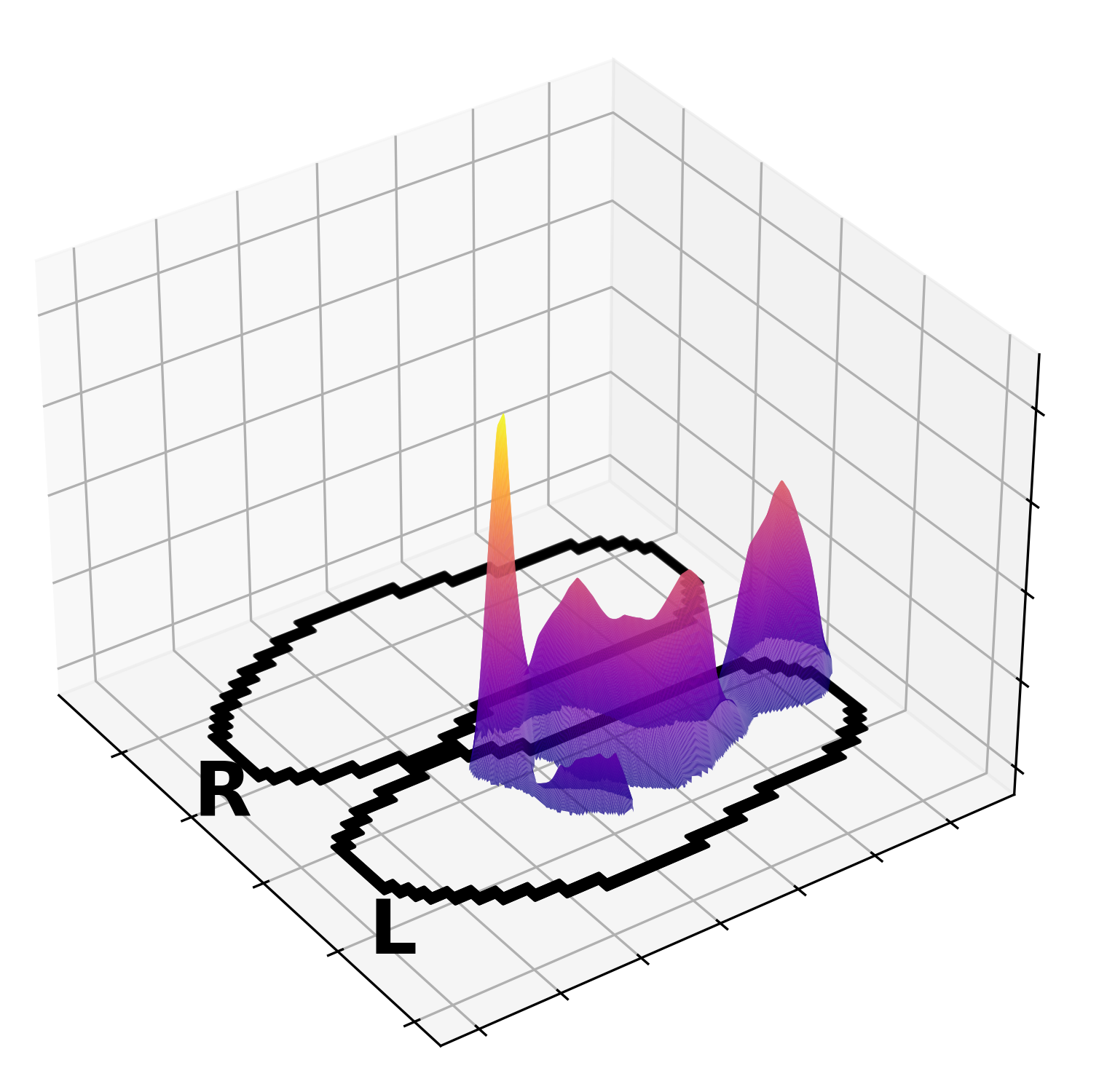} &
\includegraphics[width=0.16\linewidth]{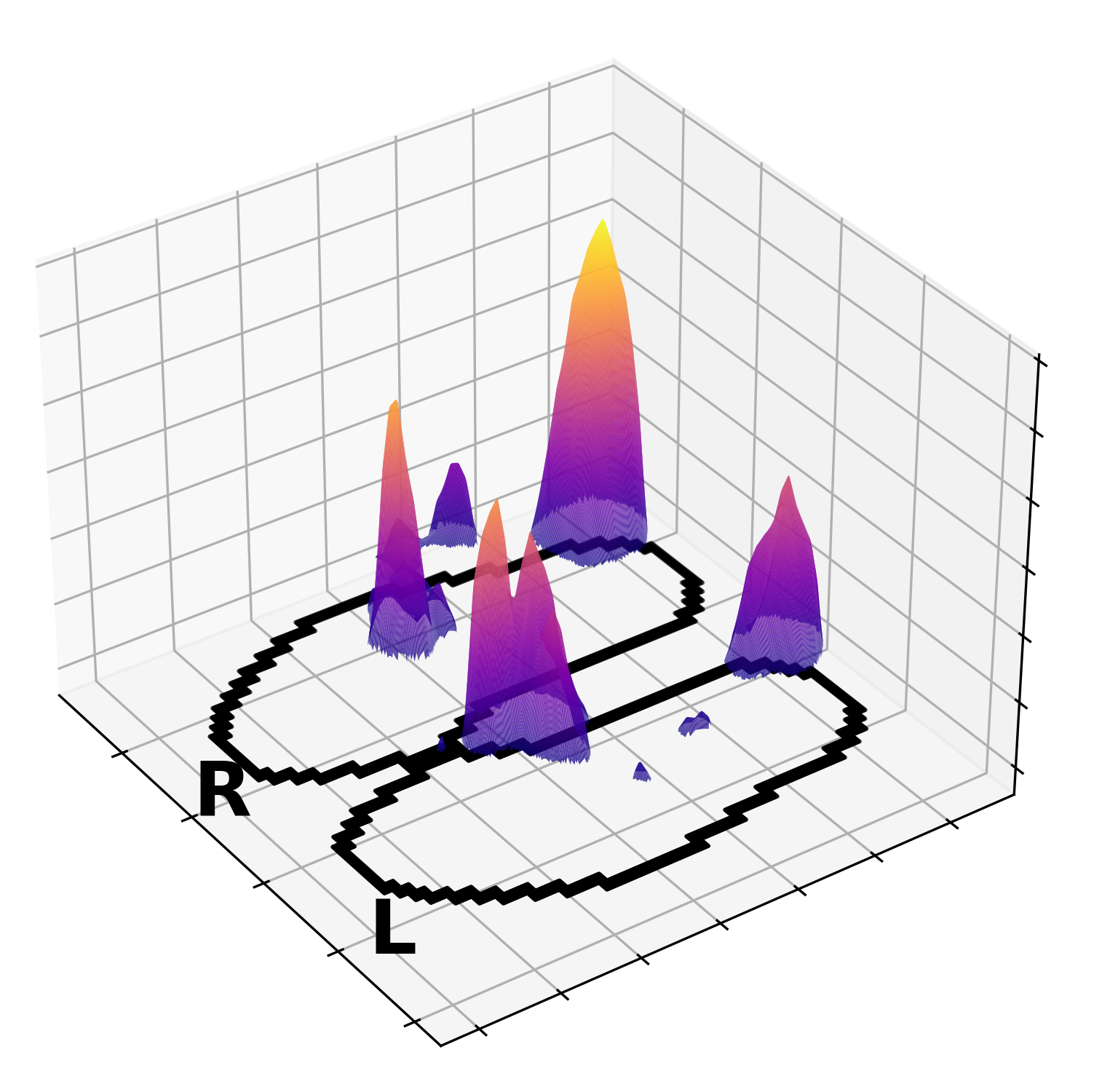} \\

\rotatebox{45}{\small{FPP-Net}} &
\includegraphics[width=0.16\linewidth]{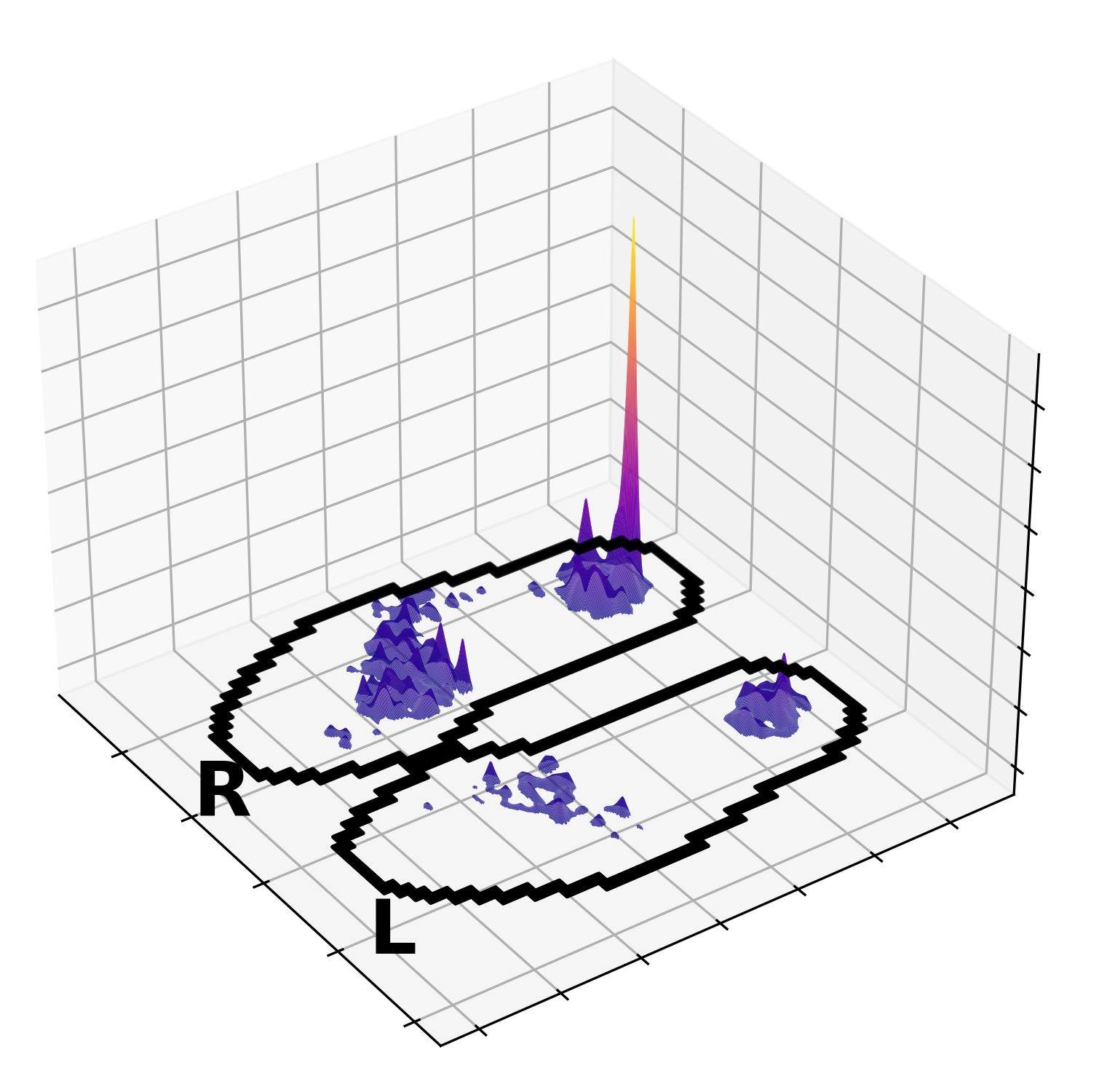} &
\includegraphics[width=0.16\linewidth]{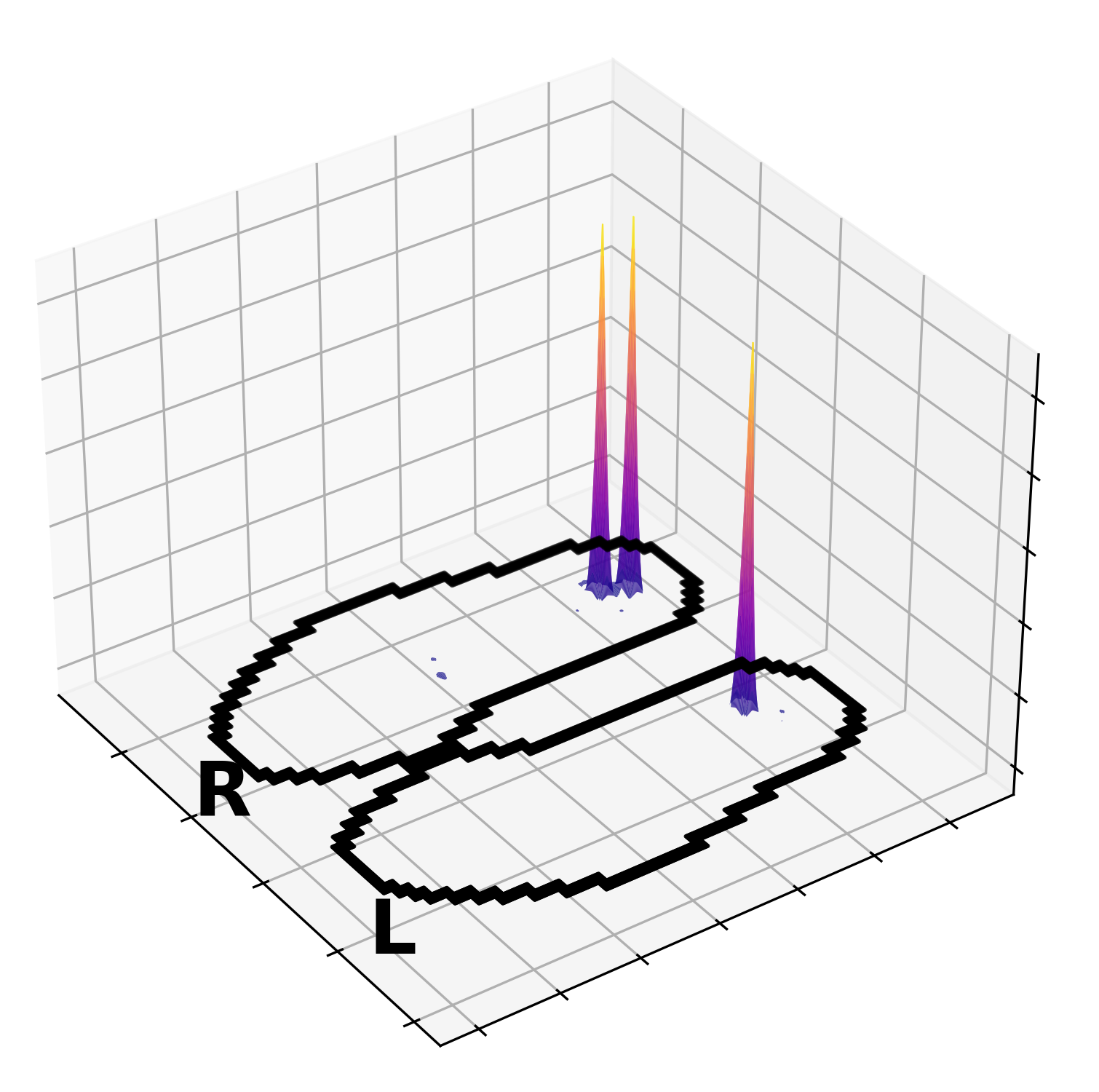} &
\includegraphics[width=0.16\linewidth]{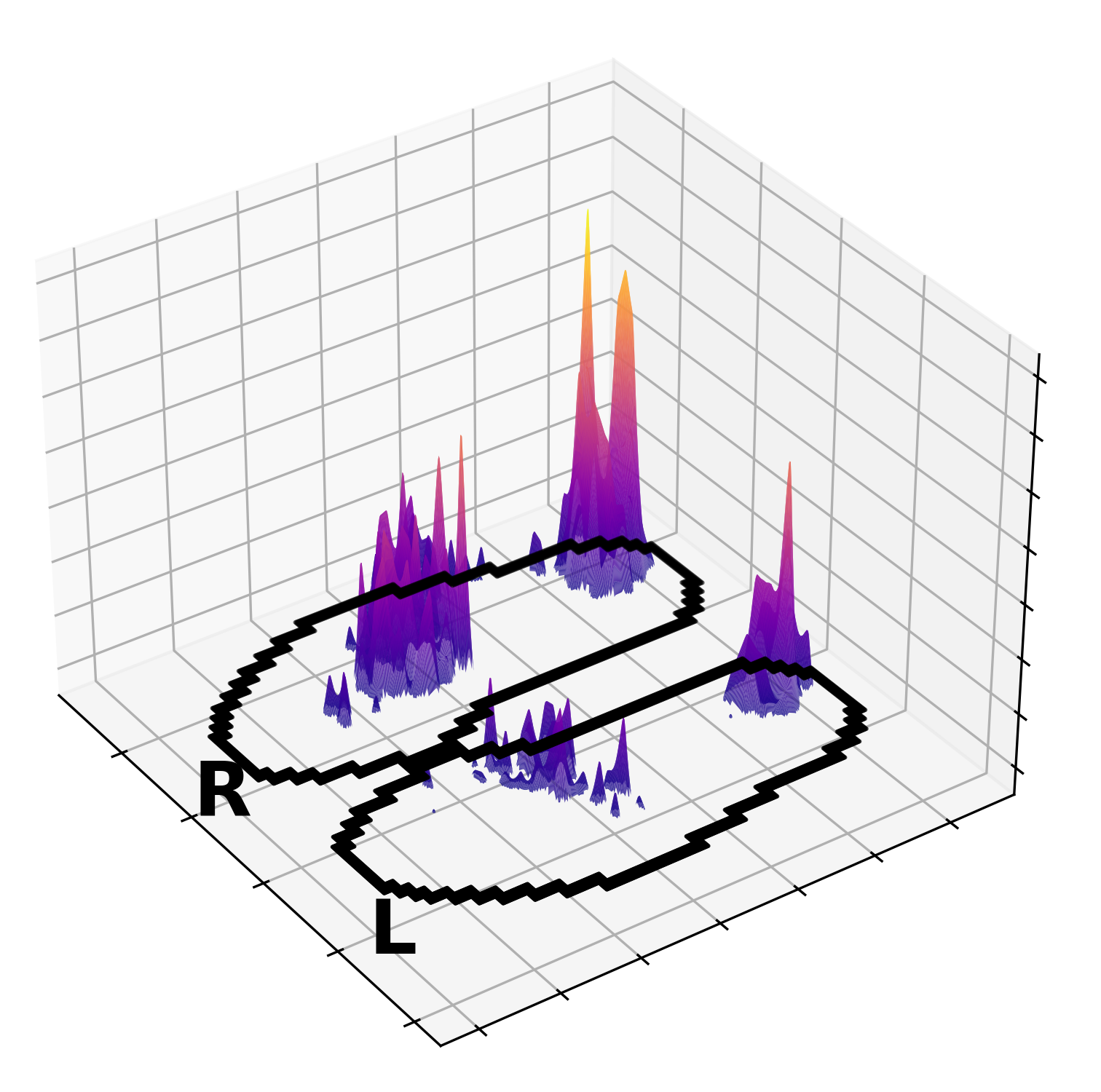} &
\includegraphics[width=0.16\linewidth]{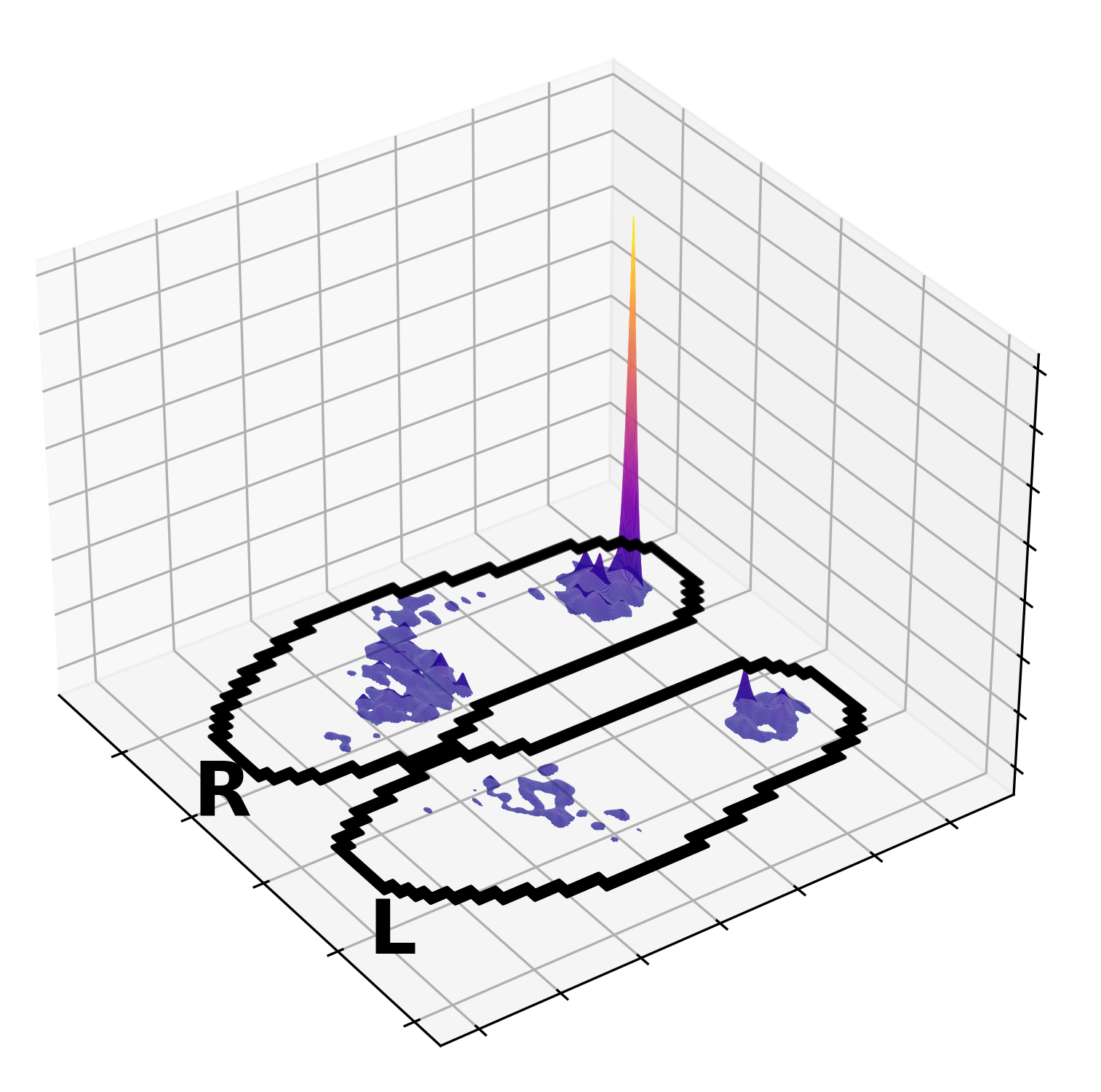} &
\includegraphics[width=0.16\linewidth]{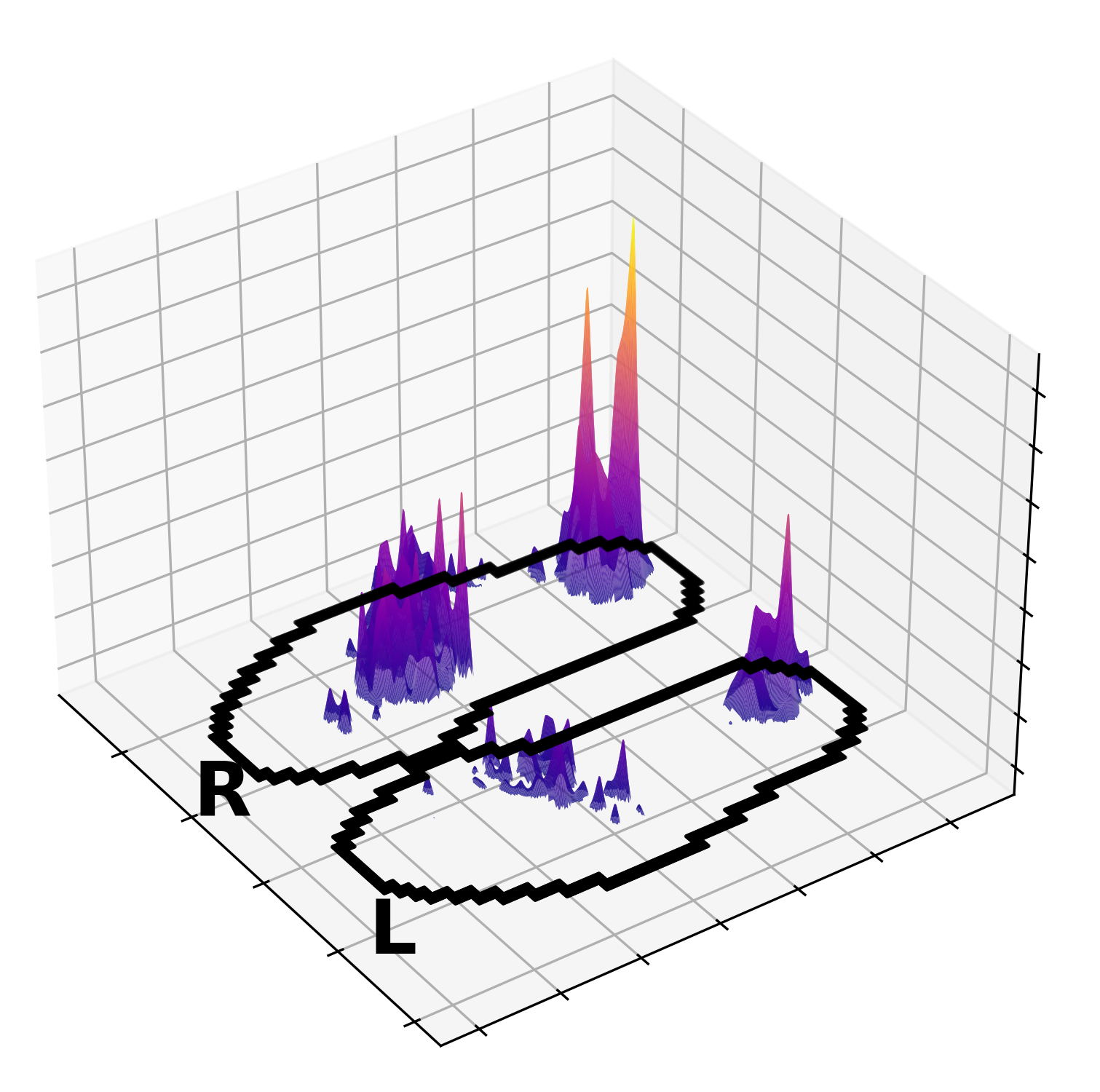} \\

\rotatebox{45}{\small{UP}} &
\includegraphics[width=0.16\linewidth]{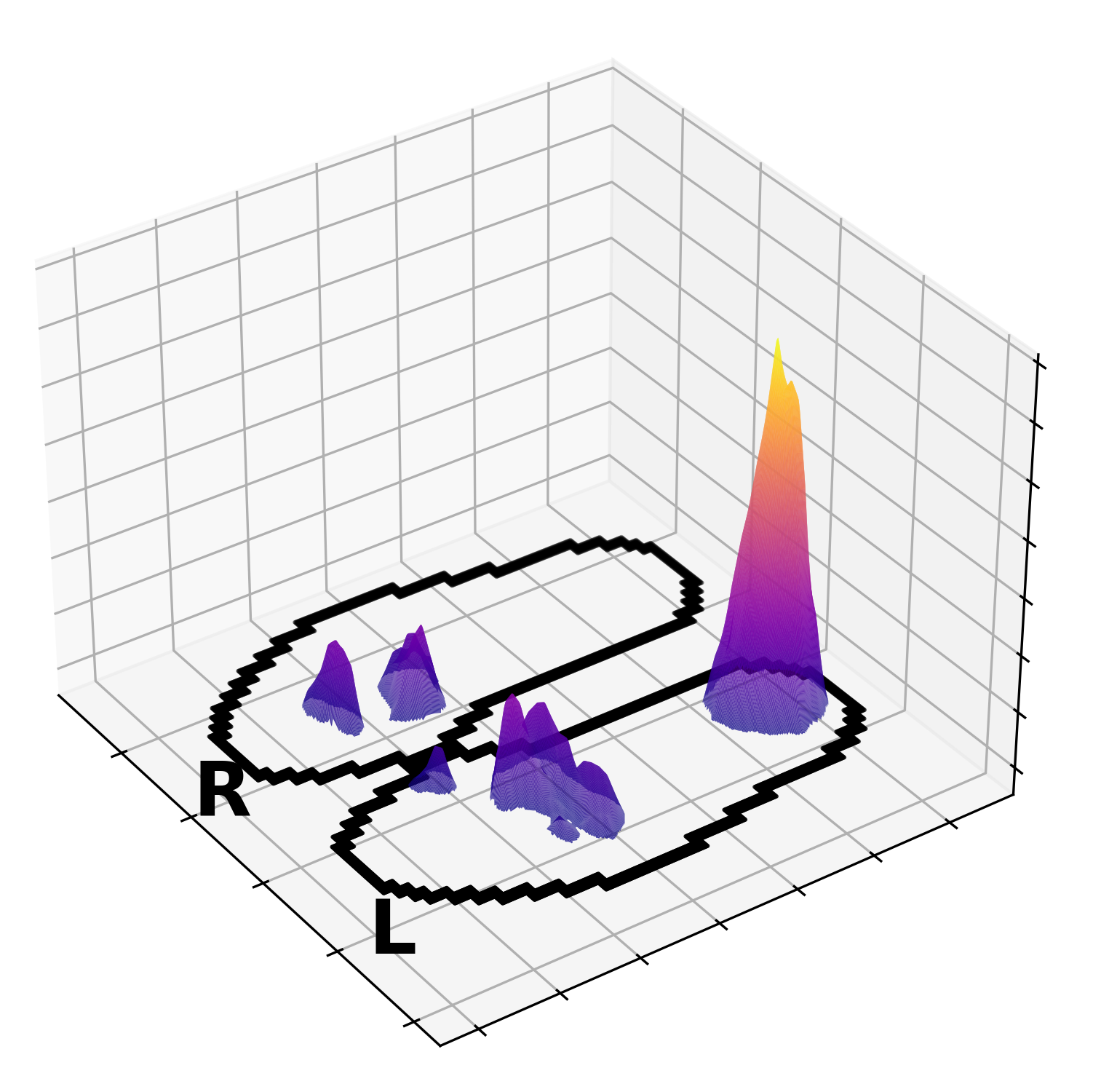} &
\includegraphics[width=0.16\linewidth]{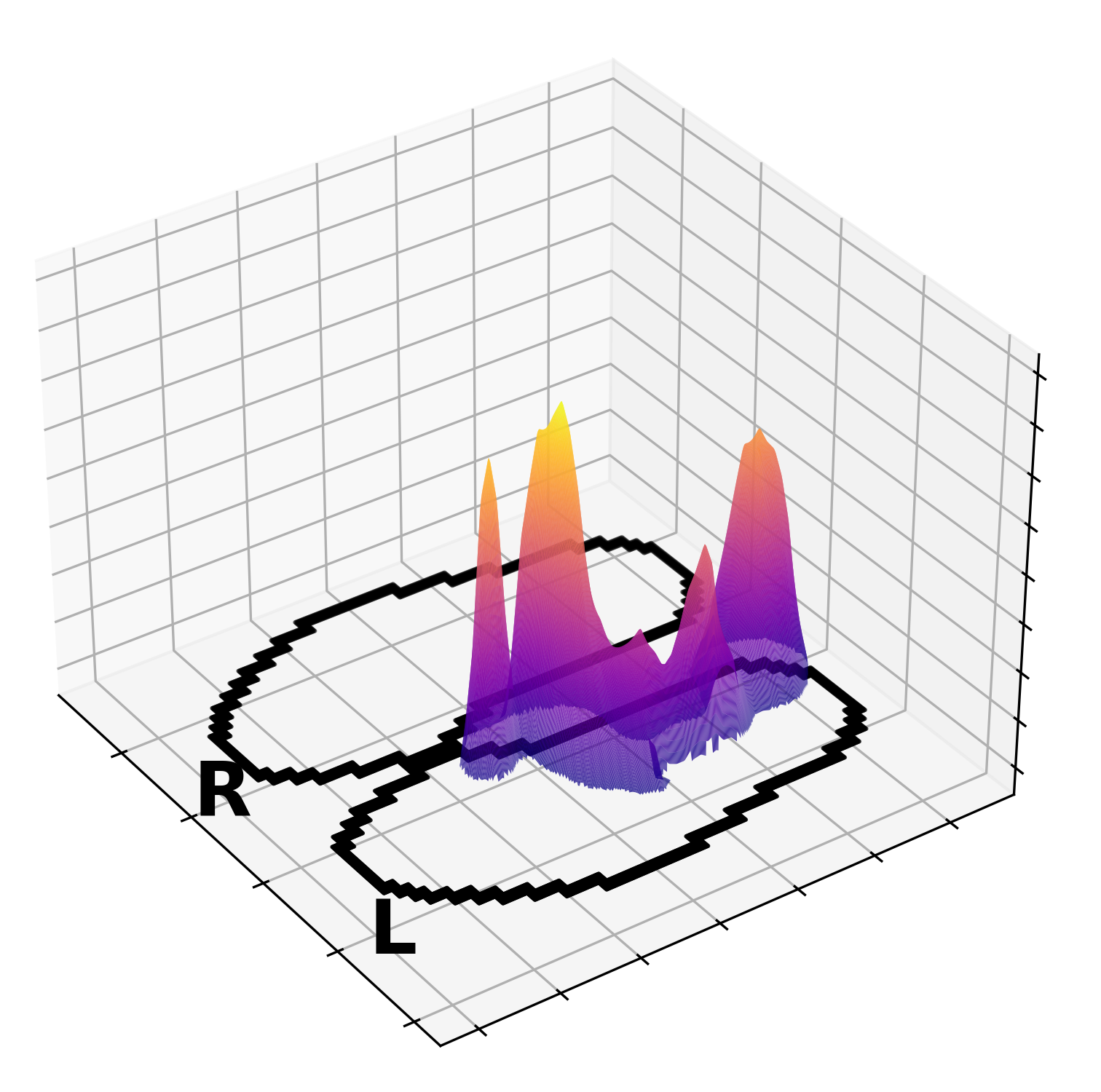} &
\includegraphics[width=0.16\linewidth]{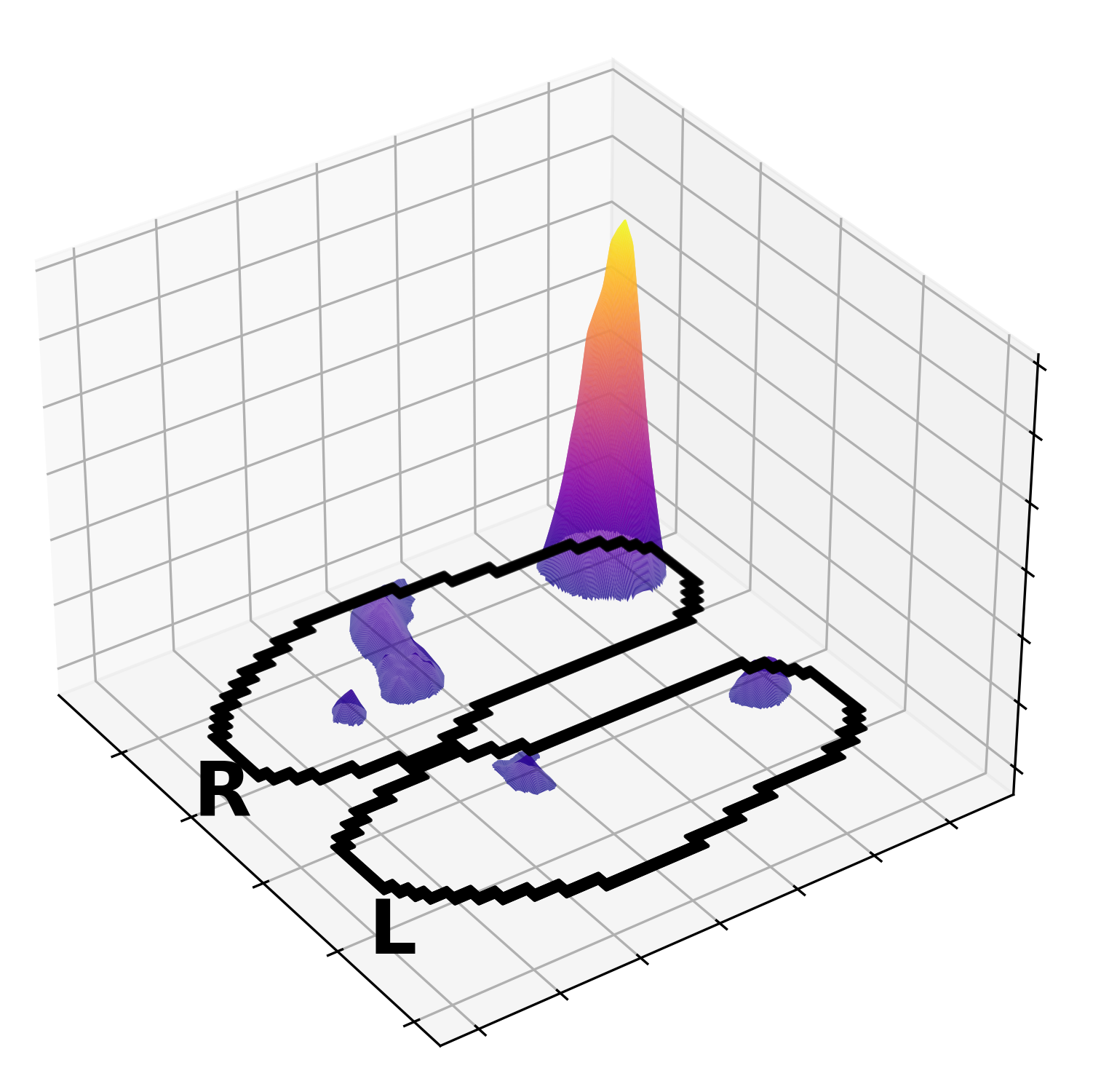} &
\includegraphics[width=0.16\linewidth]{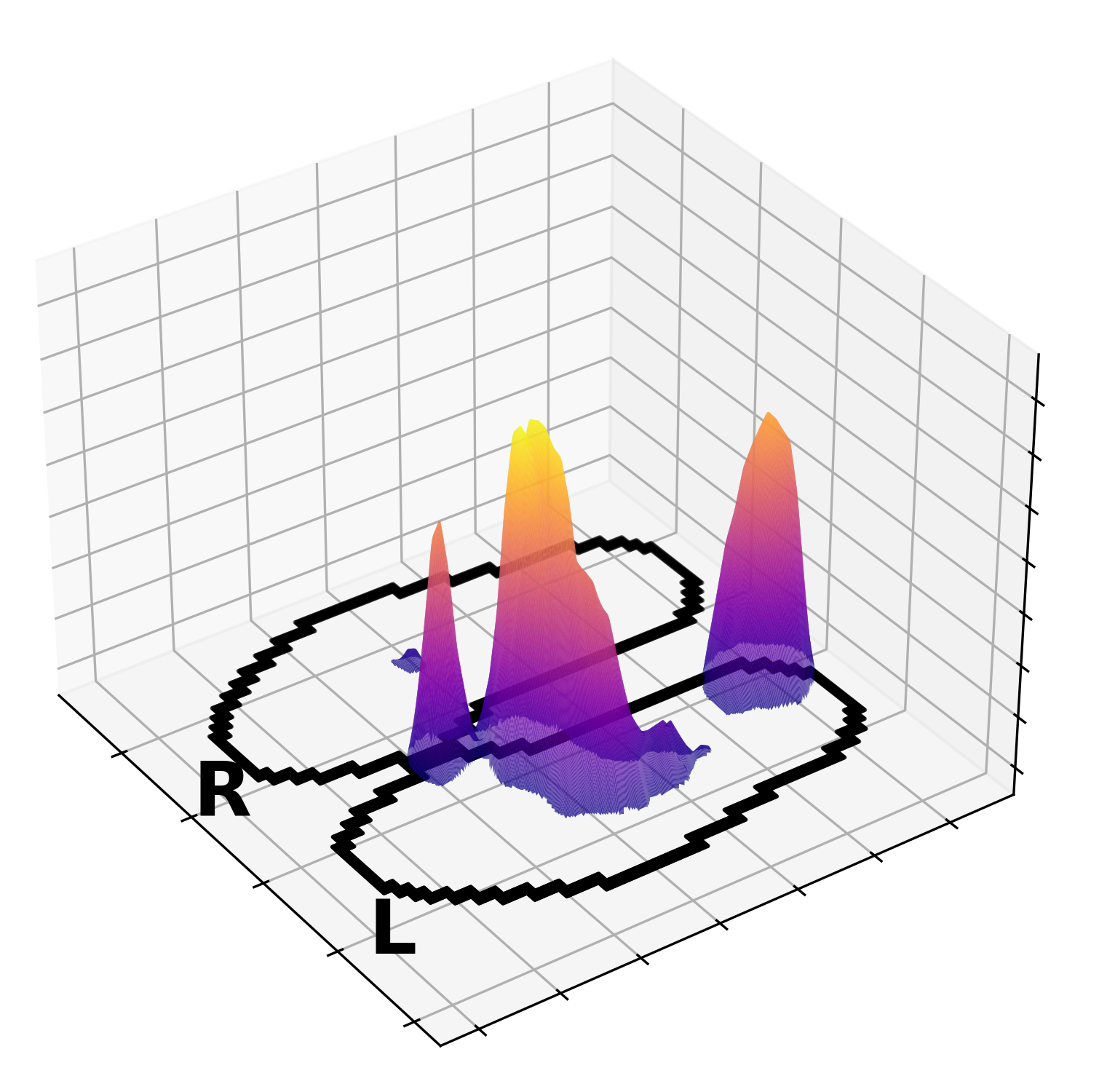} &
\includegraphics[width=0.16\linewidth]{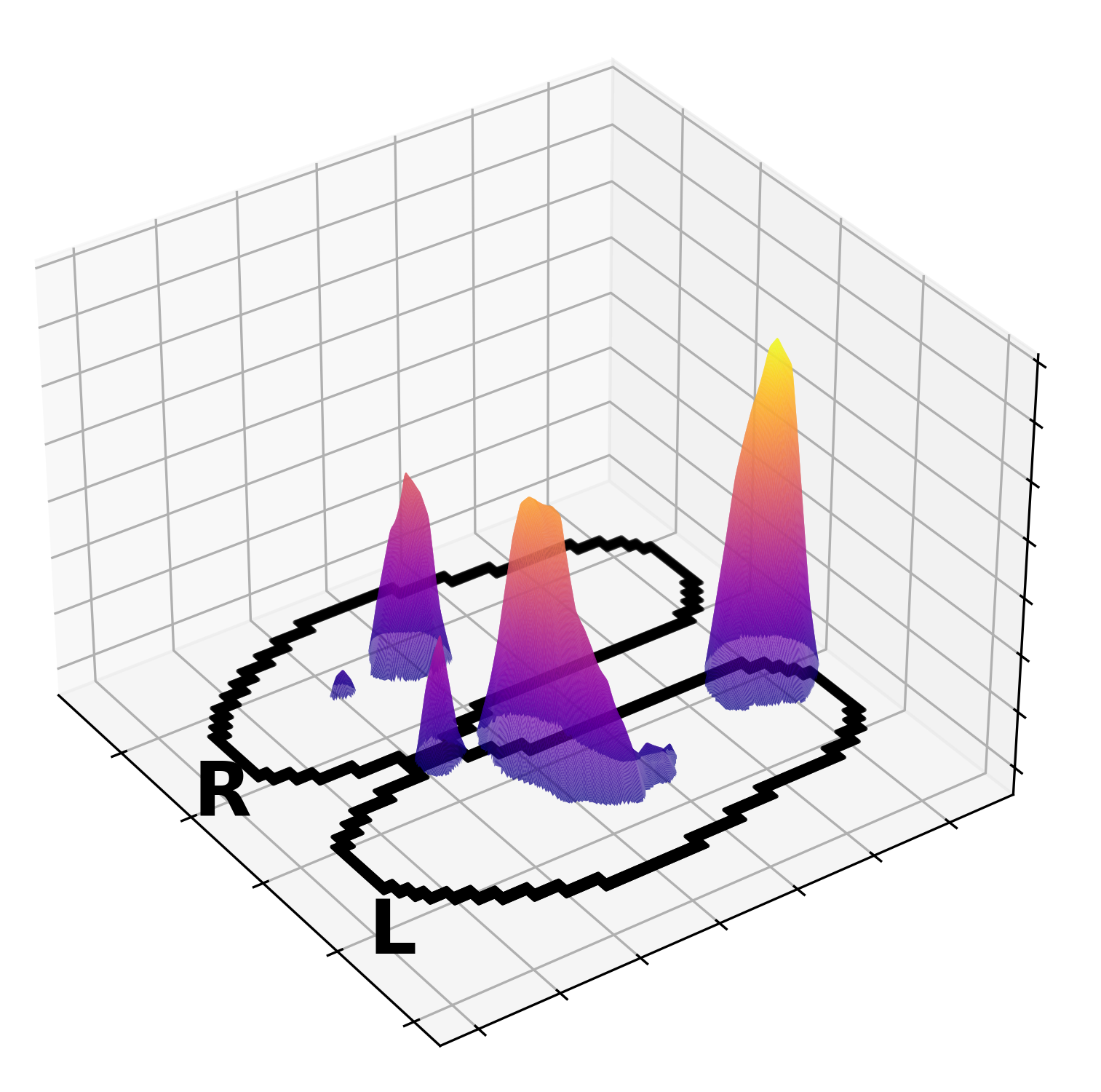} \\
\end{tabular}
\vspace{0.3cm}
\caption{Qualitative comparison of predicted foot pressure maps across five OM activities for FootFormer (Ours), PNS~\cite{ScottECCV2020}, FPP-Net~\cite{mmvp}, and UP~\cite{UnderPressure}. Column headers indicate action; row headers indicate source view or prediction method.}
\label{fig:quals}
\end{figure}

We provide qualitative examples of the different camera views and synchronized pressure along with predictions from FootFormer and the baseline models in Figure~\ref{fig:quals}.

\subsection{Additional Ordinary Movements Evaluation}

\begin{figure}[h!]
    \centering
    \includegraphics[width=0.95\linewidth]{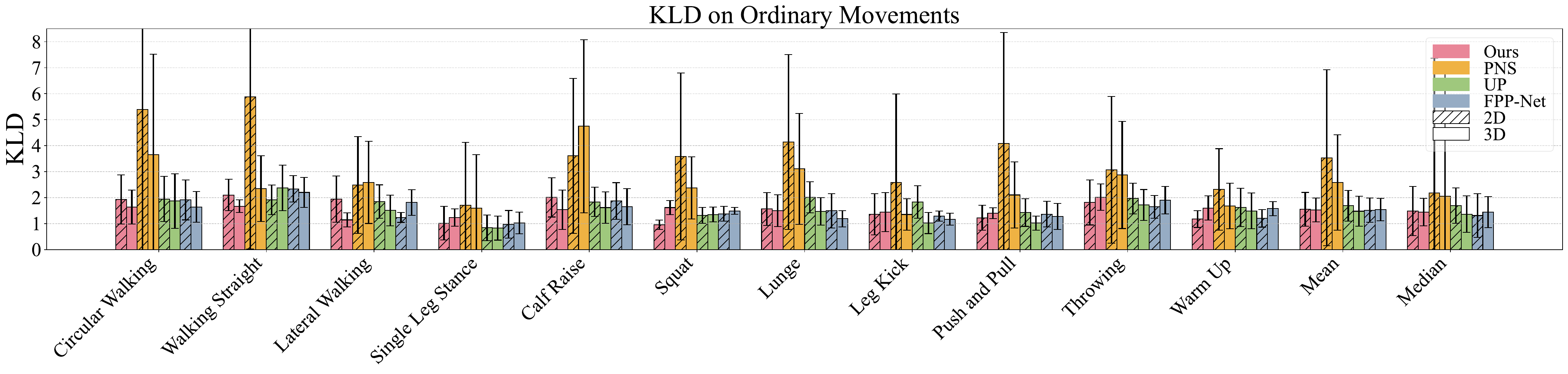}
    \vspace{0.3cm}
    \caption{KLD foot pressure estimation results on the Ordinary Movements for each movement with mean and median values. Lower is better. Statistical significance values are presented in Table~\ref{tab:pval_OM}.}
    \label{fig:OM_pressure_metrics}
\end{figure}

``Ordinary" movements (OMs), performed by a participant in the original Taiji dataset, were composed of commonplace motions and exercises (walking, squats, lunges, etc.). To evaluate how well our model can generalize to non-Taij movement, we perform training on the \PSUTMM dataset, and test on the unseen OMs. Importantly, \textbf{we use the model trained on the left-out subject}, meaning the model had not been trained with data including the performer. We report per-movement and across all movement results in Figure~\ref{fig:OM_pressure_metrics}. Table~\ref{tab:pval_OM} reports statistical significance values of paired \textit{t}-tests comparing FootFormer with the baseline models.

\begin{table}[h!]
\centering
\small
\resizebox{0.95\linewidth}{!}{%
\begin{tabular}{lcc}
\toprule
\textbf{Method} 
& \textbf{OM 2D KLD $\downarrow$ (\textit{p}-value vs Ours)} 
& \textbf{OM 3D KLD $\downarrow$ (\textit{p}-value vs Ours)} \\
\midrule
PNS~\cite{ScottECCV2020}      
& 3.53 $\pm$ 1.23$^{\dagger}$ {\scriptsize($1.65\mathrm{e}{-04}$)} 
& 2.59 $\pm$ 0.94$^{\dagger}$ {\scriptsize($4.35\mathrm{e}{-03}$)} \\
FPP-Net~\cite{mmvp}           
& \textbf{1.52 $\pm$ 0.37} {\scriptsize($6.89\mathrm{e}{-01}$)} 
& 1.54 $\pm$ 0.34 {\scriptsize($9.04\mathrm{e}{-01}$)} \\
UP~\cite{UnderPressure}       
& 1.69 $\pm$ 0.34 {\scriptsize($1.28\mathrm{e}{-01}$)} 
& \textbf{1.48 $\pm$ 0.41} {\scriptsize($6.66\mathrm{e}{-01}$)} \\
Ours                 
& 1.56 $\pm$ 0.40 & 1.53 $\pm$ 0.22 \\
\bottomrule
\end{tabular}
}
\vspace{0.3cm}
\caption{Comparison of FootFormer (Ours) with baselines on Ordinary Movements (OM). 
Each entry shows mean $\pm$ std KLD and the paired \textit{t}-test \textit{p}-value vs.~Ours.
\textbf{Bold} indicates the best (lowest KLD);
$^{\dagger}$ denotes a statistically significant difference from Ours (\textit{p}~<~0.05).}
\label{tab:pval_OM}
\end{table}

\section{Additional Foot Contact Estimation Results}
Table~\ref{tab:contact_pval} reports statistical significance values of paired \textit{t}-tests comparing FootFormer with FPP-Net and UP on the MMVP~\cite{mmvp} and UnderPressure~\cite{UnderPressure} datasets respectively.

\begin{table}[h!]
\centering
\small
\resizebox{0.95\linewidth}{!}{%
\begin{tabular}{llcccc}
\hline
\textbf{Model} & \textbf{Dataset} 
& \textbf{Precision $\uparrow$ (\textit{p}-value)} 
& \textbf{Recall $\uparrow$ (\textit{p}-value)} 
& \textbf{F1 $\uparrow$ (\textit{p}-value)} 
& \textbf{IoU $\uparrow$ (\textit{p}-value)} \\
\hline\hline
FPP-Net~\cite{mmvp} & MMVP 
& $0.635^{\dagger}$ \, {\scriptsize($1.28\mathrm{e}{-2}$)} 
& \textbf{0.600} \, {\scriptsize($6.19\mathrm{e}{-2}$)} 
& 0.583 \, {\scriptsize($6.19\mathrm{e}{-1}$)} 
& 0.448 \, {\scriptsize($7.41\mathrm{e}{-2}$)} \\
Ours & MMVP 
& \textbf{0.650} & 0.588 & \textbf{0.586} & \textbf{0.450} \\
\hline
UP~\cite{UnderPressure} & UnderPressure 
& $0.936^{\dagger}$ \, {\scriptsize($1.62\mathrm{e}{-7}$)} 
& $0.954^{\dagger}$ \, {\scriptsize($9.05\mathrm{e}{-11}$)} 
& $0.945^{\dagger}$ \, {\scriptsize($7.56\mathrm{e}{-7}$)} 
& $0.896^{\dagger}$ \, {\scriptsize($7.16\mathrm{e}{-11}$)} \\
Ours & UnderPressure 
& \textbf{0.942} & \textbf{0.972} & \textbf{0.956} & \textbf{0.917} \\
\hline
\end{tabular}
}
\vspace{0.3cm}
\caption{Foot contact estimation comparison on the MMVP~\cite{mmvp} and UnderPressure~\cite{UnderPressure} datasets.
Each entry shows the mean metric value and the paired \textit{t}-test \textit{p}-value vs.~Ours.
\textbf{Bold} indicates the best (highest metric);
$^{\dagger}$ denotes a statistically significant difference from Ours (\textit{p}~<~0.05).}
\label{tab:contact_pval}
\end{table}

\section{Additional Stability Component Estimation Evaluation}
Table~\ref{tab:cop_bos_pval} reports statistical significance values of paired \textit{t}-tests comparing FootFormer with all baseline models on CoP and BoS estimation across all pressure thresholds (0-25 kPa) for all 10 subjects in \PSUTMM. Table~\ref{tab:com_pval} reports statistical significance values of paired \textit{t}-tests comparing FootFormer with baseline models for CoM estimation across all 10 subjects in \PSUTMM.

\begin{table}[h!]
\centering
\small
\resizebox{0.85\linewidth}{!}{%
\begin{tabular}{lcc}
\toprule
\textbf{Model} & \textbf{CoP Error (mm)} $\downarrow$ & \textbf{BoS IoU} $\uparrow$ \\
\midrule
PNS~\cite{ScottECCV2020} & 65.33 $\pm$ 11.33$^{\dagger}$ {\scriptsize(1.50e-02)} & 0.40 $\pm$ 0.06$^{\dagger}$ {\scriptsize(2.29e-03)} \\
UP~\cite{ScottECCV2020} & 54.33 $\pm$ 13.12$^{\dagger}$ {\scriptsize(1.27e-02)} & 0.48 $\pm$ 0.12$^{\dagger}$ {\scriptsize(1.80e-03)} \\
FPP-Net~\cite{ScottECCV2020} & 96.69 $\pm$ 49.80$^{\dagger}$ {\scriptsize(4.73e-03)} & 0.40 $\pm$ 0.15$^{\dagger}$ {\scriptsize(2.41e-04)} \\
\textbf{Ours} & \textbf{45.85 $\pm$ 11.13} & \textbf{0.56 $\pm$ 0.10} \\
\bottomrule
\end{tabular}}
\vspace{0.3cm}
\caption{Comparison of FootFormer (Ours) with baselines for CoP and BoS metrics on \PSUTMM~\cite{ScottECCV2020}. Each entry shows mean $\pm$ std averaged across subjects and thresholds, and the paired \textit{t}-test \textit{p}-value vs.~Ours. \textbf{Bold} indicates the best (lowest error / highest IoU); $^{\dagger}$ denotes a statistically significant difference from Ours (\textit{p}~<~0.05).}
\label{tab:cop_bos_pval}
\end{table}

\begin{table}[h!]
\centering
\small
\resizebox{0.95\linewidth}{!}{
\begin{tabular}{llcc}
\toprule
\textbf{Metric} & \textbf{Model} & \textbf{Mean L$_2$ Error (mm)} $\downarrow$ & \textbf{Median L$_2$ Error (mm)} $\downarrow$ \\
\midrule
\multirow{3}{*}{CoM} 
& Dempster~\cite{dempster1967properties}    & $48.54 \pm 33.03^{\dagger}$    {\scriptsize(6.72e-06)} & $44.86 \pm 7.53^{\dagger}$ {\scriptsize(2.34e-06)} \\
& CoMNet~\cite{JesseRehabJournal}           & $18.80 \pm 6.66^{\dagger}$     {\scriptsize(4.67e-02)} & $18.31 \pm 7.76^{\dagger}$ {\scriptsize(4.28e-02)} \\
& \textbf{Ours} & $\mathbf{15.51 \pm 7.38}$ & $\mathbf{13.90 \pm 5.63}$ \\
\bottomrule
\end{tabular}}
\vspace{0.3cm}
\caption{Comparison of FootFormer (Ours) with baselines for CoM estimation on \PSUTMM~\cite{ScottECCV2020}. Each entry reports mean $\pm$ std or median $\pm$ rSTD error and the paired \textit{t}-test \textit{p}-value vs.~Ours. \textbf{Bold} indicates the best (lowest error); $^{\dagger}$ denotes a statistically significant difference from Ours (\textit{p}~<~0.05).}
\label{tab:com_pval}
\end{table}

\section{Complete Stability Estimation Evaluation}
\begin{figure}[h!]
\centering
\includegraphics[width=0.9\linewidth]{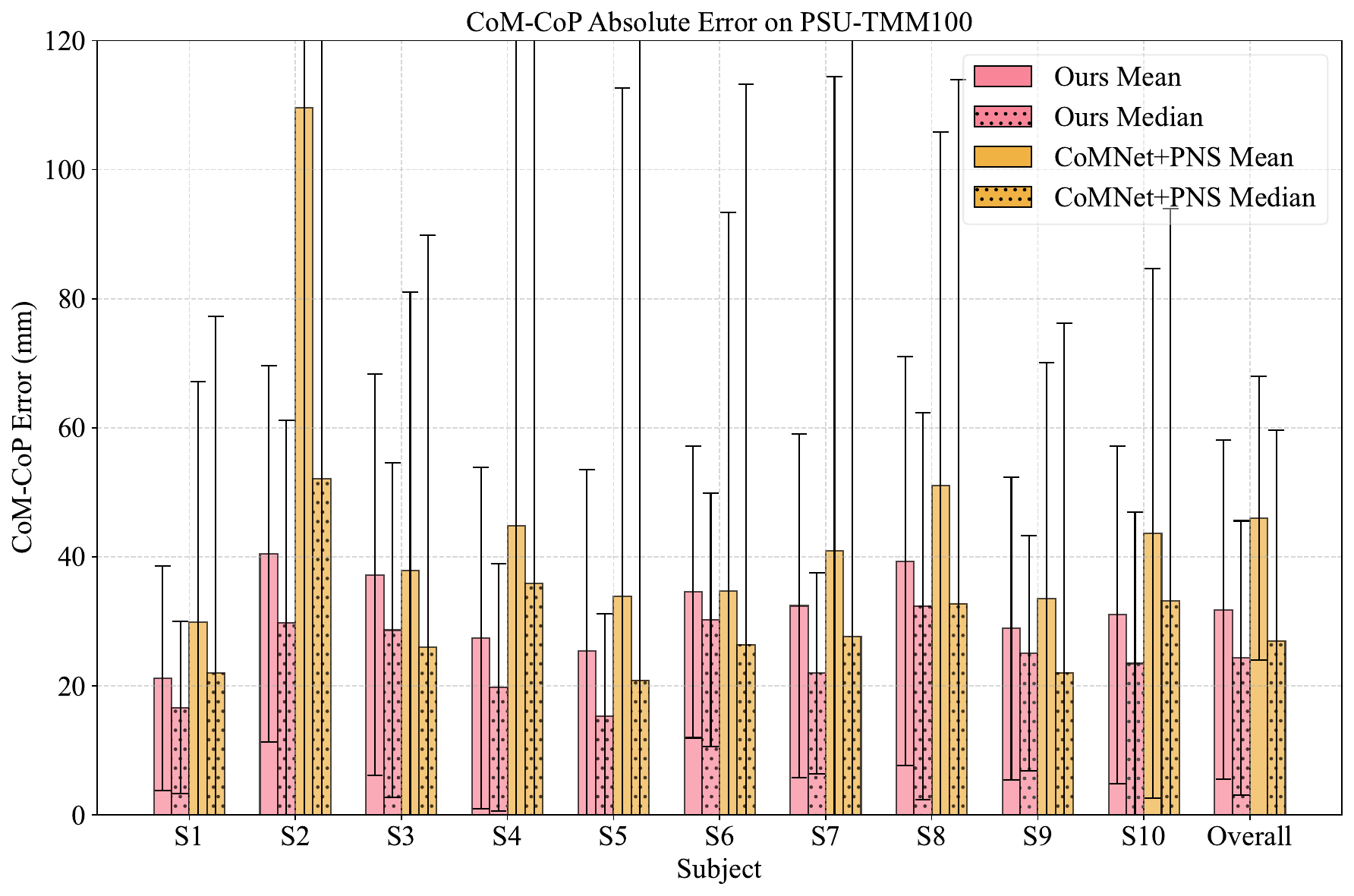}\\
    (a) CoM-to-CoP Absolute Error.
    \\[0.3cm]
    \includegraphics[width=0.9\linewidth]{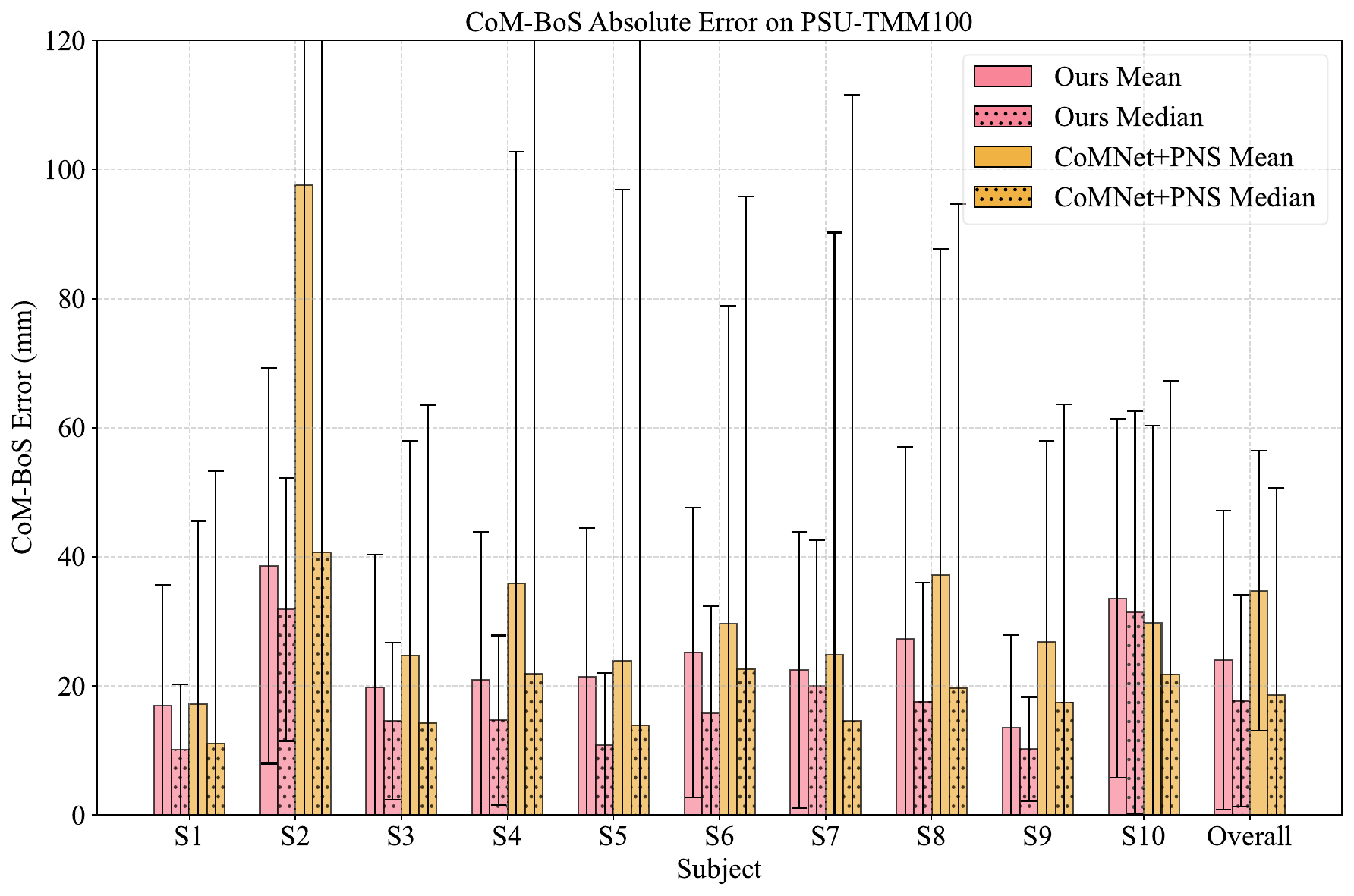}\\
    (b) CoM-to-BoS Absolute Error.
    \vspace{0.4cm}
    \caption{Comparison of CoM-to-CoP and CoM-to-BoS Absolute Error across all 10 subjects and overall in \PSUTMM~\cite{ScottECCV2020}. We report mean and median errors against CoMNet+PNS~\cite{JesseRehabJournal,ScottECCV2020}. Statistical significance values are
    presented in Table~\ref{tab:pval_stability}.}
    \label{fig:com_cop_bos_comparison}
\end{figure}

Figure~\ref{fig:com_cop_bos_comparison} reports the mean $\pm$ std and median $\pm$ rSTD CoM-CoP and CoM-BoS error in mm. Error is computed as the absolute distance between predicted and ground-truth positions derived from the insole sensors and MoCap system used in \PSUTMM. Lastly, we report mean $\pm$ std and median $\pm$ rSTD  absolute error in mm and paired \textit{t}-tests comparing FootFormer with baseline methods in Table~\ref{tab:pval_stability}.

\begin{table}[h!]
\centering
\small
\resizebox{0.95\linewidth}{!}{
\begin{tabular}{llcc}
\toprule
\textbf{Metric} & \textbf{Model} 
& \textbf{Mean Abs. Error (mm) $\downarrow$ (\textit{p}-value)} 
& \textbf{Median Abs. Error (mm) $\downarrow$ (\textit{p}-value)} \\
\midrule

\multirow{2}{*}{CoM-CoP} 
& CoMNet+PNS~\cite{JesseRehabJournal} 
& $46.00 \pm 22.0^{\dagger}$ \, {\scriptsize($2.57\mathrm{e}{-02}$)} 
& $29.86 \pm 13.2^{\dagger}$ \, {\scriptsize($3.63\mathrm{e}{-02}$)} \\
& \textbf{Ours} 
& \textbf{31.80 $\pm$ 27.60} & \textbf{24.33 $\pm$ 8.3} \\
\midrule

\multirow{2}{*}{CoM-BoS} 
& CoMNet+PNS~\cite{JesseRehabJournal} 
& $34.73 \pm 21.7^{\dagger}$ \, {\scriptsize($4.50\mathrm{e}{-02}$)} 
& $19.79 \pm 11.7$ \, {\scriptsize($1.48\mathrm{e}{-01}$)} \\
& \textbf{Ours} 
& \textbf{23.97 $\pm$ 23.16} & \textbf{17.69 $\pm$ 11.3} \\
\bottomrule
\end{tabular}
}
\vspace{0.3cm}
\caption{Comparison of FootFormer (Ours) with CoMNet+PNS~\cite{JesseRehabJournal} on \PSUTMM~\cite{ScottECCV2020}. 
Each entry shows mean $\pm$ std or median $\pm$ rSTD error and the paired \textit{t}-test \textit{p}-value vs.~Ours. 
\textbf{Bold} indicates the best (lowest error); 
$^{\dagger}$ denotes a statistically significant difference from Ours (\textit{p}~<~0.05). 
Corresponding per-subject results are in Fig.~\ref{fig:com_cop_bos_comparison}.}
\label{tab:pval_stability}
\end{table}

\end{document}